\documentclass{article}

\PassOptionsToPackage{round}{natbib}

\usepackage[preprint]{nips_2018}

\usepackage[utf8]{inputenc} 
\usepackage[T1]{fontenc}    
\usepackage[colorlinks]{hyperref}
\usepackage{url}            
\usepackage{booktabs}       
\usepackage{amsfonts}       
\usepackage{amssymb}        
\usepackage{nicefrac}       
\usepackage{microtype}      
\usepackage{graphicx}
\usepackage{subcaption}
\usepackage{wrapfig,booktabs}
\usepackage{enumitem}
\usepackage{xcolor}
\usepackage{kky}
\usepackage{xspace}

\usepackage{amsthm,amsmath}
\usepackage{xcolor}
\usepackage{multirow}

\newcommand*{\eg}{e.g.\@\xspace}
\newcommand*{\ie}{i.e.\@\xspace}

\title{Point Cloud GAN}

\author{
  Chun-Liang Li$^*$, Manzil Zaheer$^*$, Yang Zhang, Barnab{\'a}s P{\'o}czos, Ruslan Salakhutdinov\\
  Machine Learning Department, Carnegie Mellon University, Pittsburgh, PA 15213 \\
  \texttt{\{chunlial,manzilz,yz6,bapoczos,rsalakhu\}@cs.cmu.edu} \\
}

\begin{document}

\maketitle
\begin{abstract}
Generative Adversarial Networks (GAN) can achieve promising performance on learning complex data distributions on different types of data. 
In this paper, we first show a straightforward extension of existing GAN algorithm is not applicable to point clouds, because the constraint required for discriminators is undefined for set data.
We propose a two fold modification to GAN algorithm for learning to generate point clouds (PC-GAN). 
First, we combine ideas from hierarchical Bayesian modeling and implicit generative models by learning a hierarchical and interpretable sampling process. 
A key component of our method is that we train a posterior inference network for the hidden variables.
Second, instead of using only state-of-the-art Wasserstein GAN objective, we propose a \emph{sandwiching} objective, which results in a tighter Wasserstein distance estimate than the commonly used dual form.
Thereby, PC-GAN defines a generic framework that can incorporate many existing GAN algorithms.
We validate our claims on ModelNet40 benchmark dataset.
Using the distance between generated point clouds and true meshes as metric, we find that PC-GAN trained by the sandwiching objective achieves better results on test data than the existing methods. 
Moreover, as a byproduct, PC-GAN learns versatile latent representations of point clouds, which can achieve competitive performance with other unsupervised learning algorithms on object recognition task. 
Lastly, we also provide studies on generating unseen classes of objects and transforming image to point cloud, which demonstrates the compelling generalization capability and potentials of PC-GAN.
\end{abstract}

\section{Introduction}
\label{sec:intro}
A fundamental problem in machine learning is: given a data set, learn a generative model that can efficiently  generate
\emph{arbitrary many new sample points} from the domain of the underlying distribution~\citep{bishop2006pattern}. 
Deep generative models use deep neural networks as a tool for learning complex data distributions~\citep{kingma2013auto,
oord2016pixel, goodfellow2014generative}. 
Especially, Generative Adversarial Networks (GAN; \citealt{goodfellow2014generative}) is drawing attentions because of its
success in many applications.
Compelling results have been demonstrated on different types of data, including text, images and
videos~\citep{lamb2016professor, karras2017progressive, vondrick2016generating}. 
Their wide range of applicability was also shown in many important problems, including data
augmentation~\citep{salimans2016improved}, image style transformation~\citep{zhu2017unpaired}, image
captioning~\citep{dai2017towards} and art creations~\citep{kang2017}. 

Recently, capturing 3D information is garnering attention. 
There are many different data types for 3D information, such as CAD, 3D meshes and point clouds. 
3D point clouds are getting popular since these store more information than 2D images and sensors capable of collecting point clouds have become more accessible. 
These include Lidar on self-driving cars, Kinect for Xbox and face identification sensor on phones.
Compared to other formats, point clouds can be easily represented as a set of points, which has several advantages, such as permutation invariance. 
The algorithms which can effectively learn from this type of data is an emerging 
field~\citep{qi2017pointnet, qi2017pointnet++, zaheer2017deep,kalogerakis20173d, fan2017point}. 
However, compared to supervised learning, unsupervised generative models for 3D data are still under
explored~\citep{achlioptas2017learning,oliva2018transformation}.

Extending existing GAN frameworks to handle 3D point clouds, or more generally set data, is not straightforward. 
In this paper, we begin by formally defining the problem and discussing the difficulty of the problem (Section~\ref{sec:def}).
Circumventing the challenges, we propose a deep generative adversarial network (PC-GAN) with a hierarchical sampling and
inference network for point clouds. 
The proposed architecture learns a stochastic procedure which can generate new point clouds as well as draw samples from point clouds without explicitly modeling the underlying density function  (Section~\ref{sec:algo}).
The proposed PC-GAN is a generic algorithm which can incorporate many existing GAN variants.
By utilizing the property of point clouds, 
we further	propose a \emph{sandwiching} objective by considering both upper and lower bounds of Wasserstein distance
estimate, which can lead to tighter approximation (Section~\ref{sec:div}). 
Evaluation on ModelNet40 shows excellent generalization capability of PC-GAN. 
We first show we can sample from the learned model to generate new point clouds and the 
latent representations learned by the inference network 
provide meaningful interpolations between point clouds. 
We further show the conditional generation
results on \emph{unseen} classes of objects to demonstrate the superior generalization ability of PC-GAN.
Lastly, we also provide several interesting studies, such as classification and point clouds generation from images
(Section~\ref{sec:exp}).

\section{Problem Definition and Difficulty}\label{sec:def}
We begin by defining the problem and ideal generative process for point cloud over objects.
Formally, 
a point cloud for an object $\theta$ is a \emph{set} of $n$ low dimensional vectors $X = \{x_1,...,x_n \}$ with $x_i\in
\mathbb{R}^d$, where $d$ is usually $3$ and $n$ can be infinite.
Point cloud for $M$ different objects is a collection of $M$ sets $X^{(1)}, ..., X^{(M)}$, where each $X^{(m)}$ is as defined above.
Thus, we basically need a generative model $p(X)$ for sets which should be able to:
\begin{itemize}
\item Sample entirely new sets $X$, as well as
\item Sample more points for a given set, \ie $x \sim p(x|X)$.
\end{itemize}

De-Finetti theorem allows us to express the set probability in a factored format as $p(X) = \int_\theta \prod_{i=1}^n p(x_i|\theta)p(\theta) d\theta$ for some suitably defined $\theta$. 
In case of point clouds, the latent variable 
\begin{wrapfigure}{r}{0.5\textwidth}
\vspace{-2mm}
\centering
\begin{subfigure}[b]{.23\linewidth}
\captionsetup{width=0.8\textwidth}
\captionsetup{format=hang}
\includegraphics[width=\linewidth,trim={35mm 40mm 35mm 70mm},clip]{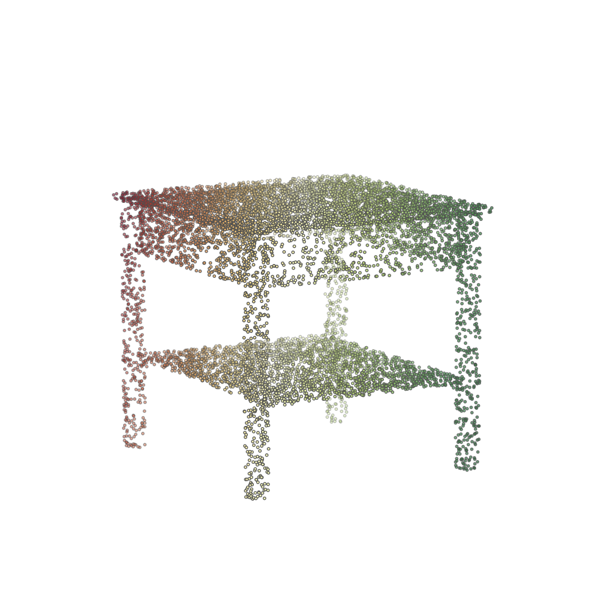}
\caption{Table: $p(x_i|\theta_1)$}
\end{subfigure}
\begin{subfigure}[b]{.23\linewidth}
\captionsetup{width=0.8\textwidth}
\captionsetup{format=hang}
\includegraphics[width=\linewidth,trim={30mm 50mm 30mm 50mm},clip]{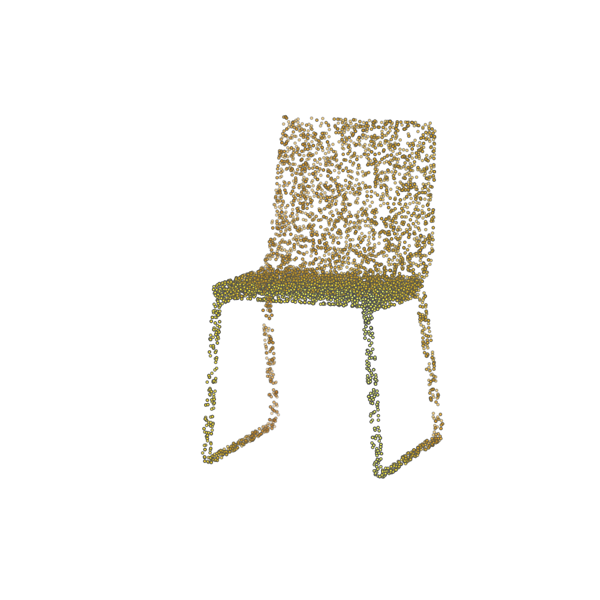}
\caption{Chair: $p(x_i|\theta_2)$}
\end{subfigure}
\begin{subfigure}[b]{.23\linewidth}
\captionsetup{width=0.8\textwidth}
\captionsetup{format=hang}
\includegraphics[width=\linewidth,trim={45mm 55mm 45mm 60mm},clip]{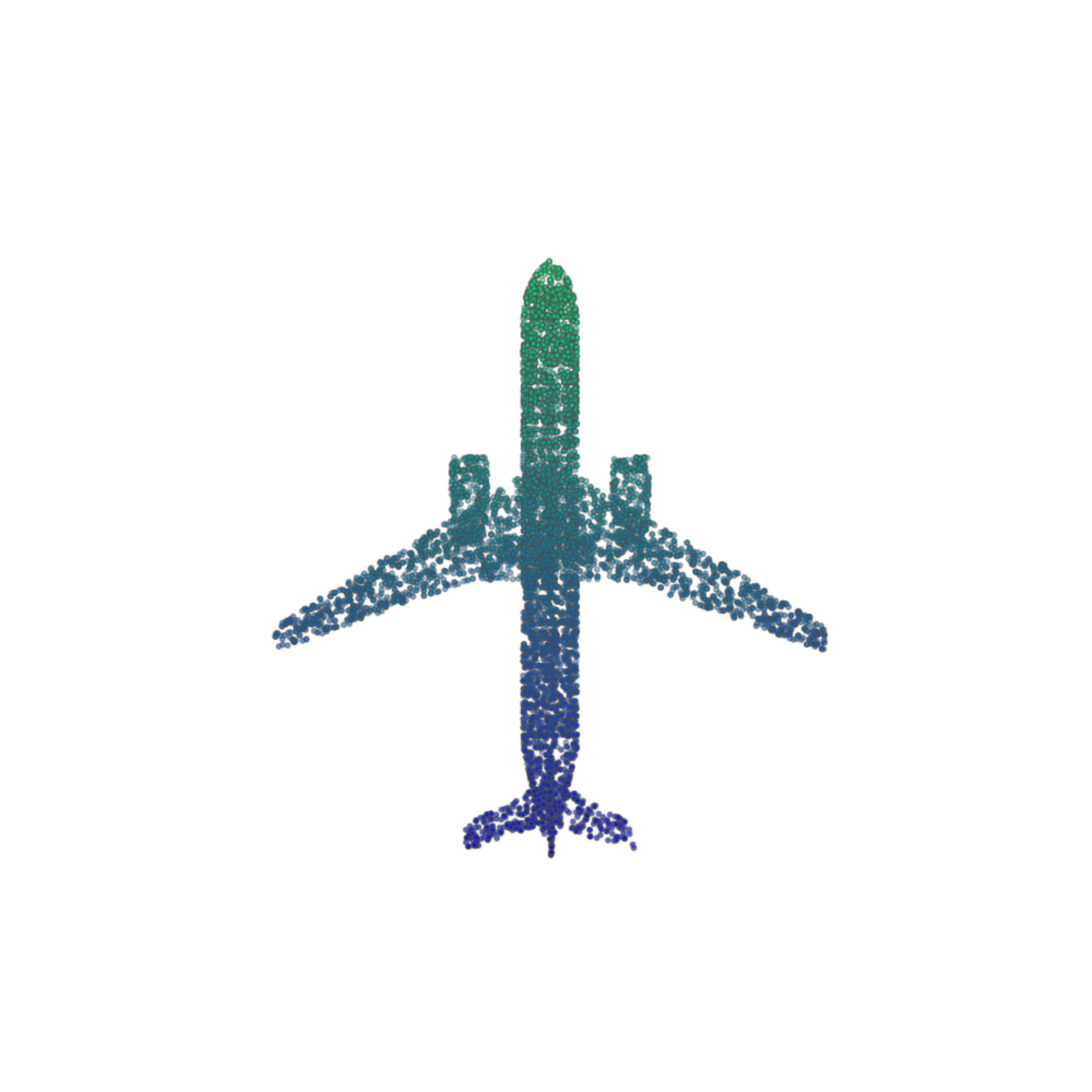}
\caption{Plane: $p(x_i|\theta_3)$}
\end{subfigure}
\begin{subfigure}[b]{.27\linewidth}
\captionsetup{width=0.85\textwidth}
\includegraphics[width=\linewidth,trim={60mm 70mm 60mm 70mm},clip]{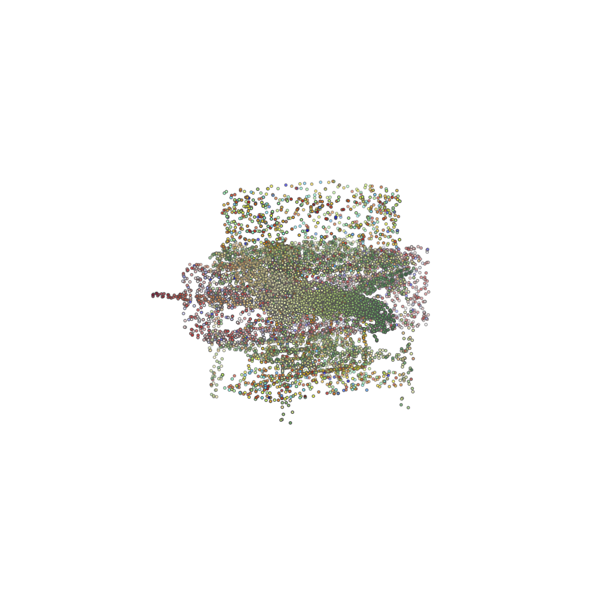}
\caption{Marginalized: $p(x_i)$}
\end{subfigure}
\caption{Unlike in case of images, the marginal distribution $p(x_i) = \int p(x_i|\theta) p(\theta) d\theta$ is not useful. We have to learn the joint distribution $p(X,\theta)$. An illustrative parallel can be drawn between pixels of an image and point cloud --- marginal distribution of pixels is quite uninformative, one has to consider the joint distribution of all pixels in one image (not across different images).}
\label{fig:marginal}
\vspace{-12mm}
\end{wrapfigure}
$\theta$ can be interpreted as an object representation.
In this view, the factoring can be understood as
follows:
Given an object, $\theta$, the points $x_i$ in the point cloud can be considered as i.i.d.
samples from $p(x|\theta)$, an unknown latent distribution representing object $\theta$.
Joint likelihood can be expressed as:
\begin{equation}
p(X, \theta) = \underbrace{p(\theta)}_{\text{object}} \; \; \; \underbrace{\prod_{i=1}^n p(x_i|\theta)}_{\text{points for object}}
\label{eq:sets:joint}
\end{equation}
Attempts have been made to characterize~\eqref{eq:sets:joint} with parametric models
like Gaussian Mixture Models or parametric hierarchical models \citep{jian2005robust,strom2010graph,eckart2015mlmd}.
However, such approaches have limited success as the point cloud conditional density $p(x|\theta)$ is highly non-linear and complicated (example of point clouds can be seen in Figure~\ref{fig:marginal}, \ref{fig:overview}, \ref{fig:reconstruction}).

With advent of implicit generative models, like GANs~\citep{goodfellow2014generative}, it is possible to model complicated  distributions, but it is not clear how to extend the existing GAN frameworks to handle the hierarchical model of point clouds as developed above. 
The aim of most GAN framework\citep{goodfellow2014generative,arjovsky2017wasserstein, li2017mmd} is to learn to generate new samples from a distribution $p(x)$, for fixed dimensional $x\in\mathbb{R}^d$, given a training dataset. We, in contrast, aim to develop a new method where the training data is a set of sets, and learning from this training data would allow us to generate points from this hierarchical distribution (e.g. point clouds of 3D objects). To re-emphasize the incompatibility, note that the training data of traditional GANs is a \emph{set} of fixed dimensional instances, in our case, however, it is a \emph{set of sets}. 
Using existing GANs to just learning marginal distribution $p(x)$, in case of point clouds $x$, is not of much use because the marginal distribution is quite uninformative as can be seen from Figure~\ref{fig:marginal}.

One approach can be used to model the distribution of the point cloud set together, i.e., $\{ \{x_i^{(1)}\}_{i=1}^n, \dots, \{x_i^{(m)}\}_{i=1}^n \}$. 
In this setting, a na\'ive application of traditional GAN is possible through treating the point cloud as finite
dimensional vector by fixing number and order of the points (reducing the problem to instances in
$\mathbb{R}^{n\times3}$) with DeepSets~\citep{zaheer2017deep} classifier as the discriminator to distinguish real sets from fake sets.
However, it would not work because the IPM guarantees behind the traditional GAN no longer hold (\eg in case of
\citet{arjovsky2017wasserstein}, nor are 1-Lipschitz functions over sets well-defined). The probabilistic divergence approximated by a
DeepSets  classifier is not clear and might be ill-defined. 
Counter examples for breaking IPM guarantees can be easily found as we show next.
 
\begin{wrapfigure}{r}{0.43\textwidth}
\vspace{-5mm}
\includegraphics[width=\linewidth]{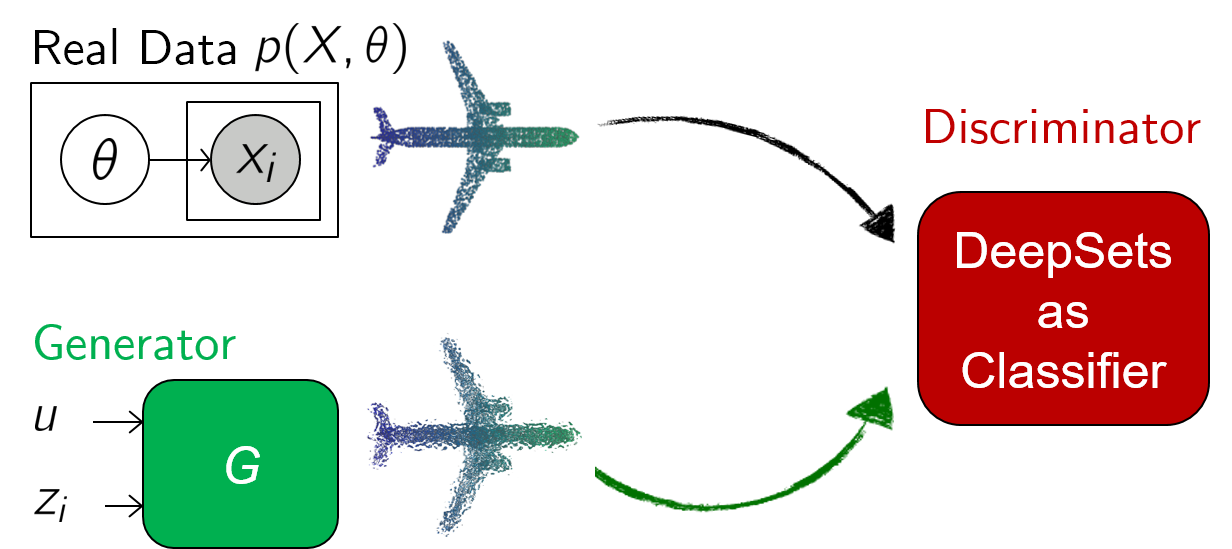}
\caption{Natural extension of GAN to handle set data does not work.}
\label{fig:failed}
\vspace{-2mm}
\end{wrapfigure}
\paragraph{Counter Example} Consider a simple GAN \citep{goodfellow2014generative} with a DeepSets classifier as the discriminator. In order to generate coherent sets of variable size, we consider a generator $G$ having two noise sources: $u$ and $z_{i}$. To generate a set, $u$ is sampled once and $z_{i}$ is sampled for $i=1,2,...,n$ to produce $n$ points in the generated set. Intuitively, fixing the first noise source $u$ selects a set and ensures the points generated by repeated sampling of $z_{i}$ are coherent and belong to same set. The setup is depicted in Figure~\ref{fig:failed}. In this setup, the GAN minimax problem would be:
\begin{equation}
 \min_{G}\max_{D} \underset{ \substack{\theta \sim p(\theta) \\ x_i \sim p(x_i|\theta)} }{\mathbb{E}} \left[\log D\left( \{ x_i \} \right)\right] \; + \underset{ \substack{u \sim p(u) \\ z_{i} \sim p(z_{i})} }{\mathbb{E}} \left[\log\left(1-D\left( \{ G(u, z_{i}) \} \right) \right)\right]
\end{equation}
Now consider the case, when there exist an `oracle' mapping $T$ which maps each sample point deterministically 
to the object it originated from, i.e. $\exists T: T(\{x_i\})=\theta$. A valid example is when different $\theta$ leads to conditional distribution $p(x|\theta)$ with non-overlapping support. Let $D=D'\circ T$ and $G$ ignores $z$ then optimization becomes
\begin{equation}
\begin{aligned}
&\min_{G}\max_{D'} \underset{ \substack{\theta \sim p(\theta) \\ x_i \sim p(x_i|\theta)} }{\mathbb{E}} \left[\log D'\left(T\left( \{ x_i \} \right)\right)\right] \; + \underset{ \substack{u \sim p(u) \\ z_{i} \sim p(z_{i})} }{\mathbb{E}} \left[\log\left(1-D' \left( T \left( \{ G(u, z_{i}) \} \right) \right) \right)\right]
\\
\Rightarrow & \min_{G}\max_{D'} \underset{ \substack{\theta \sim p(\theta) \\ x_i \sim p(x_i|\theta)} }{\mathbb{E}} \left[\log D'\left( \theta \right)\right] \; + \underset{ \substack{u \sim p(u) \\ z_{i} \sim p(z_{i})} }{\mathbb{E}} \left[\log\left(1-D'\left( T \left( \{ G(u) \} \right) \right) \right)\right]
\\
\Rightarrow & \min_{G}\max_{D'} \underset{ \substack{\theta \sim p(\theta) } }{\mathbb{E}} \left[\log D'\left( \theta \right)\right] \; + \underset{ \substack{u \sim p(u) } }{\mathbb{E}} \left[\log\left(1-D'\left( T \left( \{ G(u) \} \right) \right) \right)\right]
\end{aligned}
\end{equation}
Thus, we can achieve the lower bound $-\log(4)$ by only matching the $p(\theta)$ component, while the conditional $p(x|\theta)$ is allowed to remain arbitrary. 
We note that there still exists good solutions, which can lead to successful training,  other than this hand-crafted example. 
We found empirically that GAN with simple DeepSet-like discriminator most of the times fails to
learn to generate point clouds even after converging. However, sometimes it does results in reasonable generations.
So simply using DeepSets classifier \emph{without any constraints} in simple GAN in order to handle sets does not always lead to a valid generative model.
We need additional constraints for GANs with simple DeepSet-like discriminator to exclude such bad solutions and lead to a
more stable training.

\section{Proposed Method}
\label{sec:algo}
\begin{wrapfigure}{r}{0.43\textwidth}
\vspace{-6mm}
\centering
\includegraphics[width=\linewidth]{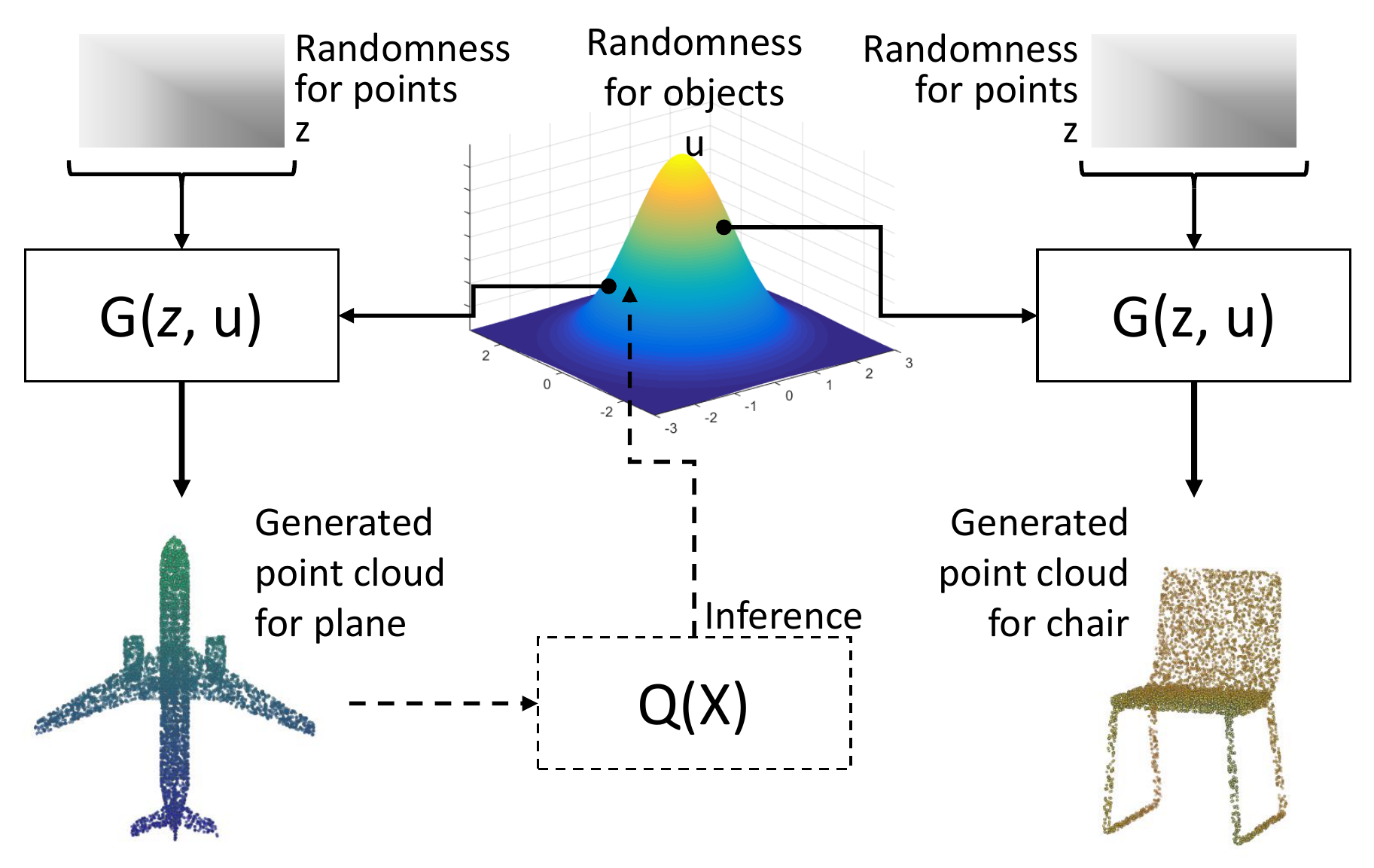}
\caption{Overview of PC-GAN.}
\label{fig:overview}
\vspace{-12mm}
\end{wrapfigure}

Although directly learning point cloud generation under GAN formulation is difficult as
described in Section~\ref{sec:def}, given a $\theta$, learning $p(X|\theta)$ is reduced to learning a 2 or 3 dimensional
distribution, which fits standard GAN settings.
Formally, given a $\theta$, we train a generator $G_{x|\theta}(z)$ such that $x = G_{x|\theta}(z)$, where $z\sim p(z)$,
follows $\GG$ by optimizing a (peudo) probabilistic divergence $D(\PP\|\GG)$ between the 
distribution $\GG$ of $G_{x|\theta}(z)$ and $p(x|\theta)$, which is denoted as $\PP$. The full objective can be written as
$\displaystyle \EE_{\theta\sim p(\theta)} \bigg[ \min_{G_x} D(\PP\|\GG) \bigg]$.

\paragraph{Inference} Although GANs have been extended to learn conditional distributions~\citep{mirza2014conditional,isola2017image}, 
they require conditioning variables to be observed, such as the one-hot label or a given image.
While in case of point clouds we only have partial knowledge of the conditional, \ie we only have groupings of point
coming from the same object but we have no representation $\theta$ of the conditional or the object other than the points themselves.
Na\"ively modeling $\theta$ to be a one-hot vector, to indicate which object the points belong to in the training data, cannot generalize to unseen testing data. 
Instead, we need a richer representation for $\theta$, which is an unobserved random variable.
Thus, we need to infer $\theta$ during the training.
The proposed algorithm has to concurrently learn the inference network $Q(X)$ which encodes $\theta$ while we learn $p(x|\theta)$. 

\paragraph{Neural Network Realization}
Our solution comprises of a generator $G_x(z, \psi)$ which takes in a noise source $z\in\mathbb{R}^{d_1}$ and a descriptor $\psi\in\mathbb{R}^{d_2}$ encoding information about distribution of $\theta$. 
Here $d_1$ is dimensionality of the per-point noise source $z$ and $d_2$ is the size of the descriptor vector. 
Typically, we would need $d_1 \ll d_2$, as $psi$ needs to encode much more information because it decides the whole shape of the conditional distribution whereas $z$ just corresponds to different samples from it.
Another interpretation of the descriptor $\psi$ can be as an embedding of distribution over $\theta$.
For a given $\theta_0$, the descriptor $\psi$ would encode 
information about the distribution $\delta(\theta-\theta_0)$ and samples 
generated as $x=G_x(z,\psi)$ would follow the distribution $p(x|\theta_0)$. 
More generally, $\psi$ can be used to encode more complicated distributions regarding $\theta$ as
well. In particular, it could be used to encode the posterior $p(\theta|X)$ for a given sample set $X$,  
such that $x=G_x(z,\psi)$ follows the posterior predictive distribution:
\[
    p(x|X) = \int p(x|\theta)p(\theta|X) d\theta.
\]

As $\psi$ is unobserved, we use an inference network $Q(X)$ to learn the 
most informative descriptor $\psi$ about the distribution $p(\theta|X)$ to minimize the 
The divergence between $p(X|\theta)$, which is denoted as $\PP$, and the distribution of $G_x(z, Q(X))$ given $Q(X)$,
which we abuse $\GG$ for convenience, that is
$\displaystyle \EE_{\theta\sim p(\theta)} \bigg[ \min_{G_x, Q} D(\PP\|\GG) \bigg]$.

A major hurdle in taking this path is that $X$ is a set of points, which can vary in size and permutation of elements.
Thus, making design of $Q$ complicated as traditional neural network can not handle this and possibly is the reason for
absence of such framework in the literature despite being a natural solution for the important problem of generative
modeling of point clouds. However, we can overcome this challenge and we propose to construct the inference network by 
utilizing the recent advance in deep learning for dealing with sets~\citep{qi2017pointnet, zaheer2017deep}.
This allows it handle variable number of inputs points in arbitrary order, yet yielding a consistent descriptor $\psi$.

\paragraph{Hierarchical Sampling} After training $G_{x}$ and $Q$, we use trained $Q$
to collect inferred $Q(X)$ and train another generator $G_\theta(u) \sim p(\psi)$ for higher hierarchical sampling.
Here $u\in \mathbb{R}^{d_3}$ is the other noise source independent of $z$ and dimensionality $d_3$ this noise source is typically much smaller than $d_2$.
In addition to layer-wise training, a joint training could further boost performance. 
The full generative process for sampling one point cloud could be represented as 
\[
\{x_i\}_{i=1}^n = \{G(z_i, u)\}_{i=1}^n = \{G_x(z_i, G_\theta(u))\}_{i=1}^n, \mbox{ where }z_1,\dots,z_n\sim p(z),
\mbox{ and }u\sim p(u).
\]
We call the proposed algorithm for point cloud generation as \emph{PC-GAN} as shown in Figure~\ref{fig:overview}.
The conditional distribution matching with a learned inference in PC-GAN can also be interpreted as an
encoder-decoder formulation~\citep{kingma2013auto}. The difference between it and the point cloud autoencoder~\citep{achlioptas2017learning,
yang2018foldingnet} will be discussed in Section~\ref{sec:related}.

\section{Different Divergences for Matching Point Clouds}
\label{sec:div}

Given two point clouds $x\sim p(X|\theta)$ and $G_x(z, Q(X))$, one commonly used heuristic distance measure is chamfer
distance~\citep{achlioptas2017learning}. 
On the other hand, if we treat each point cloud as a \emph{3 dimensional distribution}, we can 
adopt a more broader classes of probabilistic divergence $D(\PP \| \GG)$ for training $G_x$. 
Instead of using divergence esitmates which
requires density estimation~\citep{jian2005robust,strom2010graph,eckart2015mlmd}, we are interested in implicit generative models with a
GAN-like objective~\citep{goodfellow2014generative}, which has been demonstrated to learn  complicated  distributions. 

To train the generator $G_x$ using a GAN-like objective for point clouds, we need a discriminator $f(\cdot)$ which 
distinguishes between the generated samples and
true samples conditioned on $\theta$. Combining with the inference network discussed in Section~\ref{sec:algo},
if we use an IPM-based GAN~\citep{arjovsky2017wasserstein, mroueh2017fisher, mroueh2017sobolev}, the
objective can be written as
\begin{equation}
    \EE_{\theta\sim p(\theta)} \bigg[ \min_{G_x, Q} 
	\underbrace{
	\max_{f\in \Omega_{f}}  \EE_{x\sim p(X|\theta)}\left[ f(x) \right] - 
	\EE_{z\sim p(z), X\sim p(X|\theta)}\left[ f( G_x(z, Q(X))) \right]}_{D(\PP\|\GG)} 
	\bigg],
	\label{eq:full}
\end{equation}
where $\Omega_f$ is the constraint for different probabilistic distances, such as 1-Lipschitz~\citep{arjovsky2017wasserstein}, $L^2$ ball~\citep{mroueh2017fisher} or Sobolev ball~\citep{mroueh2017sobolev}. 

\subsection{Tighter Solutions via Sandwiching}
\label{sec:sandwiching}
\begin{wrapfigure}{r}{0.4\textwidth}
\vspace{-6mm}
\includegraphics[width=\linewidth,trim={0.5cm 0 0.0cm 0},clip]{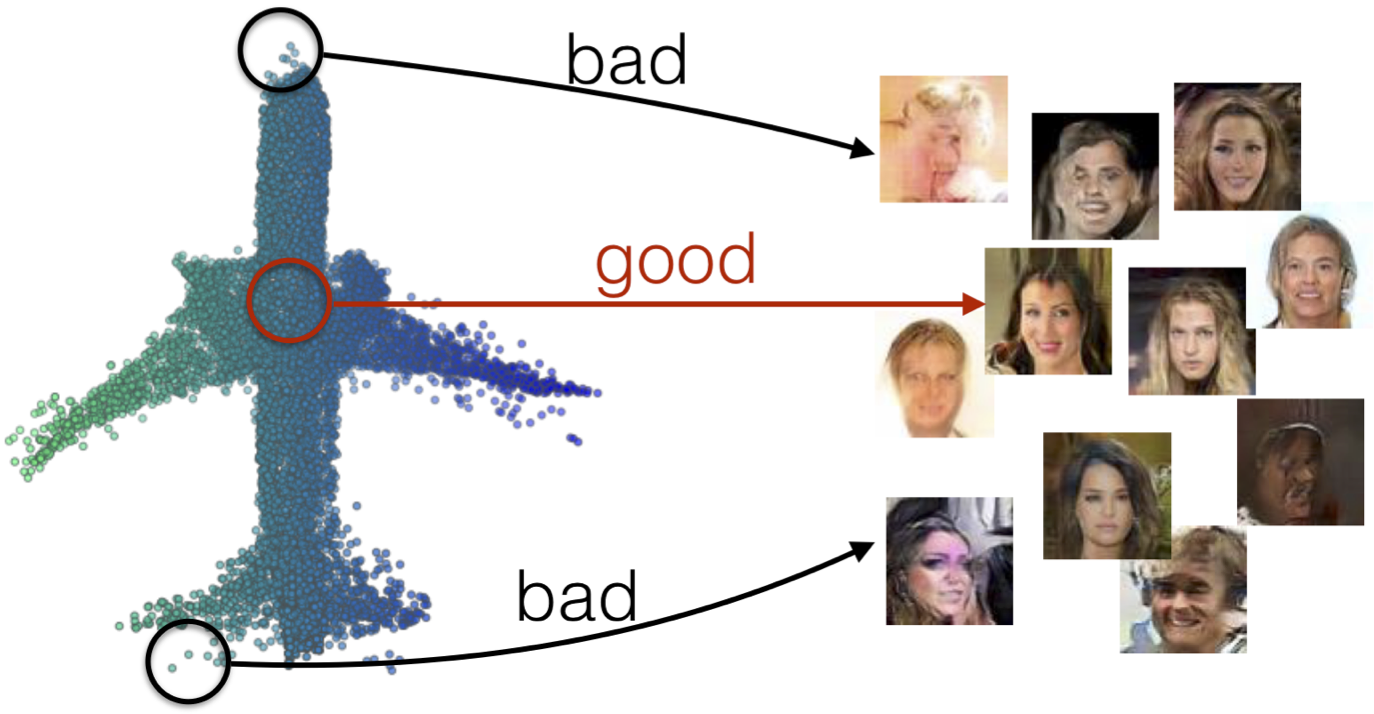}
\vspace{-5mm}
        \caption{Connection between good/bad points and faces generated from a GAN.}
        \label{fig:faces}
\end{wrapfigure}

In our setting, each point $x_i$ in the point cloud can be considered to correspond to single images when we train
GANs over images. 
An example is illustrated in Figure~\ref{fig:faces}, where the samples from MMD-GAN
\citep{li2017mmd} trained on CelebA consists of both good and bad faces. In case of images, when quality is
evaluated, it primarily focuses on coherence individual images and the few bad ones are usually left out. Whereas in
case of point cloud, to get representation of an object we need many sampled points together and presence of outlier points
degrades the quality of the object. 
Thus, when training a generative model for point cloud, we need to ensure a much
lower distance $D(\PP\|\GG)$ between true distribution $\PP$ and generator distribution $\GG$ than would be needed for images.

We begin by noting that the popular Wasserstein GAN \citep{arjovsky2017wasserstein}, aims to optimize $G$ 
by $\min w(\PP, \GG)$, where $w(\PP, \GG)$ is 
the Wasserstein distance $w(\PP,\GG)$ between the truth $\PP$ and generated distribution $\GG$ of $G$. 
Many GAN works (\eg \citet{arjovsky2017wasserstein}) approximate $w(\PP,\GG)$ in dual form (a maximization problem),
such as~\eqref{eq:full}, by
neural networks. The resulting estimate $W_L(\PP,\GG)$ is a lower bound of the true Wasserstein distance, 
as neural networks can only recover a subset of 1-Lipschitz functions~\citep{arora2017generalization} required in the dual form.
However, finding a lower bound $W_L(\PP, \GG)$ for $w(\PP, \GG)$ may not be an ideal surrogate for solving a minimization problem $\min w(\PP, \GG)$.
In optimal transport literature, Wassertein distance is usually estimated by approximate matching cost,
$W_U(\PP,\GG)$, which gives us an upper bound of the true Wasserstein distance.

We propose to combine, in general, a lower bound and upper bound estimate by sandwiching the solution between the two, i.e. we solve the following minimization problem:
\begin{equation}
\begin{aligned}
&\min_G && W_U(\PP,\GG) &&\\
&\text{such that} && W_U(\PP,\GG) - W_L(\PP,\GG) < \lambda &&
\end{aligned}
\end{equation}
The problem can be simplified and solved using method of Lagrange multipliers as follows:
\begin{equation}
\min_G W_\lambda(\PP,\GG) := (1-\lambda) W_U(\PP,\GG) + \lambda W_L(\PP,\GG)
\label{eq:sandwich}
\end{equation}
By solving the new \emph{sandwiched problem} \eqref{eq:sandwich}, we show that under certain conditions we obtain a better
estimate of Wasserstein distance in the following lemma:

\begin{lemma}
\label{lem:sandwiching}
Suppose we have two approximators to Wasserstein distance: an upper bound $W_U$ and a lower $W_L$, such that $\forall
P,G: (1+\epsilon_1)w(\PP,\GG) \leq W_U(\PP,\GG) \leq (1+\epsilon_2)w(\PP,\GG)$ and $\forall P,G: (1-\epsilon_2)w(\PP,\GG) \leq W_L(\PP,\GG) \leq (1-\epsilon1)w(\PP,\GG)$ respectively, for some $\epsilon_2>\epsilon_1>0$ and $\epsilon_1>\epsilon_2/3$. Then, using the sandwiched estimator $W_\lambda$ from \eqref{eq:sandwich}, we can achieve tighter estimate of the Wasserstein distance than using either one estimator, \ie
\begin{equation}
\exists \lambda: |W_\lambda(\PP,\GG) - w(\PP,\GG)| < \min\{|W_U(\PP,\GG) - w(\PP,\GG)|, |W_L(\PP,\GG) - w(\PP,\GG)|\}
\end{equation}
\end{lemma}

\subsubsection{Upper Bound Implementation}
\label{sec:upper}
The primal form of Wasserstein distance is defined as 
\[
	w(\PP,\GG) = \inf_{\gamma\in \Gamma(\PP,\GG)} \int \| x-y \|_1 d\gamma(x, y), 
\]
where $\gamma$ is the \emph{coupling} of $P$ and $G$.
The Wasserstein distance is also known as optimal transport (OT) or earth moving distance (EMD). 
As the name suggests, when $w(\PP,\GG)$ is estimated with finite number of samples $X={x_1, \dots, x_n}$ and $Y={y_1,\dots, y_n}$, we find the one-to-one matching between $X$ and $Y$ such that the total pairwise distance is minimal. 
The resulting minimal total (average) pairwise distance is $w(X, Y)$.
In practice, finding the exact matching efficiently is non-trivial and still an open research problem~\citep{peyre2017computational}. 
Instead, we consider an approximation provided by \citet{bertsekas1985distributed}.
It is an iterative algorithm where each iteration operates like an auction whereby unassigned points $x\in X$ bid simultaneously for closest points $y\in Y$, thereby raising their prices. 
Once all bids are in, points are awarded to the highest bidder.
The crux of the algorithm lies in designing a non-greedy bidding strategy.
One can see by construction the algorithm is embarrassingly parallelizable, which is favourable for GPU implementation. 
One can show that algorithm terminates with a valid matching and the resulting matching cost $W_U(X,Y)$ is an $\epsilon$-approximation of $w(X, Y)$. 
Thus, the estimate can serve as an upper bound, i.e.
\begin{equation}
	w(X,Y) \leq W_U(X,Y) \leq (1+\epsilon)w(X, Y),
\end{equation}

We remark estimating Wasserstein distance $w(\PP,\GG)$ with \emph{finite samples} via primal form is only favorable to low dimensional data, 
such as point clouds. 
The error of empirical estimate in primal is $O(1/n^{1/d})$~\citep{weed2017sharp}.
When the dimension $d$ is large (\eg images), we cannot accurately estimate $w(\PP, \GG)$ in primal as well as its upper bound 
with a small minibatch. 

Finding a modified primal form with low sample complexity, especially for high dimensional data, is still an open research
problem~\citep{cuturi2013sinkhorn, genevay2018learning}. Combining those into the proposed sandwiching objective for high dimensional data is left for future works. 

\subsection{Lower Bound Implementation}
\label{sec:lower}
The dual form of Wasserstein distance is defined as 
\begin{equation}
	w(\PP,\GG) = \sup_{f \in \Lcal_1} \EE_{x \sim P} f(x) - \EE_{x \sim G}f(x),
	\label{eq:wdual}
\end{equation}
where $\Lcal_k$ is the set of $k$-Lipschitz functions whose  Lipschitz constant is no larger than $k$.
In practice, deep neural networks parameterized by $\phi$ with constraints $f_\phi \in \Omega_{\phi}$~\citep{arjovsky2017wasserstein}, result in a distance approximation 
\begin{equation}
	W_L(\PP,\GG) = \max_{f_\phi \in \Omega_{\phi}} \EE_{x \sim P} f_\phi(x) - \EE_{x \sim G}f_\phi(x).
	\label{eq:wapprox}
\end{equation}

If there exists $k$ such that $\Omega_f\subseteq \Lcal_k$, then 
$W_L(\PP,\GG)/k \leq w(\PP,\GG)~\forall P, G$ is a lower bound.
To enforce $\Omega_\phi\subseteq \Lcal_k$, \citet{arjovsky2017wasserstein} propose a weight
clipping constraint~$\Omega_c$, which constrains every weight to be in $[-c, c]$ and guarantees that 
$\Omega_c \subseteq \Lcal_k$ for some $k$. 
Based on~\cite{arora2017generalization}, the Lipschitz functions $\Omega_c$ realized by neural networks is actually
$\Omega_c \subset \Lcal_k$.

In practice, choosing clipping range $c$ is non-trivial. Small ranges limit the capacity of networks,
while large ranges result in numerical issues during the training. 
On the other hand, in addition to weight clipping, several constraints (regularization) have bee proposed with better
empirical performance, such as gradient penalty~\citep{gulrajani2017improved} and $L^2$ ball~\citep{mroueh2017fisher}. 
However, there is no guarantee the resulted functions are still Lipschitz or the resulted distances are lower bounds of
Wasserstein distance.    
To take the advantage of those regularization with the Lipschitz guarantee, we propose a simple variation by combining
weight clipping, which always ensures Lipschitz functions. 

\begin{lemma}\label{prop:wc}
There exists $k>0$ such that 
\begin{equation}
	\max_{f\in \Omega_c \cap \Omega_{\phi}} \EE_{x \sim P}[f_\phi(x)] - \EE_{x \sim G}[f_\phi(x)] \leq \frac{1}{k}w(\PP,\GG)	
\end{equation}
\end{lemma}
Note that, if $c\rightarrow\infty$, then $\Omega_c \cap \Omega_{\phi} = \Omega_{\phi}$. 
Therefore, from Proposition~\ref{prop:wc}, for any regularization of discriminator~\citep{gulrajani2017improved, mroueh2017fisher,
mroueh2017sobolev}, we can always combine it with a weight clipping constraint $\Omega_c$ to ensure a valid lower bound
estimate of Wasserstein distance and enjoy the advantage that it is numerically stable when we use large $c$ compared
with original weight-clipping WGAN~\citep{arjovsky2017wasserstein}.

\section{Related Works}
\label{sec:related}

Generative Adversarial Network~\citep{goodfellow2014generative} aims to learn a generator that can sample data
followed by the data distribution.  Compelling results on learning complex data distributions with GAN have
been shown on images~\citep{karras2017progressive}, speech~\citep{lamb2016professor},
text~\citep{yu2016seqgan,HjelmJCCB17}, vedio~\citep{vondrick2016generating} and 3D voxels~\citep{wu2016learning}.
However, the GAN algorithm on 3D point cloud is still under
explored~\citep{achlioptas2017learning}.
Many alternative objectives for training GANs have been
studied.
Most of them are the \emph{dual form} of $f$-divergence~\citep{goodfellow2014generative, mao2016least,
nowozin2016f}, integral probability metrics (IPMs)~\citep{zhao2016energy, li2017mmd, arjovsky2017wasserstein, gulrajani2017improved}
or IPM extensions~\citep{mroueh2017fisher, mroueh2017sobolev}.
\citet{genevay2018learning} learn the generative model by the approximated primal form of Wasserstein distance~\citep{cuturi2013sinkhorn}.

Instead of training a generative model on the data space directly, one popular approach is combining with autoencoder
(AE), which is called adversarial autoencoder (AAE)~\citep{makhzani2015adversarial}. AAE constrain the encoded data to
follow normal distribution via GAN loss, which is similar to VAE~\citep{kingma2013auto} by replacing the KL-divergence 
on latent space via any GAN loss. 
\citet{tolstikhin2017Wasserstein} provide a theoretical explanation for AAE by connecting it with the primal form of
Wasserstein distance.
The other variant of AAE is training the other generative model to learn the distribution of the encoded data instead of
enforcing it to be similar to a known distribution~\citep{engel2017latent, kim2017adversarially}.
\citet{achlioptas2017learning} explore a AAE variant for point cloud. 
They use a specially-designed encoder network~\citep{qi2017pointnet} for learning a compressed representation for point clouds before training GAN on the latent space.
However, their
decoder is restricted to be a MLP which generates $m$ fixed number of points, where $m$ has to be pre-defined. 
That is, the output of their decoder is fixed to be $3m$ for 3D point clouds, while the output of the proposed $G_x$ is
only 3 dimensional and $G_x$ can generate arbitrarily many points by sampling different random noise $z$ as input.  
\citet{yang2018foldingnet, groueix2018papier} propose similar decoders to $G_x$ with fixed grids 
to break the limitation of~\citet{achlioptas2017learning} aforementioned, but they use heuristic Chamfer distance without any theoretical
guarantee and do not exploit generative models for point clouds.
The proposed PC-GAN can also be interpreted as an encoder-decoder formulation. However, the underlying interpretation is
different. We start from De-Finetti theorem to learn both $p(X|\theta)$ and $p(\theta)$ with inference network
interpretation of $Q$, while ~\citet{achlioptas2017learning} focus on learning $p(\theta)$ without modeling
$p(X|\theta)$.

GAN for learning conditional distribution (conditional GAN)
has been studied in images with single conditioning~\citep{mirza2014conditional,pathak2016context,isola2017image,chang2017one}
or multiple conditioning~\citep{wang2016generative}. The case on point cloud is still under explored. Also, most of the
works assume the conditioning is given (\eg labels and base images) without learning the inference during the training. 
Training GAN with inference is studied by~\citet{dumoulin2016adversarially, li2017alice}; however, their goal is to infer
the random noise $z$ of generators instead of semantic latent variable of the data.  
\citet{li2018graphical} is a parallel work aiming to learn GAN and unseen latent variable simultaneously, but they
only study image and video datasets.

Lastly, we briefly review some recent development of deep learning on point clouds. 
Instead of transforming into 3D voxels and projecting objects into different views to use
convolution~\citep{su2015multi, maturana2015voxnet, wu20153d, qi2016volumetric, tatarchenko2017octree}, which have the concern of memory usage, 
one direction is designing permutation invaraint operation for dealing with set data directly~\citet{qi2017pointnet,
zaheer2017deep, qi2017pointnet++}. 
\citet{wang2018dynamic} use graph convolution~\citep{bronstein2017geometric} to utilize local neighborhood information.
Most of the application studied by those works focus on classification and segmentation tasks, but they can be used to 
implement the inference network $Q$ of PC-GAN.

\section{Experiments}\label{sec:exp}
In this section we demonstrate the point cloud generation capabilities of PC-GAN. 
As discussed in
Section~\ref{sec:related}, we refer \citet{achlioptas2017learning} as AAE as it could be treated as an AAE extension to
point clouds and we use the implementation provided by the authors for experiments.
The sandwitching objective $W_s$ for PC-GAN combines $W_L$  and $W_U$ with the mixture 1:20 without tunning for all experiment.
$W_L$ is a GAN loss by combining~\citet{arjovsky2017wasserstein} and~\citet{mroueh2017fisher} 
and we adopt~\citep{bertsekas1985distributed} for $W_U$ as discussed in Section~\ref{sec:lower} and
Section~\ref{sec:upper}.
We parametrize $Q$ in PC-GAN by DeepSets~\citep{zaheer2017deep}.
The review of DeepSets is in Appendix~\ref{sec:pelayer}. 
Other detailed configurations of each experiment can be found in Appendix~\ref{sec:config}.
Next, we study both synthetic 2D point cloud and ModelNet40 benchmark datasets. 

\subsection{Synthetic Datasets}
\label{sec:exp_syn}
\begin{figure}
\centering
\begin{subfigure}[b]{.23\linewidth}
\includegraphics[width=\linewidth]{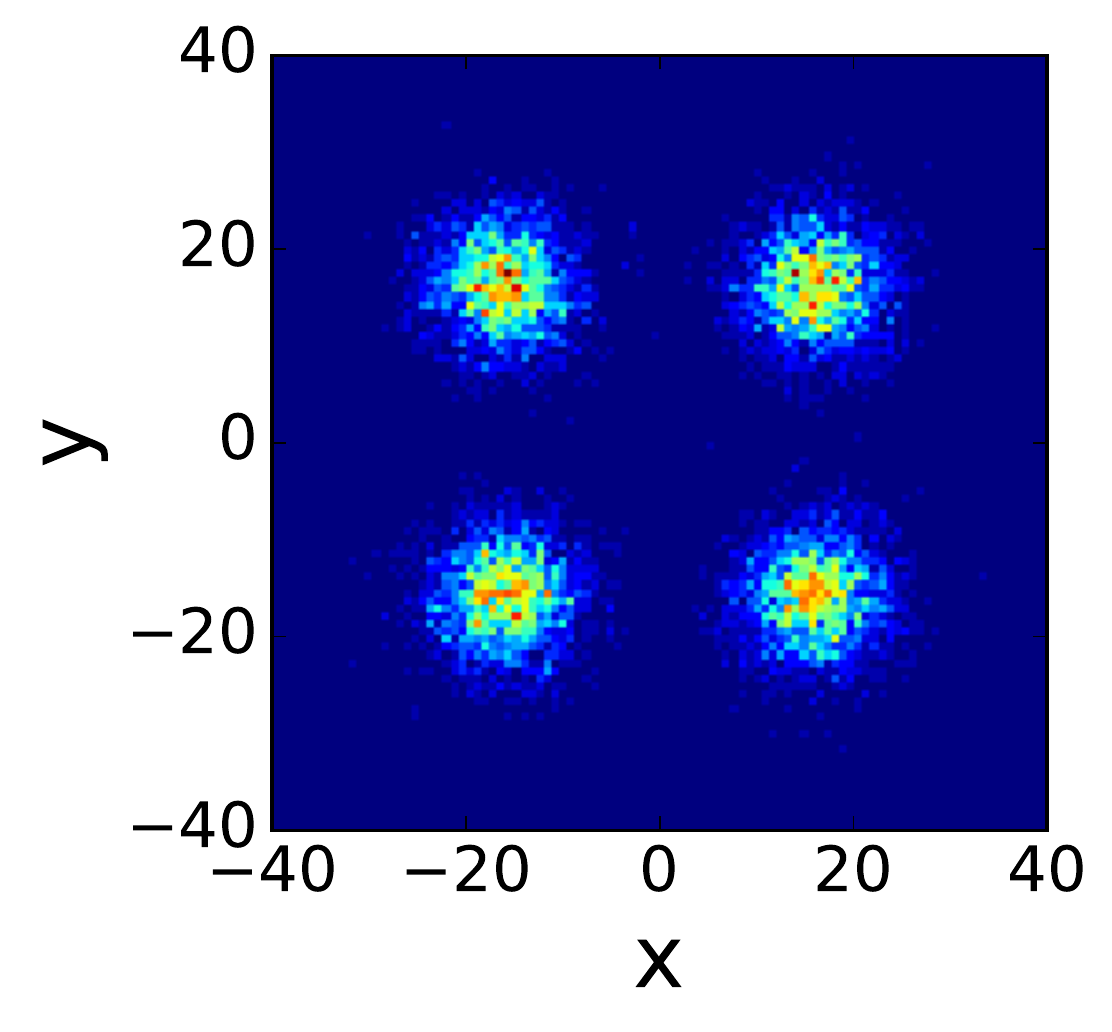}
\end{subfigure}
\begin{subfigure}[b]{.23\linewidth}
\includegraphics[width=\linewidth]{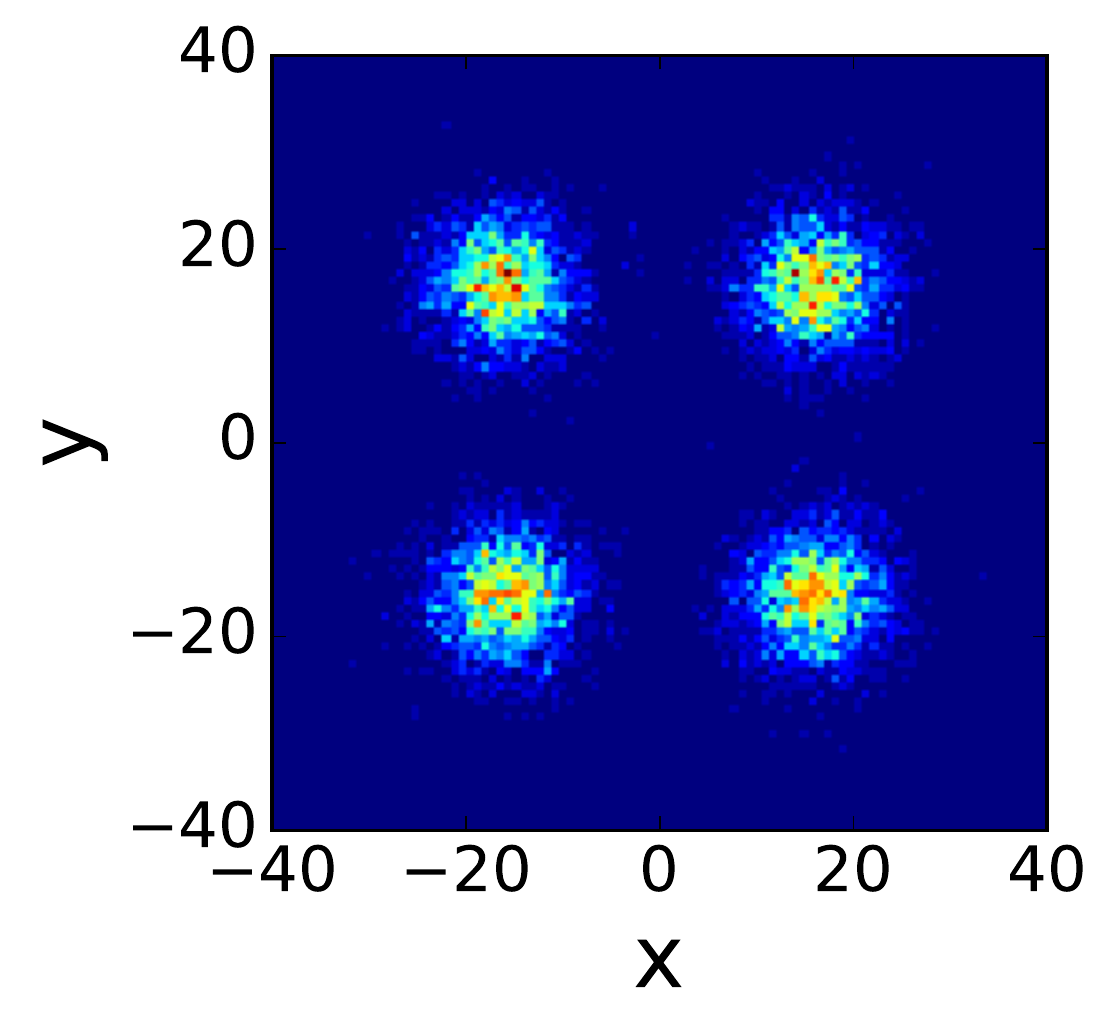}
\end{subfigure}
\begin{subfigure}[b]{.23\linewidth}
\includegraphics[width=\linewidth]{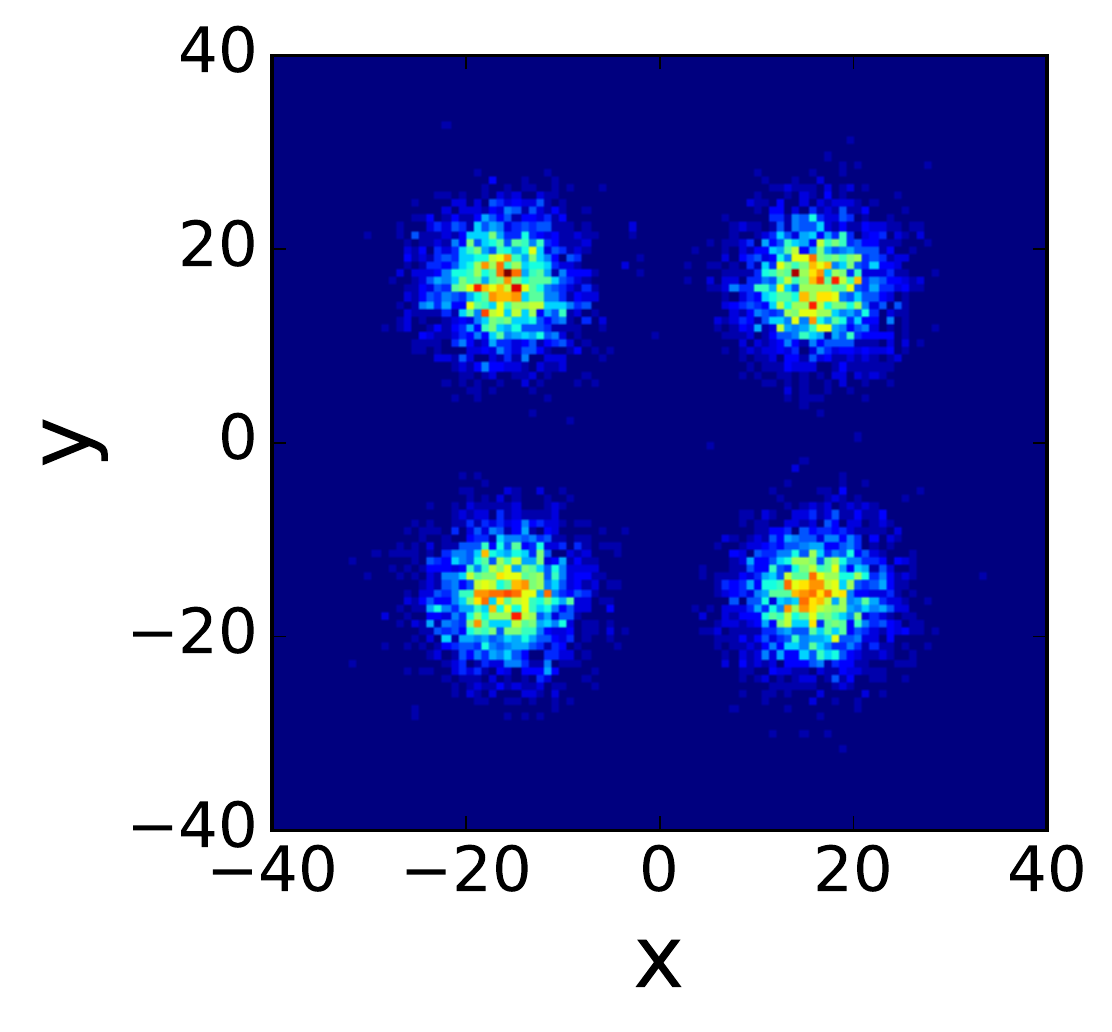}
\end{subfigure}
\begin{subfigure}[b]{.23\linewidth}
\includegraphics[width=\linewidth]{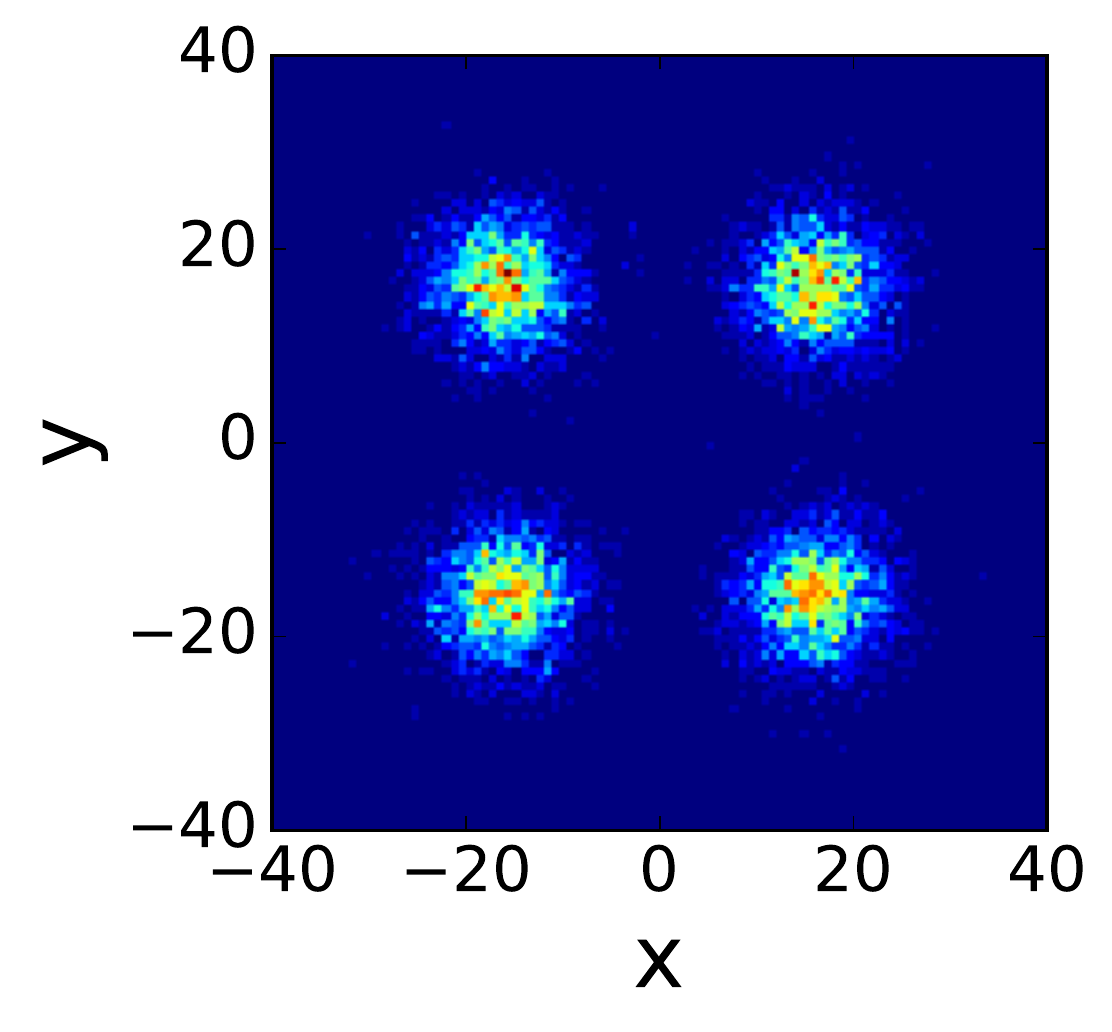}
\end{subfigure}
\begin{subfigure}[b]{.23\linewidth}
\includegraphics[width=\linewidth]{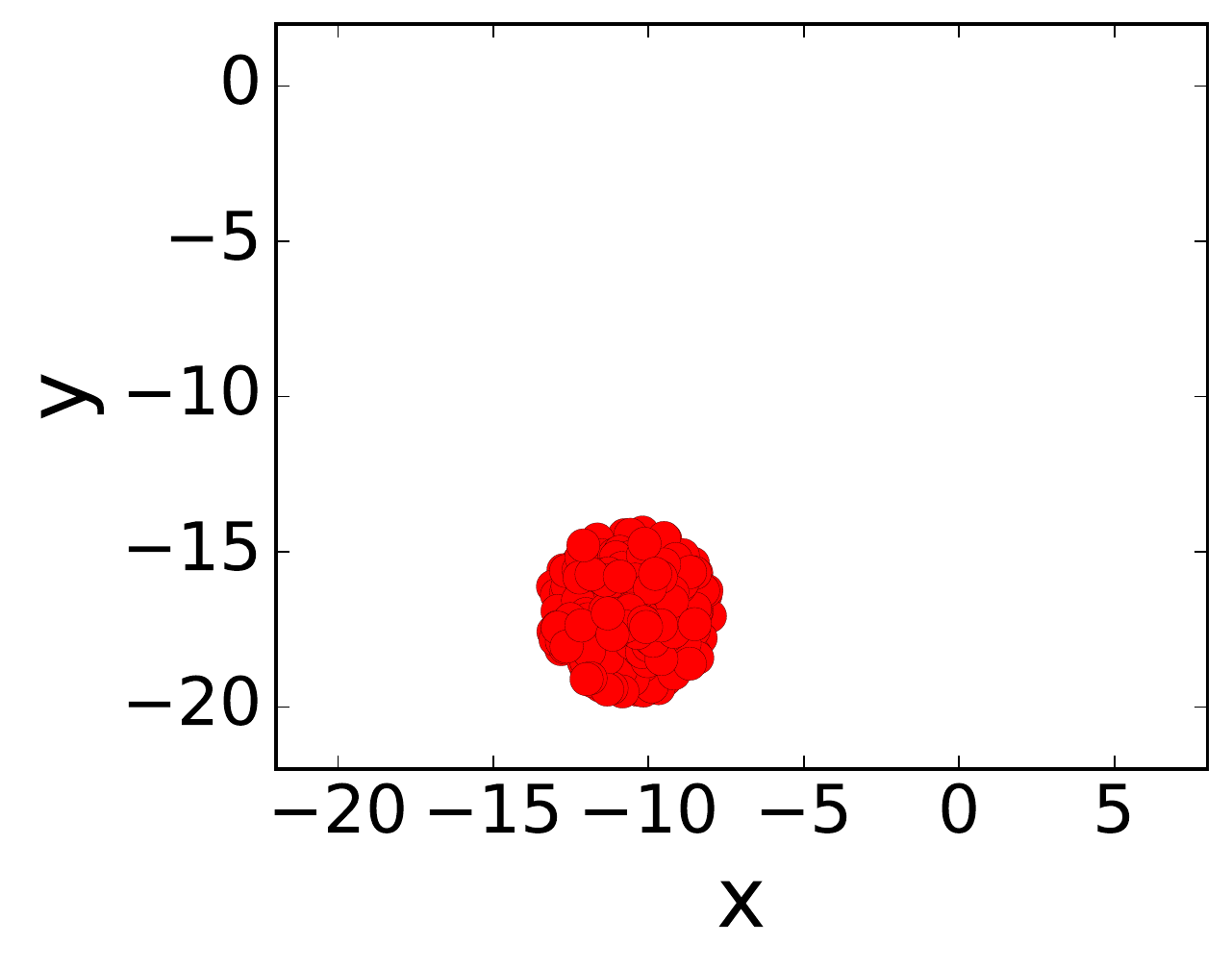}
\caption{True Data}
\label{fig:circle_true}
\end{subfigure}
\begin{subfigure}[b]{.23\linewidth}
\includegraphics[width=\linewidth]{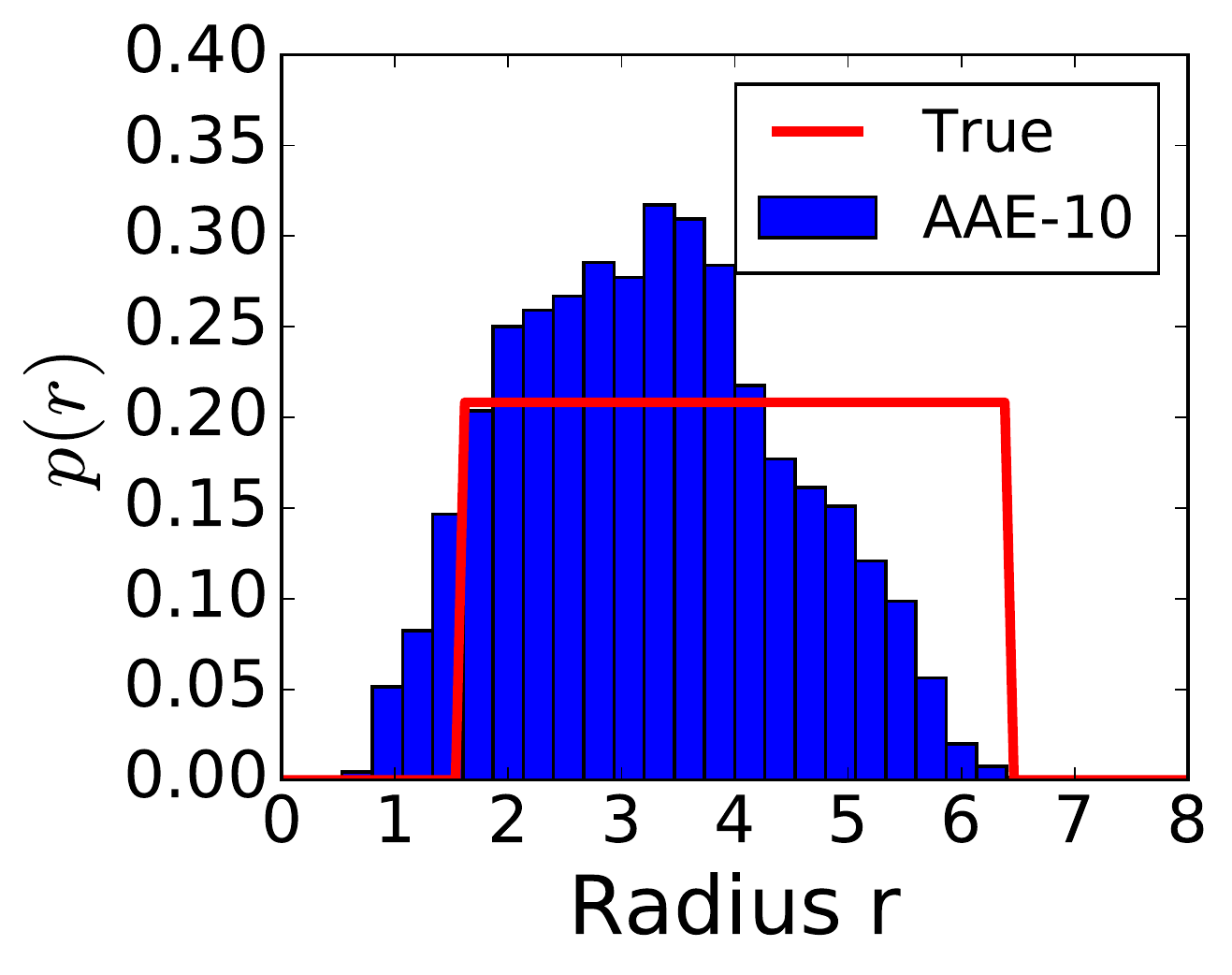}
\caption{AAE-10}
\end{subfigure}
\begin{subfigure}[b]{.23\linewidth}
\includegraphics[width=\linewidth]{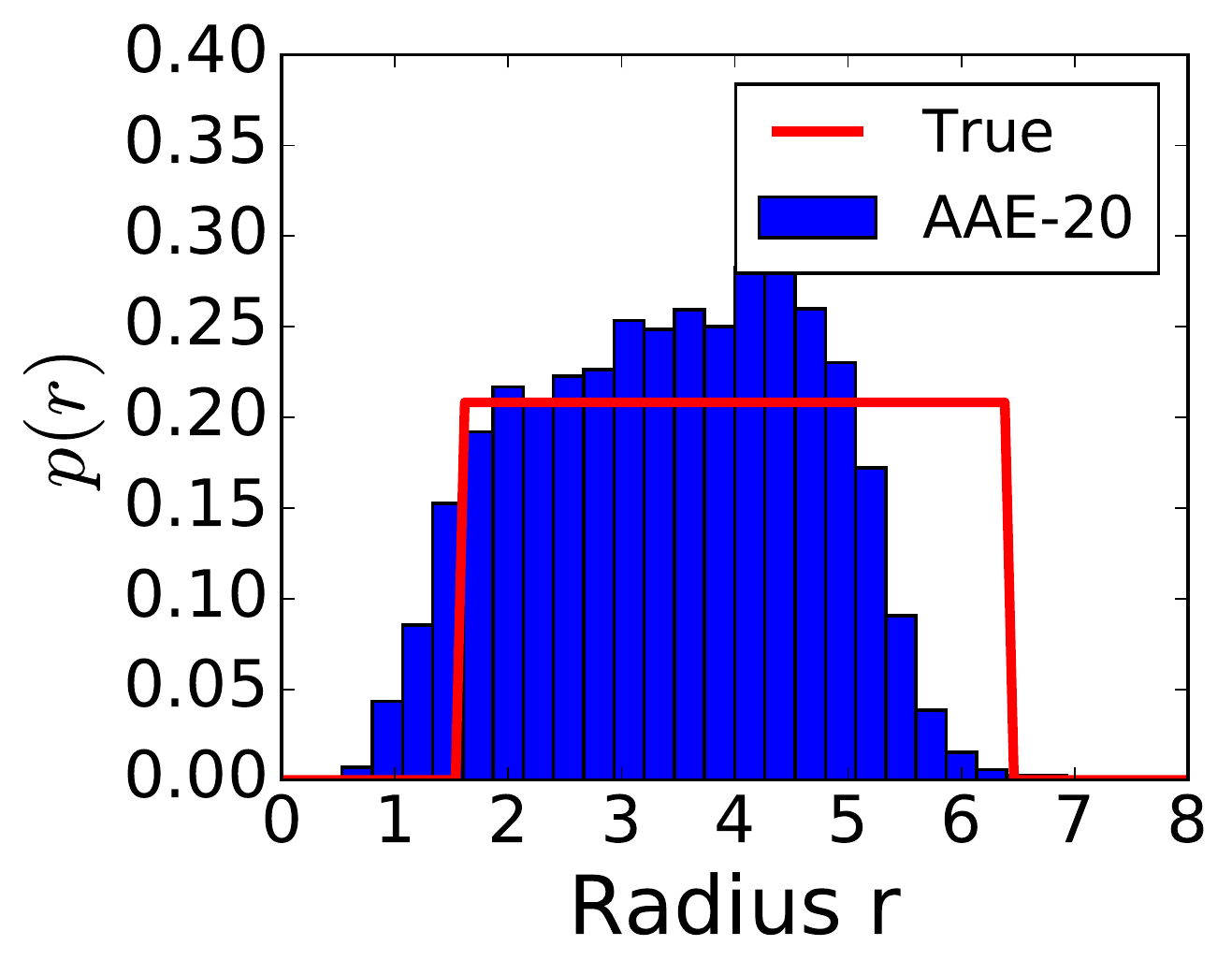}
\caption{AAE-20}
\end{subfigure}
\begin{subfigure}[b]{.23\linewidth}
\includegraphics[width=\linewidth]{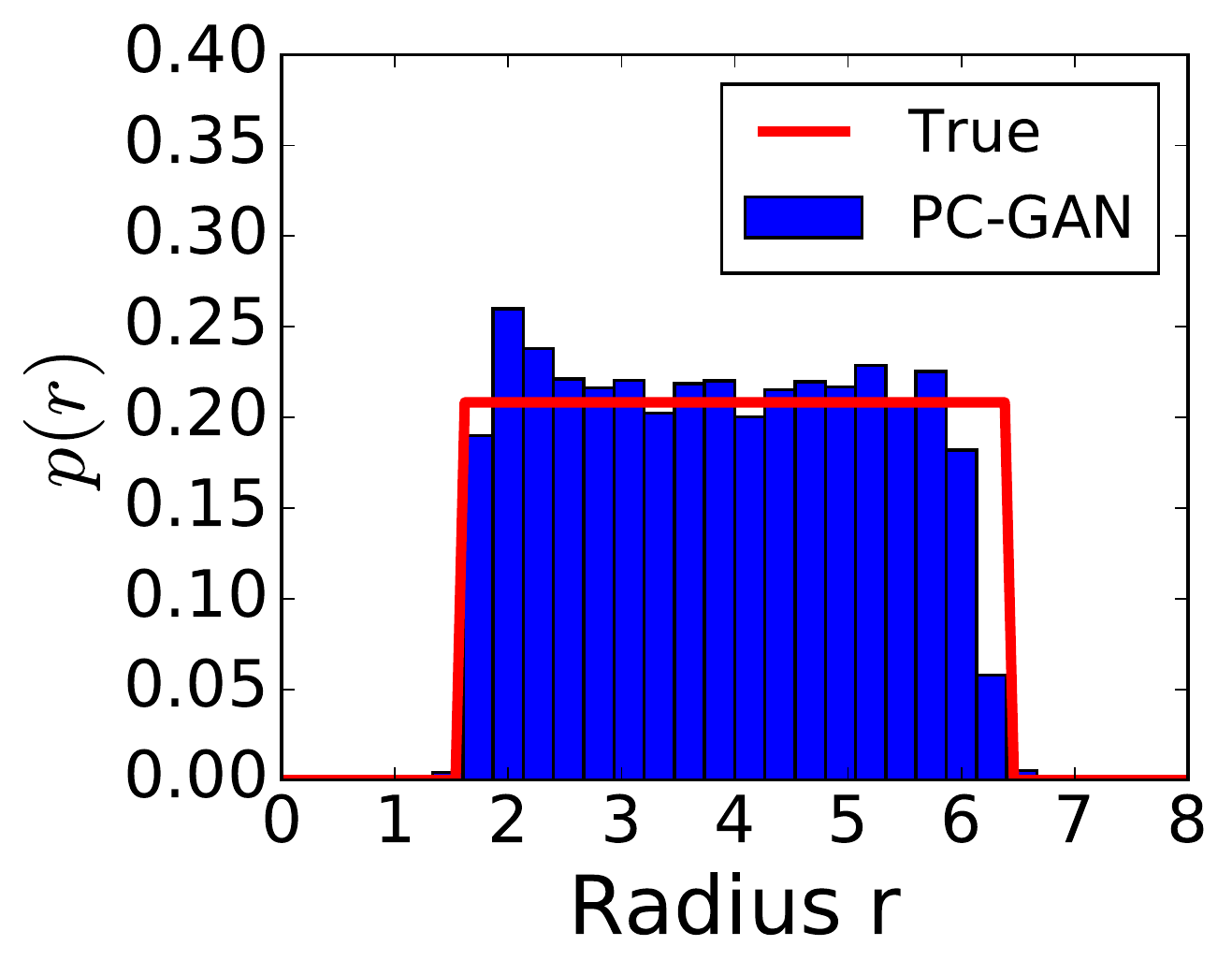}
\caption{PC-GAN}
\end{subfigure}
\caption{The reconstructed center and radius distributions. (a) (top) the true center distribution and (bottom) one example of the
2D circle point cloud. (b-d) are the reconstructed center and radius distributions of different algorithms. }
\label{fig:circle}
\end{figure}

We created a simple 2D synthetic point cloud datasets from parametric distributions on which we can carry out
thorough evaluations of the prposed PC-GAN and draw comparisons with AAE~\citet{achlioptas2017learning}.
We generate 2D point clouds for circles, where the center of circles is followed a mixture of four Gaussians
with means equal to $\{\pm 16\}\times\{\pm 16\}$. The covariance matrices were set to be $16I$ and we used equal mixture weights.
The radius of the circles was drawn from a uniform distribution $\mbox{\it Unif}(1.6, 6.4)$. One sampled circile is shown in
Figure~\ref{fig:circle_true}. We sampled $10,000$ circles for the training and testing data, respectively. 

For PC-GAN, the inference network $Q$ is a stack of 3 Permutation Equivariance Layers with hidden layer size to be $30$
and output size to be $15$ (\eg $Q(X)\in \RR^{15}$).
In this experiment the total number of parameters for PC-GAN is $12K$.
For AAE encoder, we follow the same setting in~\citet{achlioptas2017learning}; for decoder, we
increase it from 3 to be 4 layers with larger capacities and set output size to be $500\times 2$ dimensions for $500$ points. 
We consider two model configurations AAE-10 and AAE-20, which use $10$ and $20$ units for the hidden layers of AAE
decoder, respectively. The total number of parameters (encoder+decoder) are $14K$ and $24K$ for AAE-10 and AAE-20, respectively. 
Detailed model configurations are provided in the supplementary material. 

We evaluated the conditional distributions on the $10,000$ testing circles.
For the proposed PC-GAN, we pass the same points into the inference network $Q$, then sample $500$
points with the conditional generator $G_x$ to match the output number of AAE. 
We measured the empirical distributions of the centers and the radius
of the generated circles conditioning on the testing data for PC-GAN. Similarly, we measured the reconstructed circles of
the testing data for AAE.
The results are shown in Figure~\ref{fig:circle}. 

From Figure~\ref{fig:circle}, both AAE and PC-GAN can successfully recover the center distribution, but AAE does not
learn the radius distribution well.  Even if we increase number the hidden layer unit of the decoder to be $20$
(AAE-20), which almost doubles the number of parameters, the performance is still not satisfactory.  Compared with AAE,
the proposed PC-GAN recovers the both center and radius distributions well with less parameters.  The gap of memory
usage could be larger if we configure AAE to generate more points, while the model size required for PC-GAN is
independent of the number of points. The reason is MLP decoder adopted by~\citet{achlioptas2017learning} wasted
parameters for nearby points. A much larger model (more parameters) can potentially boost the performance, yet would be
still restricted to generate a fixed number of points for each object as discussed in
Section~\ref{sec:related}.

\subsection{Conditional Generation on ModelNet40}
We consider the ModelNet40~\citep{wu20153d} benchmark, which contains 40 classes of objects. There are $9,843$ training
and $2,468$ testing instances. We follow~\citet{zaheer2017deep} to do pre-processing. 
For each object, we sampled $10,000$ points from the mesh representation and normalize it to have zero mean (for each
axis) and unit (global) variance. 
During the training, we augment the data by uniformly rotating $0, \pi/8, \dots, 7\pi/8$ rad on the $x$-$y$ plane.
For PC-GAN, the random noise $z_2$ is fixed to be $10$ dimensional for all experiments. 
For other settings, we follow~\citet{achlioptas2017learning}. 

\paragraph{Training on Single Class} We start from a smaller model which is only trained on single class of objects. 
For AAE, the latent code size for its encoder is $128$ and the decoder outputs $2,048$ points for each object. 
The number of parameters for encoder and decoder are $15M$ in total.
Similarly, we set the size of PC-GAN latent variable (the output of $Q$) to be $128$ dimensional. 
The number of parameters for $G_x$ and $Q$ is less than $1M$ in total.

\paragraph{Training on All Classes.} We also train the proposed model on all $9,843$ objects in the training set. 
The size of AAE latent code of is increased to be $256$. The number of parameters of its encoder and decoder is $15.2M$. 
We set the size of PC-GAN latent variable to be $256$ dimensional as well. 
The number of parameters for $G_x$ and $Q$ are around $3M$ in total. 

\subsubsection{Quantitative Comparison}

\begin{table}[t]
\newcommand{\ra}[1]{\renewcommand{\arraystretch}{#1}}
\ra{1.2}
\centering
\caption{The quantitative results of different models trained on different subsets of ModelNet40 and evaluated on the
corresponding testing set. ModelNet10 is a subset containing 10 classes of objects, while ModelNet40 is a full training set.  
AAE is trained using the code provided by~\citet{achlioptas2017learning}. Three PC-GAN are trained via
upper bound $W_U$, lower bound $W_L$ and sandwiching loss $W_s$ discussed in Section~\ref{sec:div}. 
}
    \resizebox{\linewidth}{!}{
    \begin{tabular}{@{} l r r r r c r r r r @{}} 
    \toprule
    \multirow{2}{*}{Data} & \multicolumn{4}{c}{ Distance to Face $(\mbox{D2F} \downarrow)$ }&&  \multicolumn{4}{c}{Coverage $(\uparrow)$}
    \\ 
    \cline{2-5}\cline{7-10} &PC-GAN ($W_S$) & AAE  & PC-GAN ($W_U$) & PC-GAN ($W_L$) && PC-GAN ($W_S$) & AAE &  PC-GAN ($W_U$) &
	PC-GAN ($W_L$) \\
    \midrule
    Aeroplanes& 1.89E+01  & 1.99E+01 & 1.53E+01 & 2.49E+01 && 1.95E-01 & 2.99E-02 & 1.73E-01 & 1.88E-01 \\
    Benches   & 1.09E+01  & 1.41E+01 & 1.05E+01 & 2.46E+01 && 4.44E-01 & 2.35E-01 & 2.58E-01 & 3.83E-01 \\
    Cars      & 4.39E+01  & 6.23E+01 & 4.25E+01 & 6.68E+01 && 2.35E-01 & 4.98E-02 & 1.78E-01 & 2.35E-01 \\
    Chairs    & 1.01E+01  & 1.08E+01 & 1.06E+01 & 1.08E+01 && 3.90E-01 & 1.82E-01 & 3.57E-01 & 3.95E-01 \\
    Cups      & 1.44E+03  & 1.79E+03 & 1.28E+03 & 3.01E+03 && 6.31E-01 & 3.31E-01 & 4.32E-01 & 5.68E-01 \\
    Guitars   & 2.16E+02  & 1.93E+02 & 1.97E+02 & 1.81E+02 && 2.25E-01 & 7.98E-02 & 2.11E-01 & 2.27E-01 \\
    Lamps     & 1.47E+03  & 1.60E+03 & 1.64E+03 & 2.77E+03 && 3.89E-01 & 2.33E-01 & 3.79E-01 & 3.66E-01 \\
    Laptops   & 2.43E+00  & 3.73E+00 & 2.65E+00 & 2.58E+00 && 4.31E-01 & 2.56E-01 & 3.93E-01 & 4.55E-01 \\
    Sofa      & 1.71E+01  & 1.64E+01 & 1.45E+01 & 2.76E+01 && 3.65E-01 & 1.62E-01 & 2.94E-01 & 3.47E-01 \\
    Tables    & 2.79E+00  & 2.96E+00 & 2.44E+00 & 3.69E+00 && 3.82E-01 & 2.59E-01 & 3.20E-01 & 3.53E-01 \\
    \midrule                                                                     
    ModelNet10& 5.77E+00  & 6.89E+00 & 6.03E+00 & 9.19E+00 && 3.47E-01 & 1.90E-01 & 3.36E-01 & 3.67E-01 \\
    ModelNet40& 4.84E+01  & 5.86E+01 & 5.24E+01 & 7.96E+01 && 3.80E-01 & 1.85E-01 & 3.65E-01 & 3.71E-01 \\
 \bottomrule
\end{tabular}}
\label{tb:result}
\end{table}
We first evaluate the performance of trained conditional generator $G_x$ and the inference network $Q$. 
We are interested in whether the learned model can model the distribution of the unseen testing data. Therefore, for
each testing point cloud, we use $Q$ to infer the latent variable $Q(X)$, then use $G_x$
to generate points. We  then compare the distribution between the input point cloud and the generated point clouds. 

There are many criteria based on finite sample estimation can be used for evaluation, such $f$-divergence and IPM.
However, the estimator with finite samples are either biased or with high
variance~\citep{peyre2017computational,wang2009divergence,poczos2012nonparametric, weed2017sharp}. Also, it is impossible to use these estimators with infinitely many samples if they are accessible. 

\begin{wrapfigure}{r}{.2\textwidth}  
\vspace{-3.5mm}
\includegraphics[width=\linewidth]{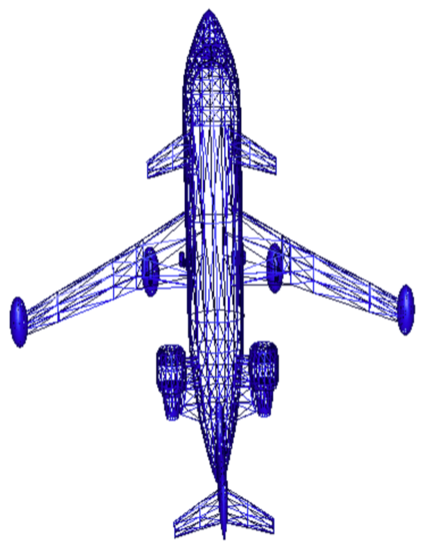}
\caption{Sample mesh from ModelNet40}
\vspace{-7mm}
\label{fig:mesh}
\end{wrapfigure}
For ModelNet40, the meshes of each object are available. 
In many statistically guaranteed distance estimates, the adopted statistics are commonly based on distance between nearest neighbors~\citep{wang2009divergence,poczos2012nonparametric}. 
Therefore, we propose to measure the performance with the following criteria. 
Given a point cloud $\{x_i\}_{i=1}^n$ and a mesh, which is a collection of faces $\{F_j\}_{j=1}^m$, we measure
the \emph{distance to face (D2F)} as 
\[
     D2F\left( \{x_i\}_{i=1}^n, \{F_j\}_{j=1}^m \right) = \frac{1}{n} \sum_{i=1}^n \min_j \Dcal(x_i, F_j),
\]
where $\Dcal(x_i, F_j)$ is the Euclidean distance from $x_i$ to the face $F_j$.
This distance is similar to Chamfer distance, which is commonly used for measuring images and point clouds~\citep{achlioptas2017learning,fan2017point}, with infinitely samples from true distributions (meshes). 

However, the algorithm can have low or zero D2F by only focusing a small portion of the point clouds (mode collapse).
Therefore, we are also interested in 
whether the generated points recover enough supports of the distribution. 
We compute the \emph{Coverage} ratio as follows. For each points, we find the its nearest face, we then treat this face
is covered\footnote{We should do thresholding to ignore outlier points. In our experiments, we observe that without
excluding outliers does not change conclusion for comparison.}. We then compute the ratio of number of faces of a mesh is covered.
A sampled mesh is showed in Figure~\ref{fig:mesh}, 
where the details have more faces (non-uniform). Thus, it is difficult to get high coverage for AAE or PC-GAN trained by
limited number of sampled points. However, the coverage ratio, on the other hand, serve as an indicator about how
much details the model recovers.

The results are reported in Table~\ref{tb:result}. We compare four different algorithm, AAE and 
PC-GAN with three objectives, including 
upper bound $W_U$ ( $\epsilon$ approximated Wasserstein distance), lower bound $W_L$ (GAN with $L^2$ ball constraints and weight clipping), 
and the sandwiching loss $W_S$ as discussed in Section~\ref{sec:sandwiching}, 
The study with $W_U$ and $W_L$ also serves as the ablation test of the proposed sandwiching loss $W_s$.

\paragraph{Comparison between Upper bound, Lower bound and Sandwiching}
Since $W_U$ directly optimizes distance between training and generated point clouds, 
$W_U$ usually results in smaller D2F than $W_L$ in Table~\ref{tb:result}. 
One the other hand, 
although $W_L$ only recovers lower bound estimate of Wasserstein distance,
its discriminator is known to focus on learning support of the distribution~\citep{bengio18support}, which results in
better coverage (support) than $W_U$.

\begin{figure}[t]
\centering
\begin{subfigure}[b]{.9\linewidth}
\includegraphics[width=0.19\linewidth,trim={40mm 50mm 50mm 40mm},clip]{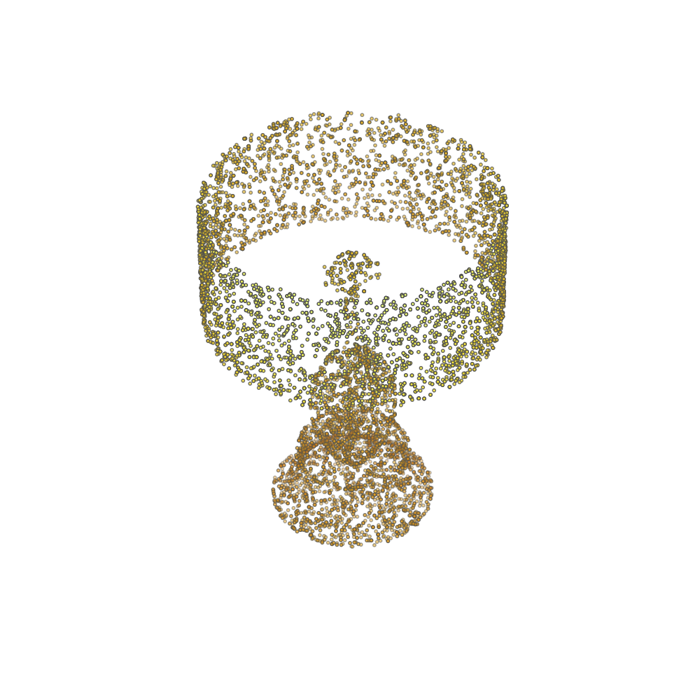}
\includegraphics[width=0.19\linewidth,trim={50mm 50mm 50mm 50mm},clip]{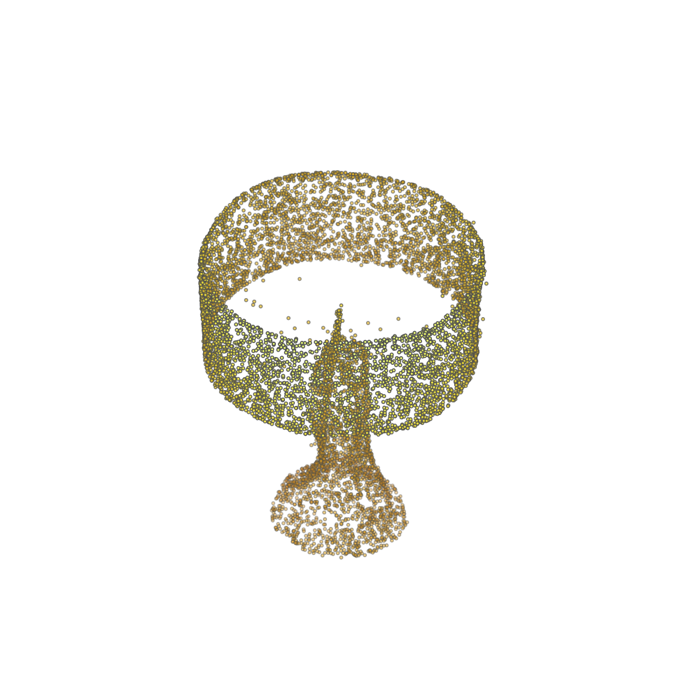}
\includegraphics[width=0.19\linewidth,trim={50mm 50mm 50mm 50mm},clip]{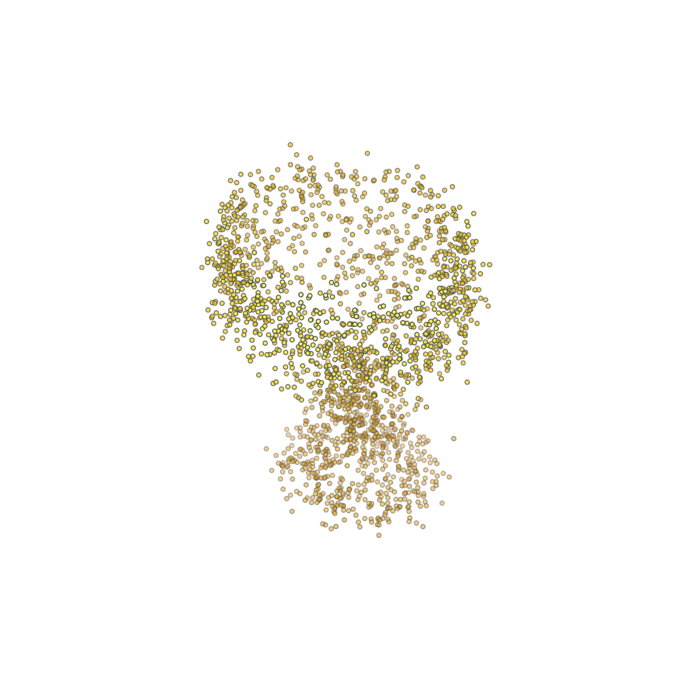}
\includegraphics[width=0.19\linewidth,trim={50mm 50mm 50mm 50mm},clip]{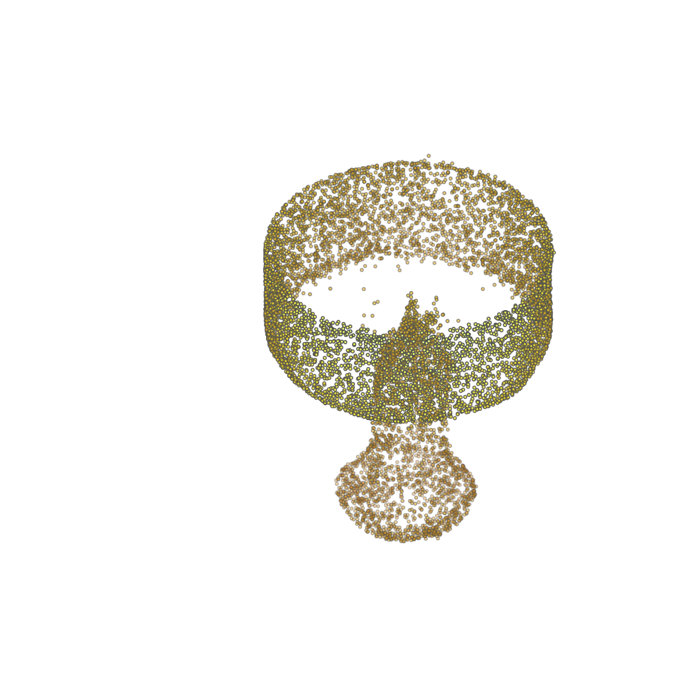}
\includegraphics[width=0.19\linewidth,trim={50mm 50mm 50mm 50mm},clip]{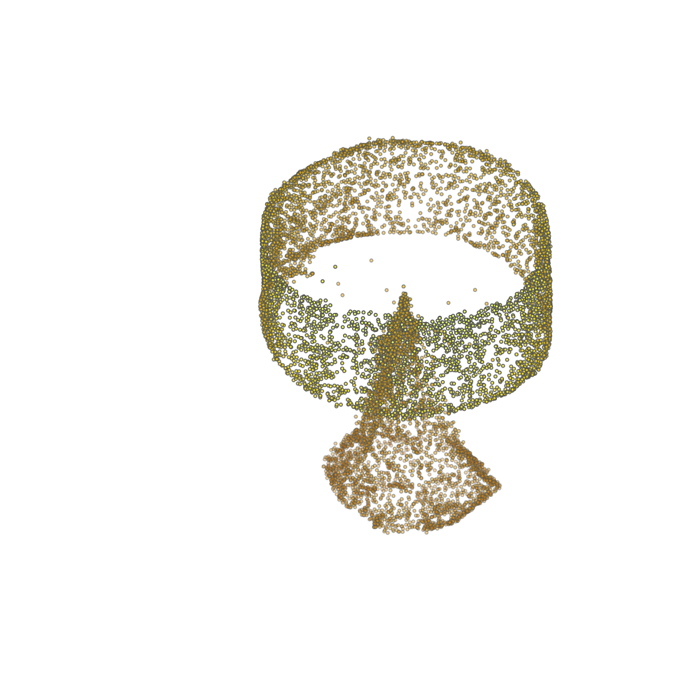}
\end{subfigure}
\begin{subfigure}[b]{.9\linewidth}
\includegraphics[width=0.19\linewidth,trim={50mm 50mm 50mm 50mm},clip]{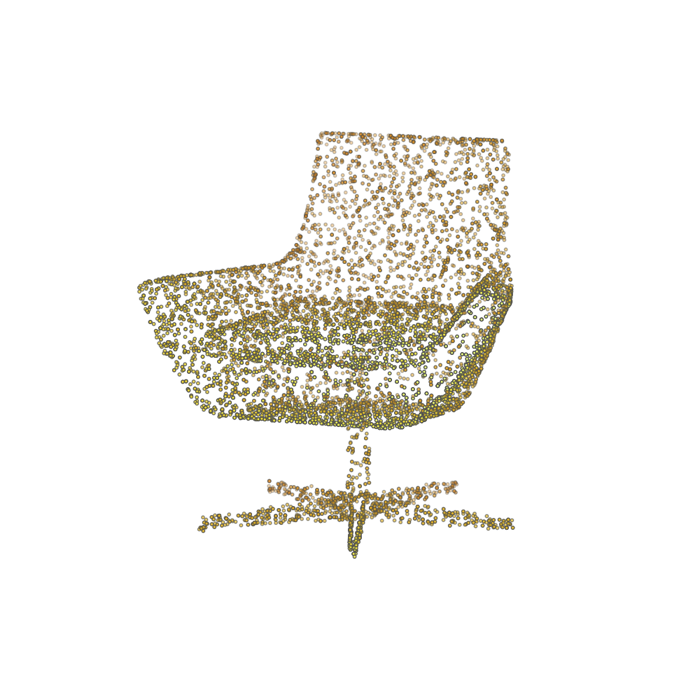}
\includegraphics[width=0.19\linewidth,trim={50mm 50mm 50mm 50mm},clip]{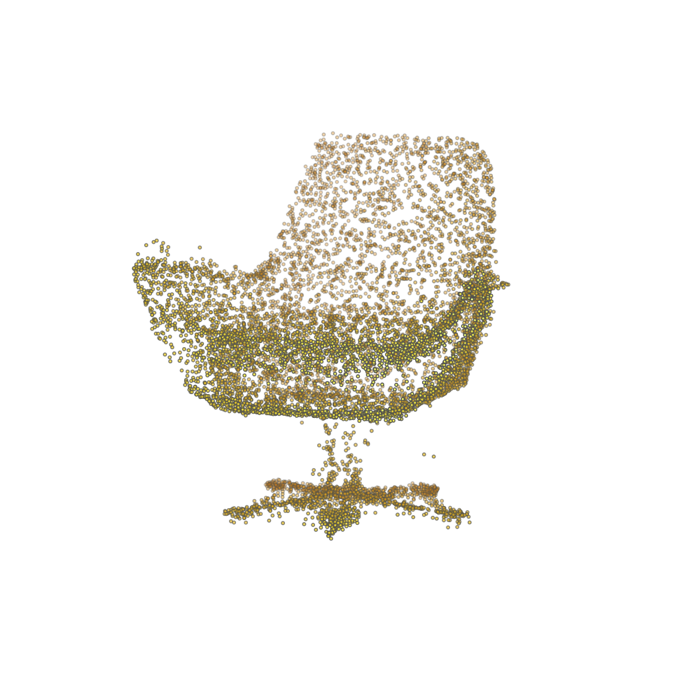}
\includegraphics[width=0.19\linewidth,trim={50mm 50mm 50mm 50mm},clip]{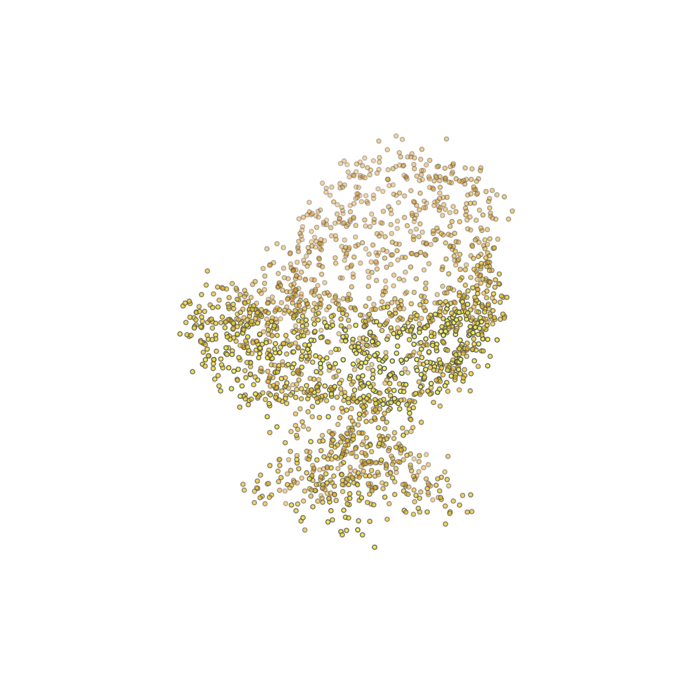}
\includegraphics[width=0.19\linewidth,trim={50mm 50mm 50mm 50mm},clip]{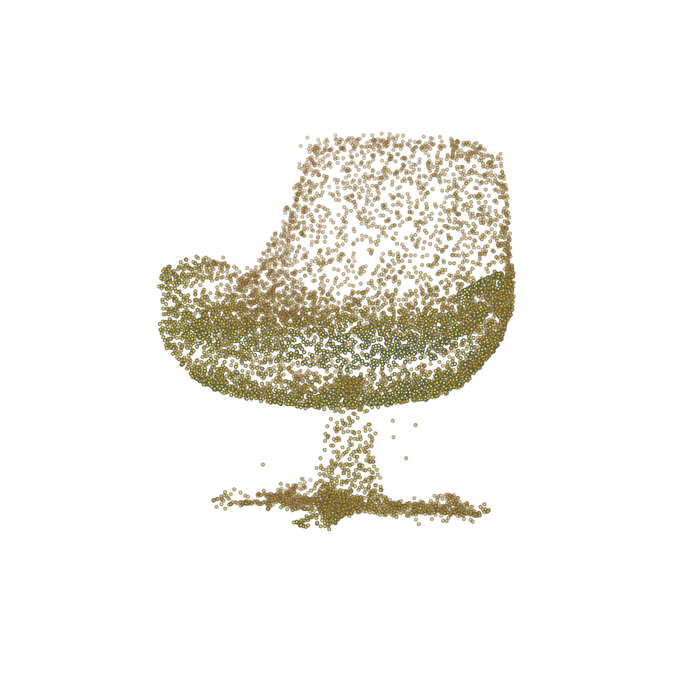}
\includegraphics[width=0.19\linewidth,trim={50mm 50mm 50mm 50mm},clip]{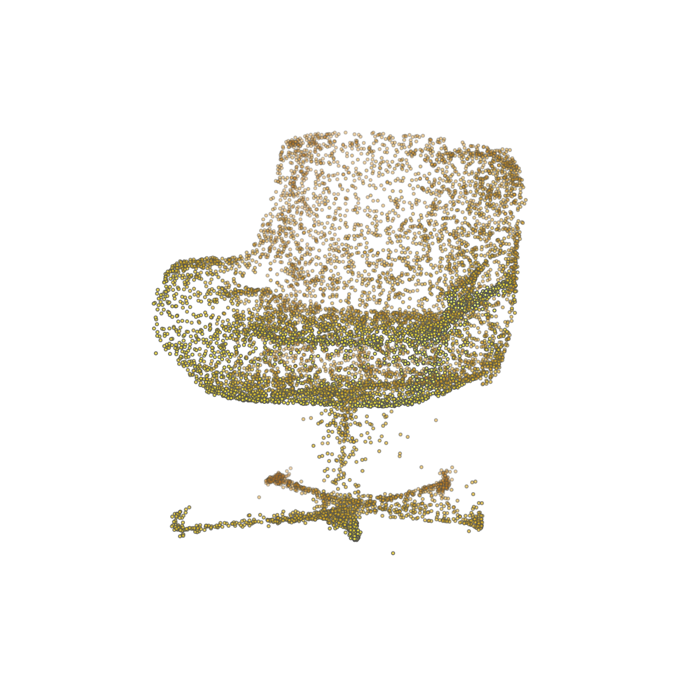}
\end{subfigure}
\begin{subfigure}[b]{.9\linewidth}
\includegraphics[width=0.19\linewidth,trim={80mm 100mm 50mm 70mm},clip]{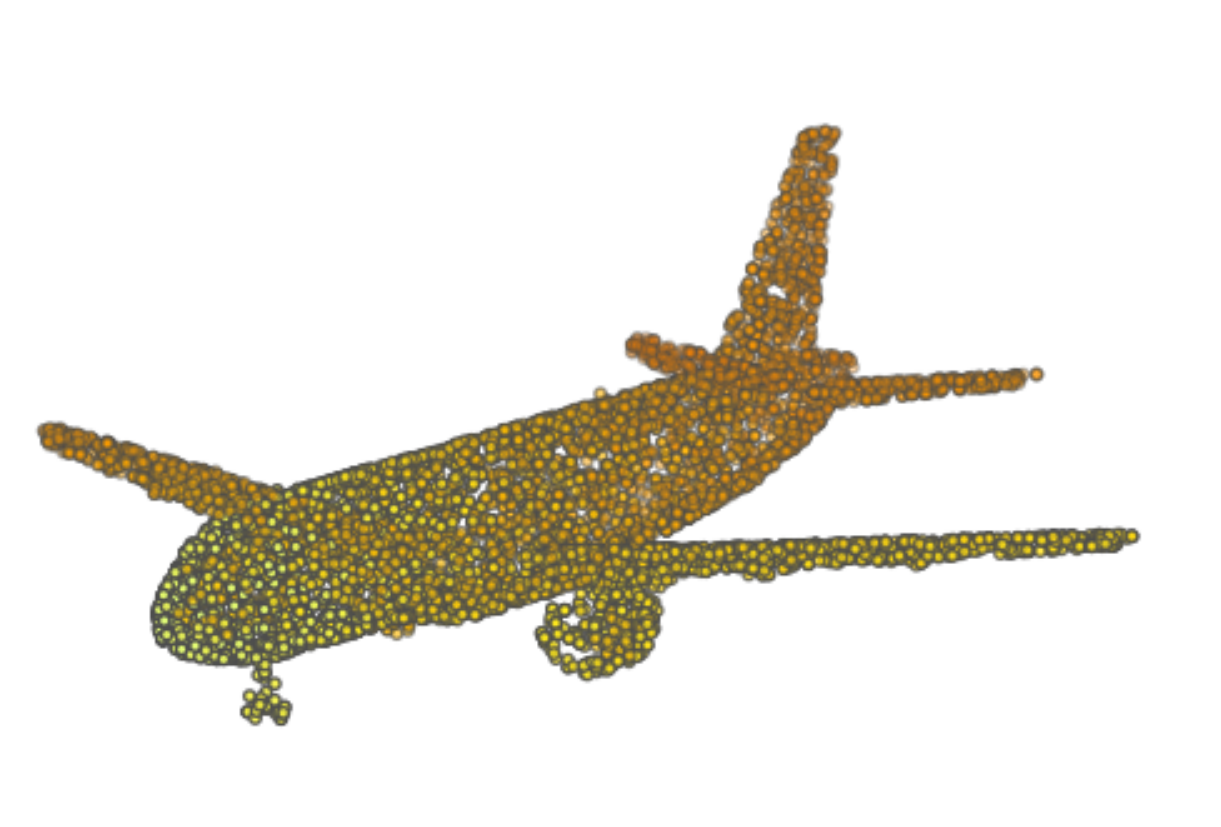}
\includegraphics[width=0.19\linewidth,trim={80mm 100mm 50mm 70mm},clip]{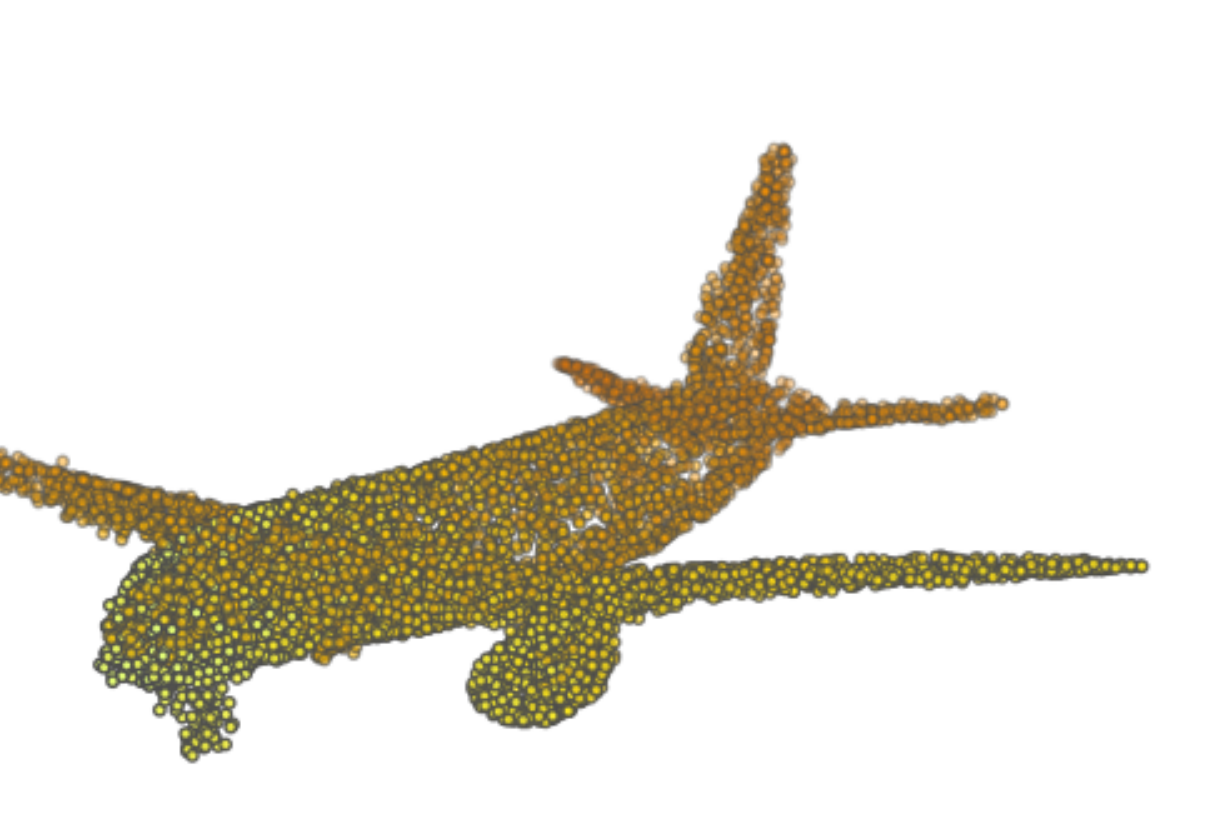}
\includegraphics[width=0.19\linewidth,trim={80mm 100mm 50mm 70mm},clip]{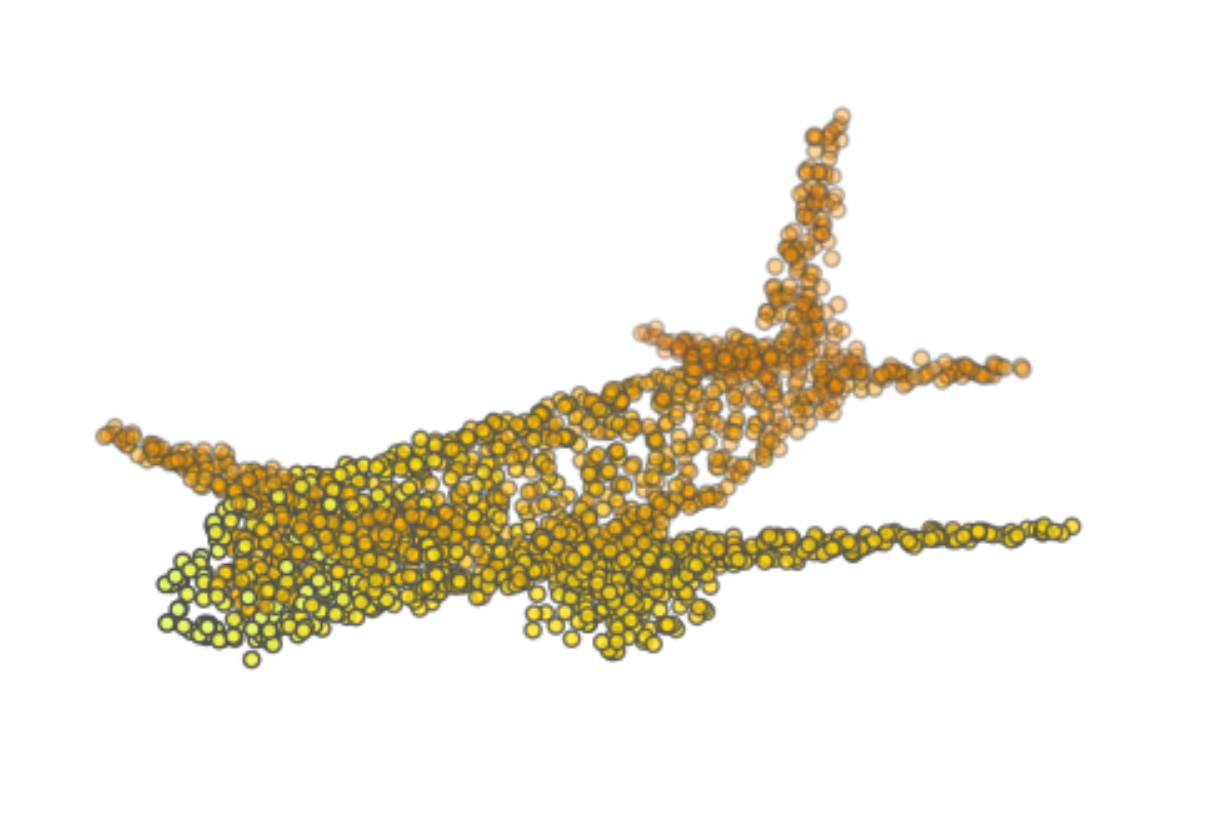}
\includegraphics[width=0.19\linewidth,trim={80mm 100mm 50mm 70mm},clip]{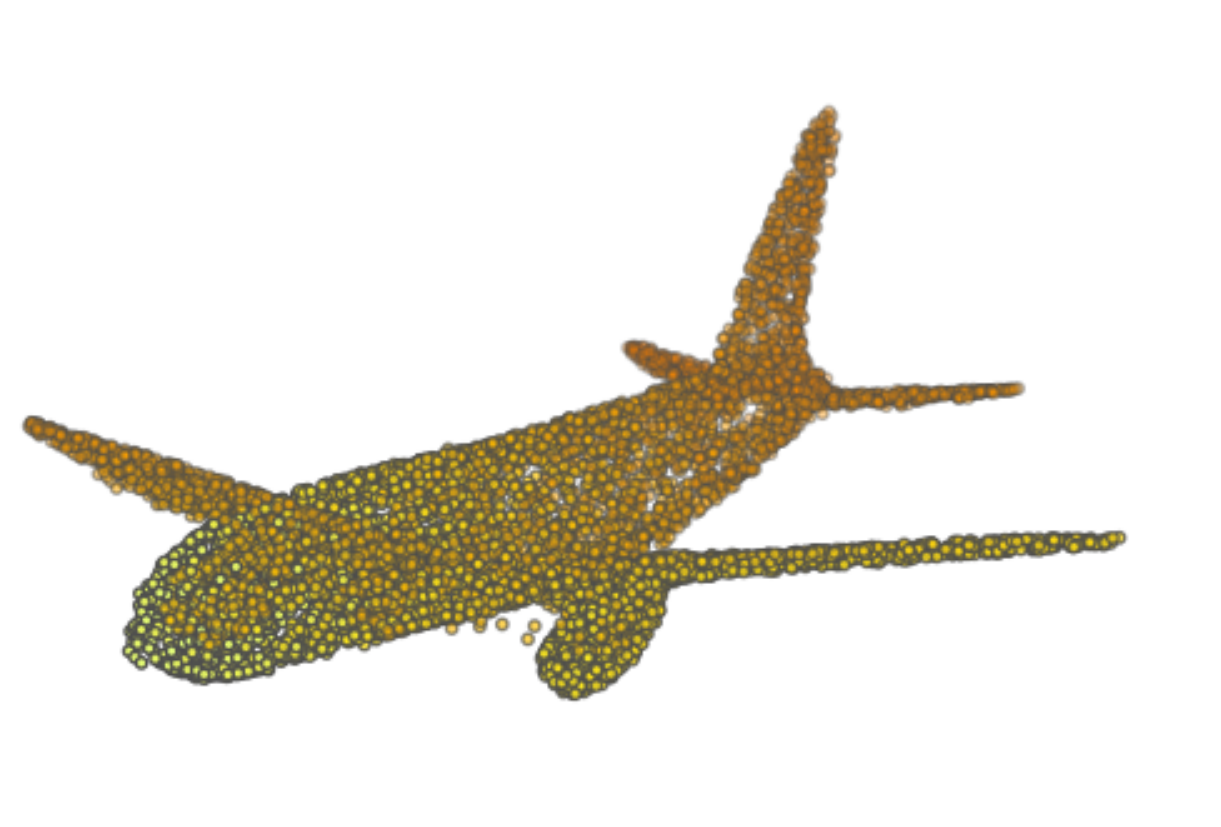}
\includegraphics[width=0.19\linewidth,trim={80mm 100mm 50mm 70mm},clip]{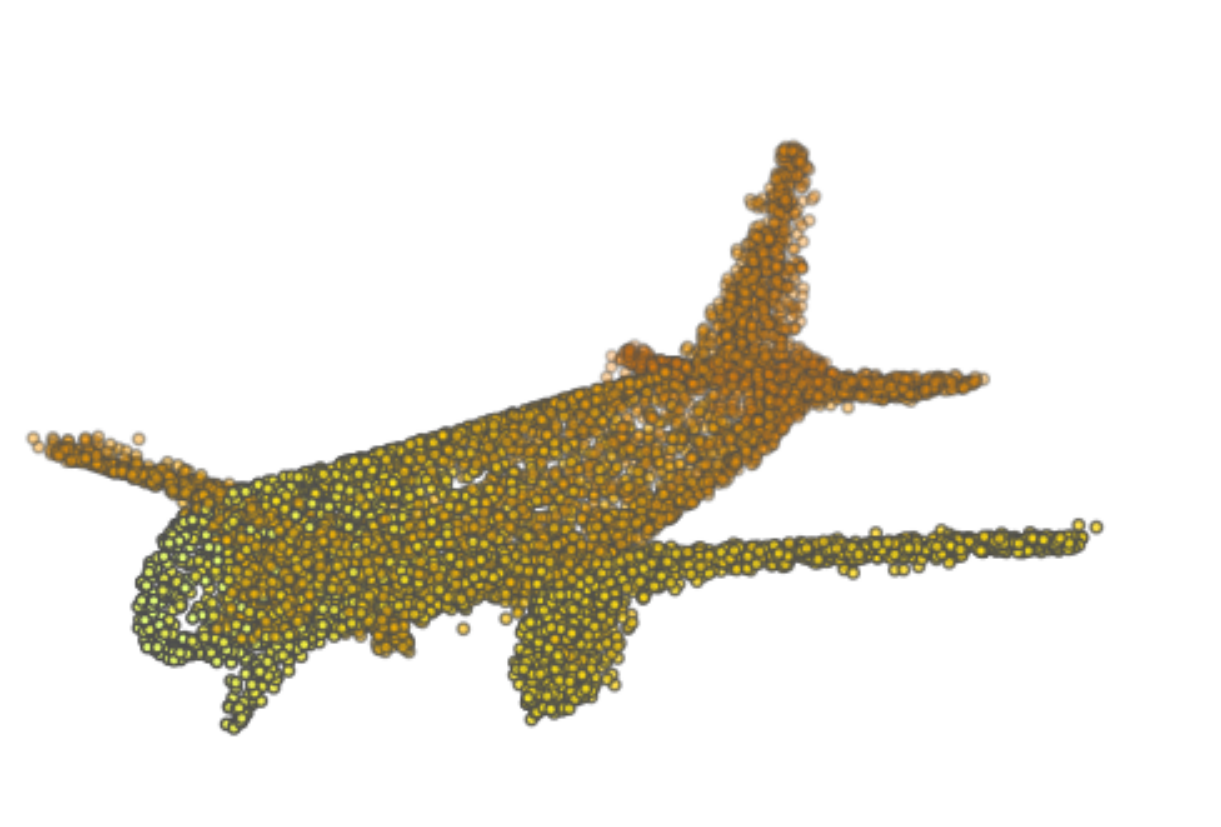}
\end{subfigure}

{\small \hfill Data \hfill PC-GAN $(W_S)$ \hfill AAE \hfill PC-GAN $(W_U)$ \hfill PC-GAN $(W_L)$ \hfill}
\caption{The reconstruction on test objects from seen categories. For each object, from left to right is training data, AAE, and PC-GAN. PC-GAN is better in capturing fine details like wheels of aeroplane or proper chair legs.}
\label{fig:reconstruction}
\end{figure}

Theoretically, the proposed sandwiching $W_s$ results in a tighter Wasserstein distance estimation than $W_U$ and $W_L$
(Lemma~\ref{lem:sandwiching}). Based on above discussion, 
it can also be understood as balancing both D2F 
and coverage by combining both $W_U$ and $W_L$ to get a desirable middle ground. 
Empirically, we even observe that $W_s$ results in better coverage than $W_L$, and competitive D2F with
$W_U$.
The intuitive explanation is that some discriminative tasks are \emph{off} to $W_U$ objective, so the discriminator can focus more on
learning distribution supports.
We argue that this difference is crucial for capturing the object details.
Some reconstructed point clouds of 
testing data are shown in Figure~\ref{fig:reconstruction}.
For aeroplane examples, $W_U$ are failed to capture aeroplane tires and $W_s$ has better tire than $W_L$.
For Chair example, $W_s$ recovers better legs than $W_U$ and better 
seat cushion than $W_L$. 
Lastly, we highlight $W_s$ outperforms others more significantly when training data is larger (ModelNet10 and
ModelNet40) in Table~\ref{tb:result}.

\paragraph{Comparison between PC-GAN and AAE}
In most of cases, PC-GAN with $W_s$ has lower D2F in Table~\ref{tb:result} with
less number of parameters aforementioned. Similar to the argument in Section~\ref{sec:exp_syn}, although AAE use
larger networks, the decoder wastes parameters for nearby points. AAE only outperforms PC-GAN ($W_s$) in Guitar
and Sofa in terms of D2F, since the variety of these two classes are low. It is easier for MLP to learn
the shared template (basis) of the point clouds. 
On the other hand, due to the limitation of the fixed
number of output points and Chamfer distance objective, AAE has worse coverage than PC-GAN, 
It can be supported by Figure~\ref{fig:reconstruction}, where AAE is also failed to recover aeroplane tire.

\subsection{Hierarchical Sampling}
\label{sec:hierarchy}
\begin{figure}[h]
\centering
\includegraphics[width=0.09\linewidth,trim={50mm 50mm 50mm 50mm},clip]{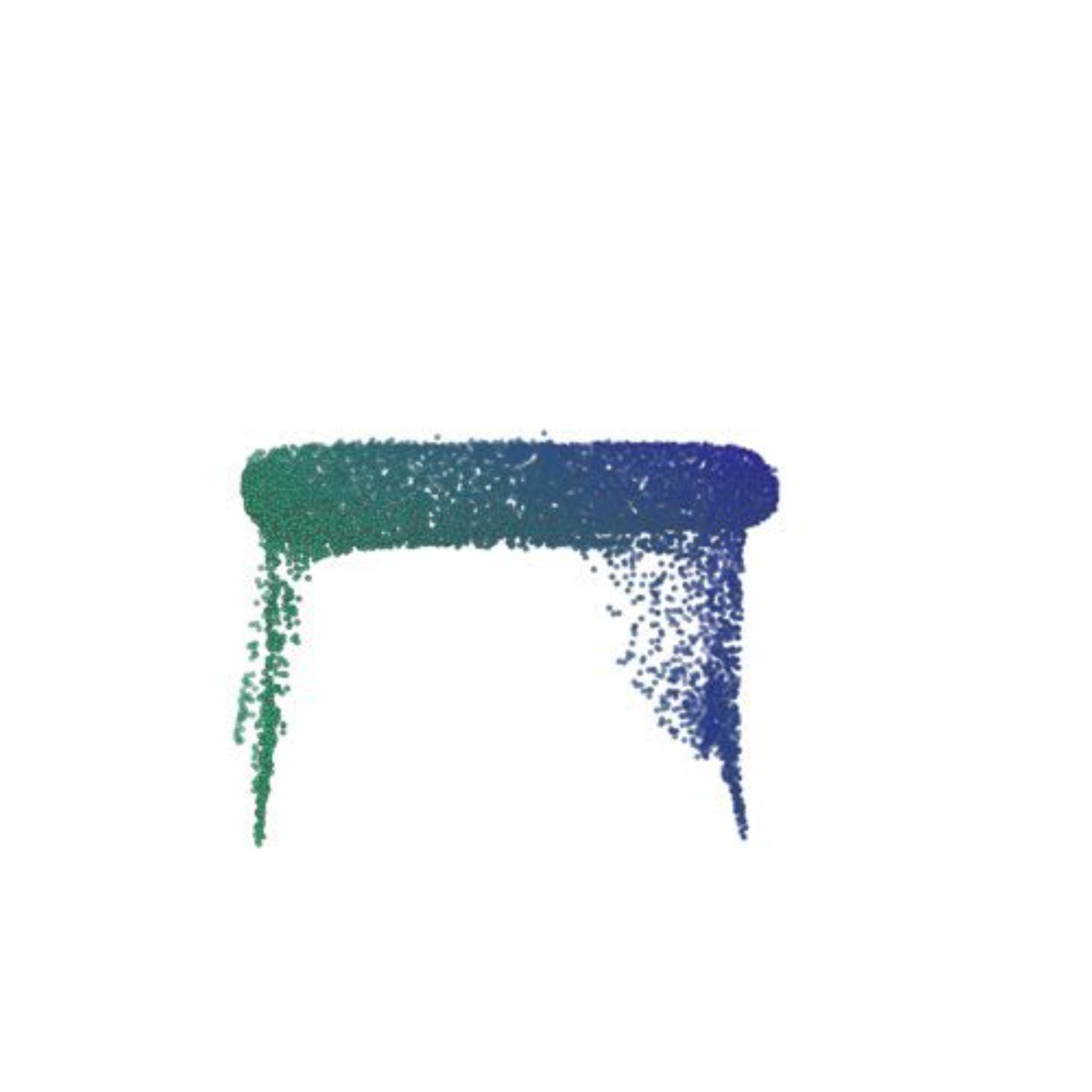}
\includegraphics[width=0.09\linewidth,trim={50mm 50mm 50mm 50mm},clip]{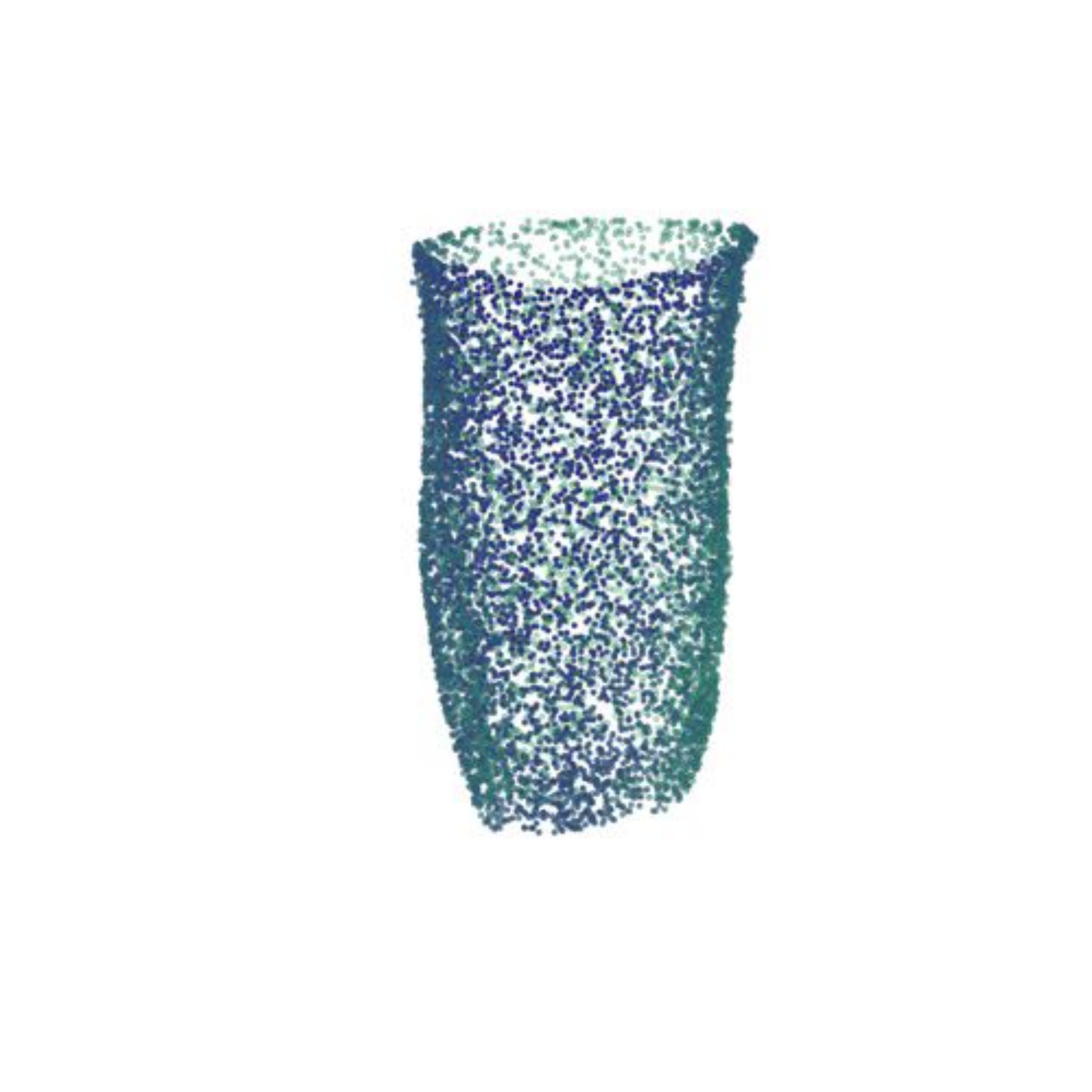}
\includegraphics[width=0.09\linewidth,trim={50mm 50mm 50mm 50mm},clip]{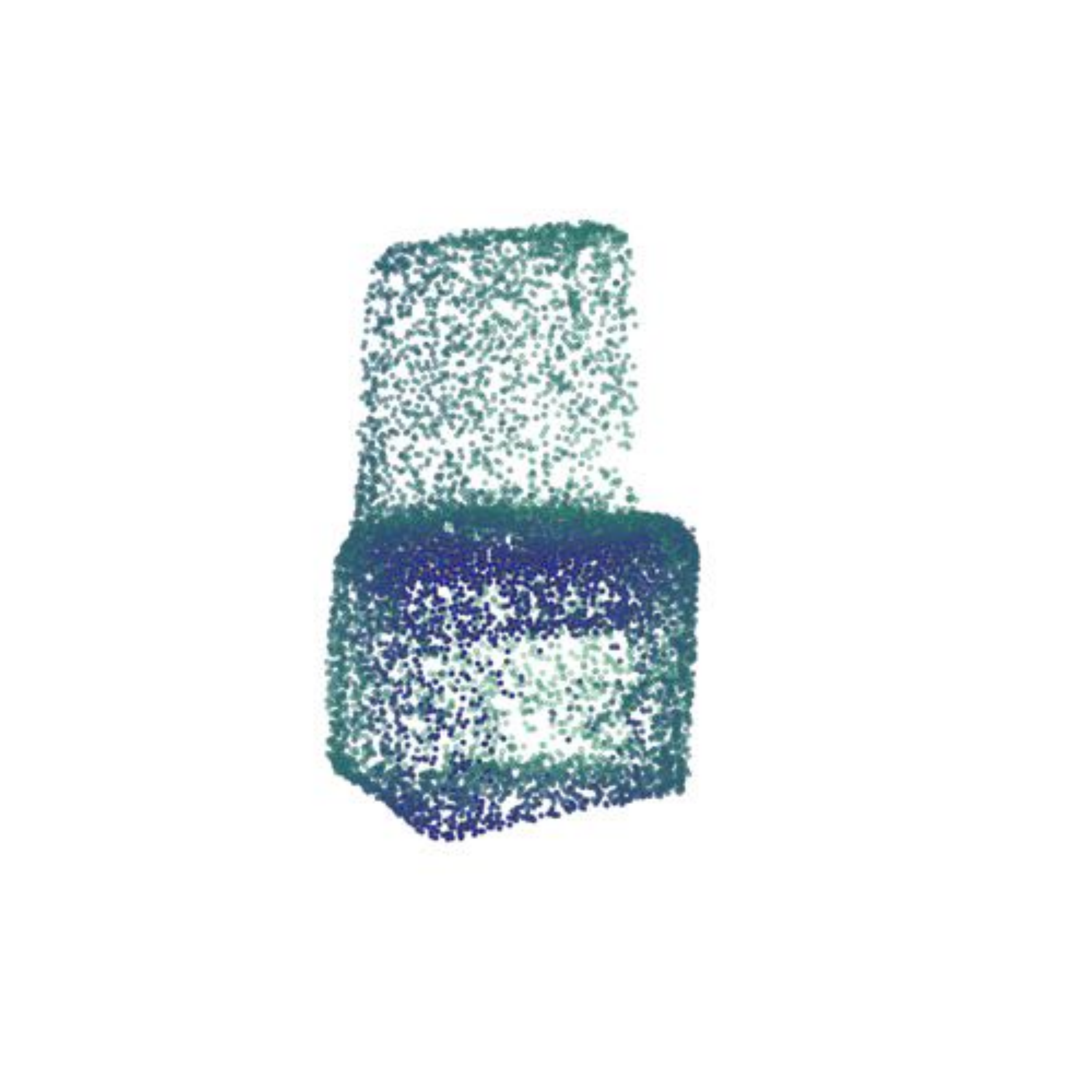}
\includegraphics[width=0.09\linewidth,trim={50mm 50mm 50mm 50mm},clip]{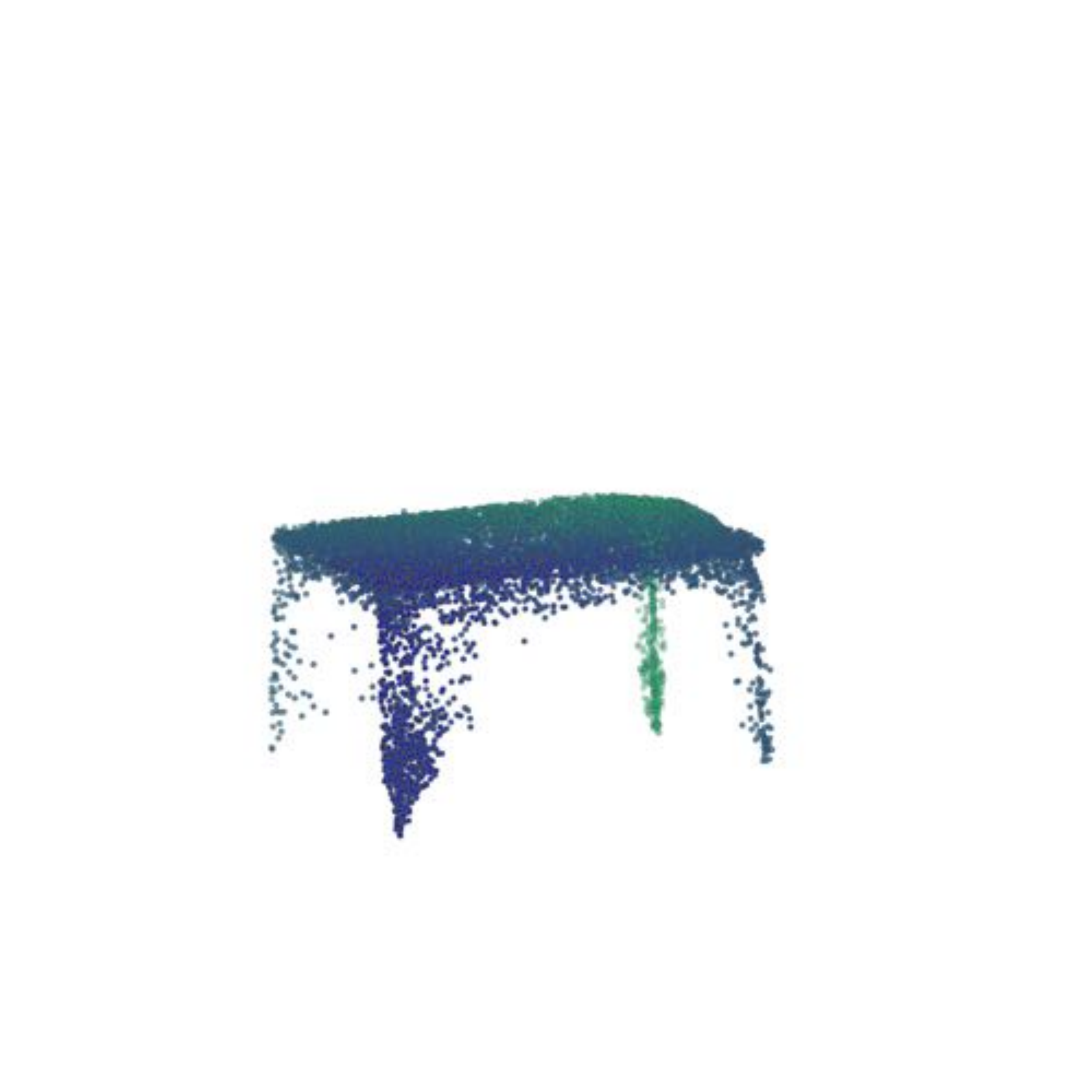}
\includegraphics[width=0.09\linewidth,trim={80mm 100mm 80mm 100mm},clip]{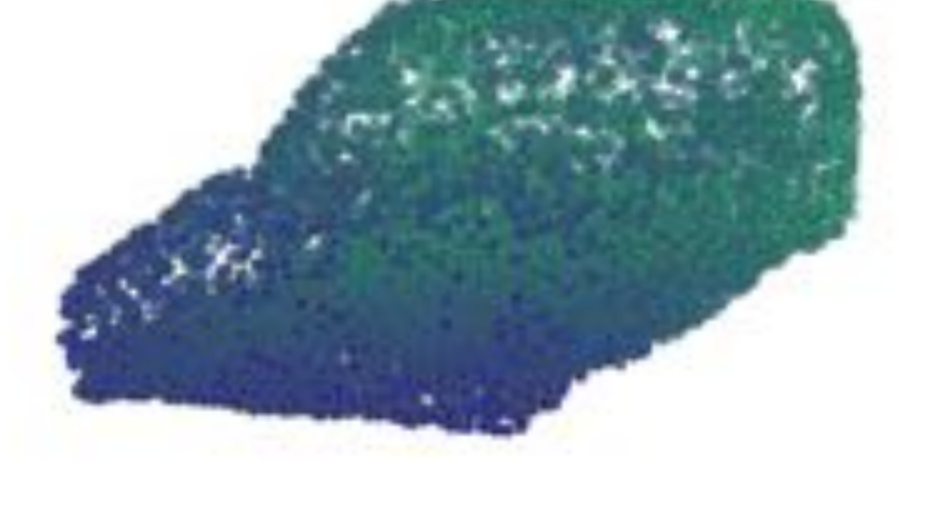}
\includegraphics[width=0.09\linewidth,trim={50mm 50mm 50mm 50mm},clip]{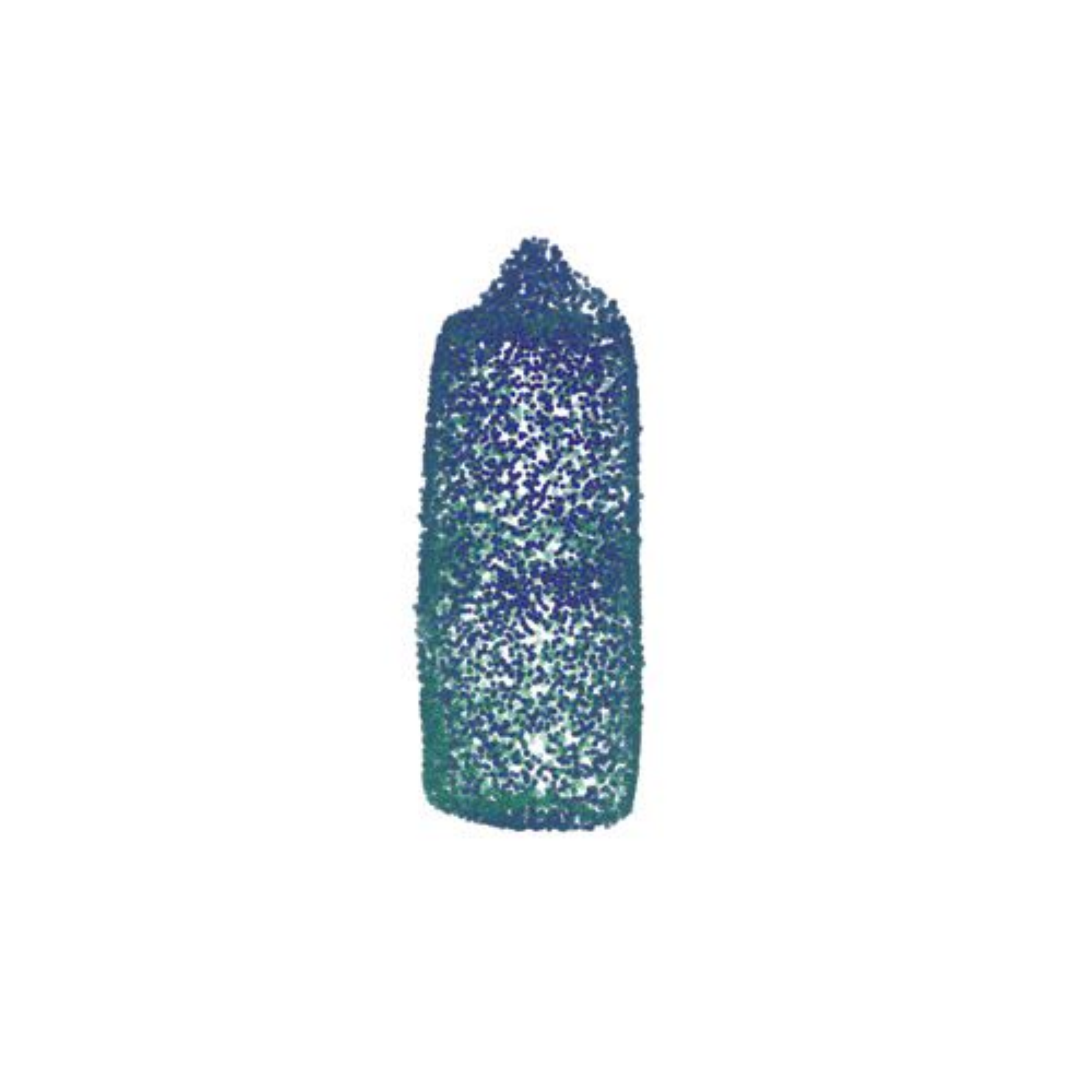}
\includegraphics[width=0.09\linewidth,trim={50mm 50mm 50mm 50mm},clip]{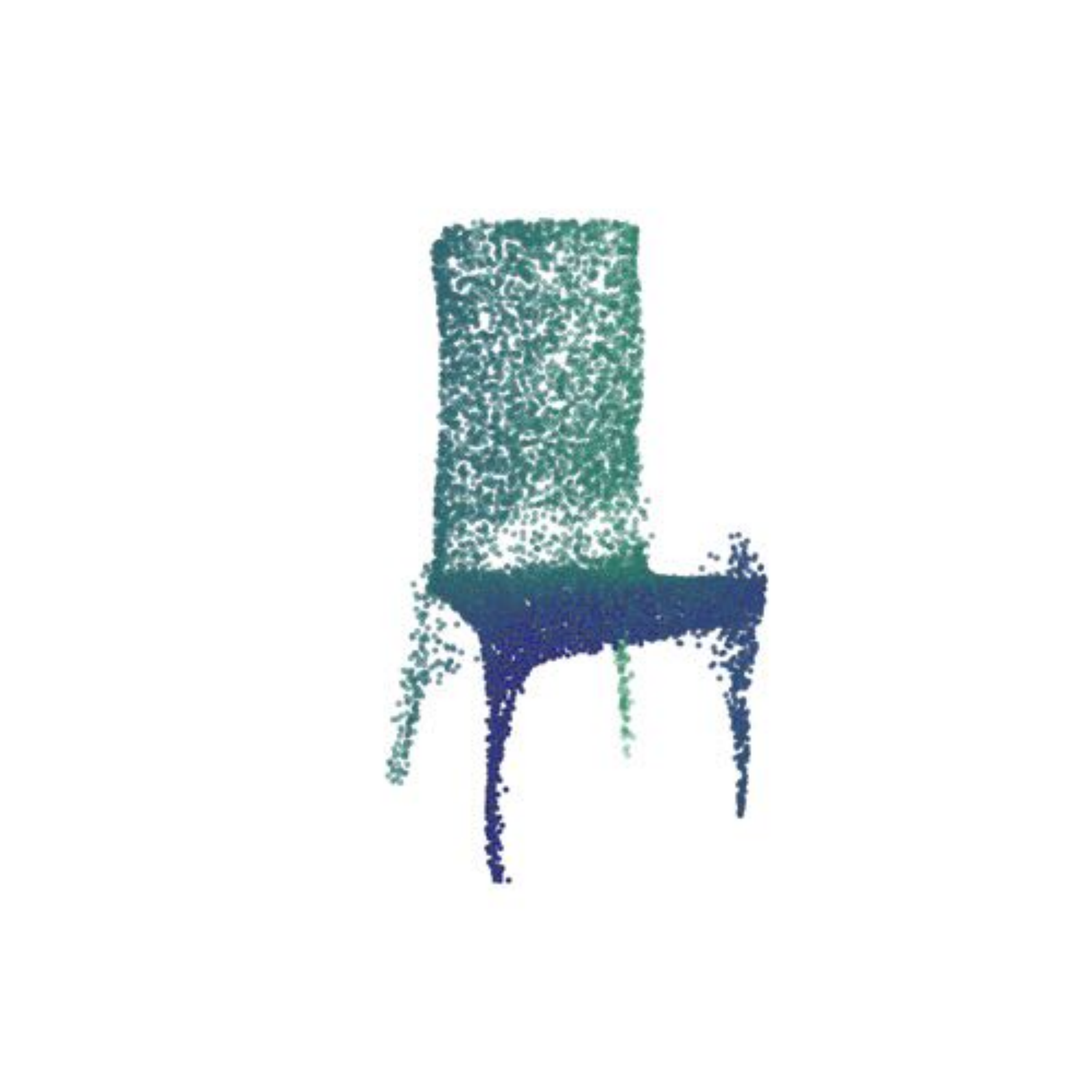}
\includegraphics[width=0.09\linewidth,trim={50mm 50mm 50mm 50mm},clip]{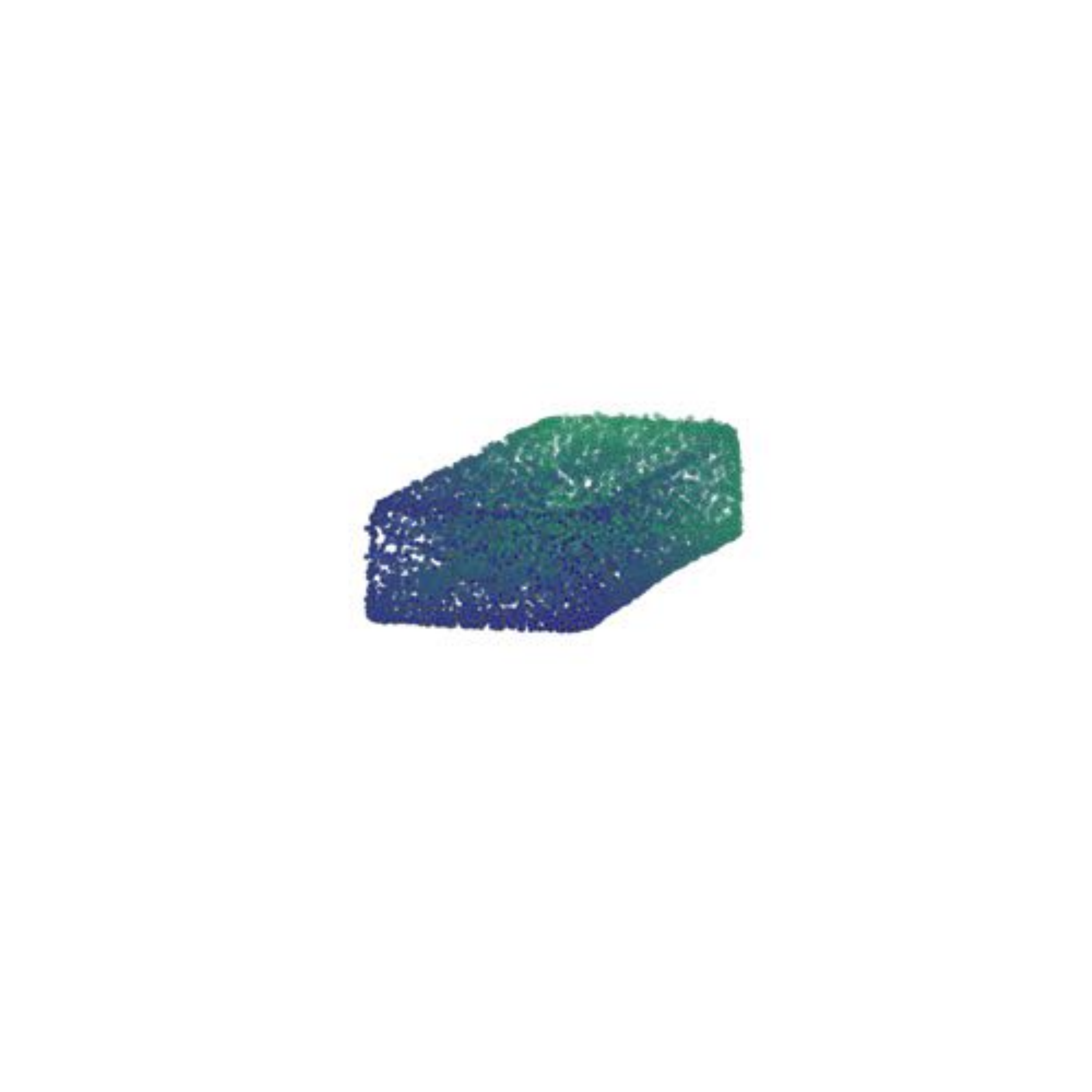}
\includegraphics[width=0.09\linewidth,trim={50mm 50mm 50mm 50mm},clip]{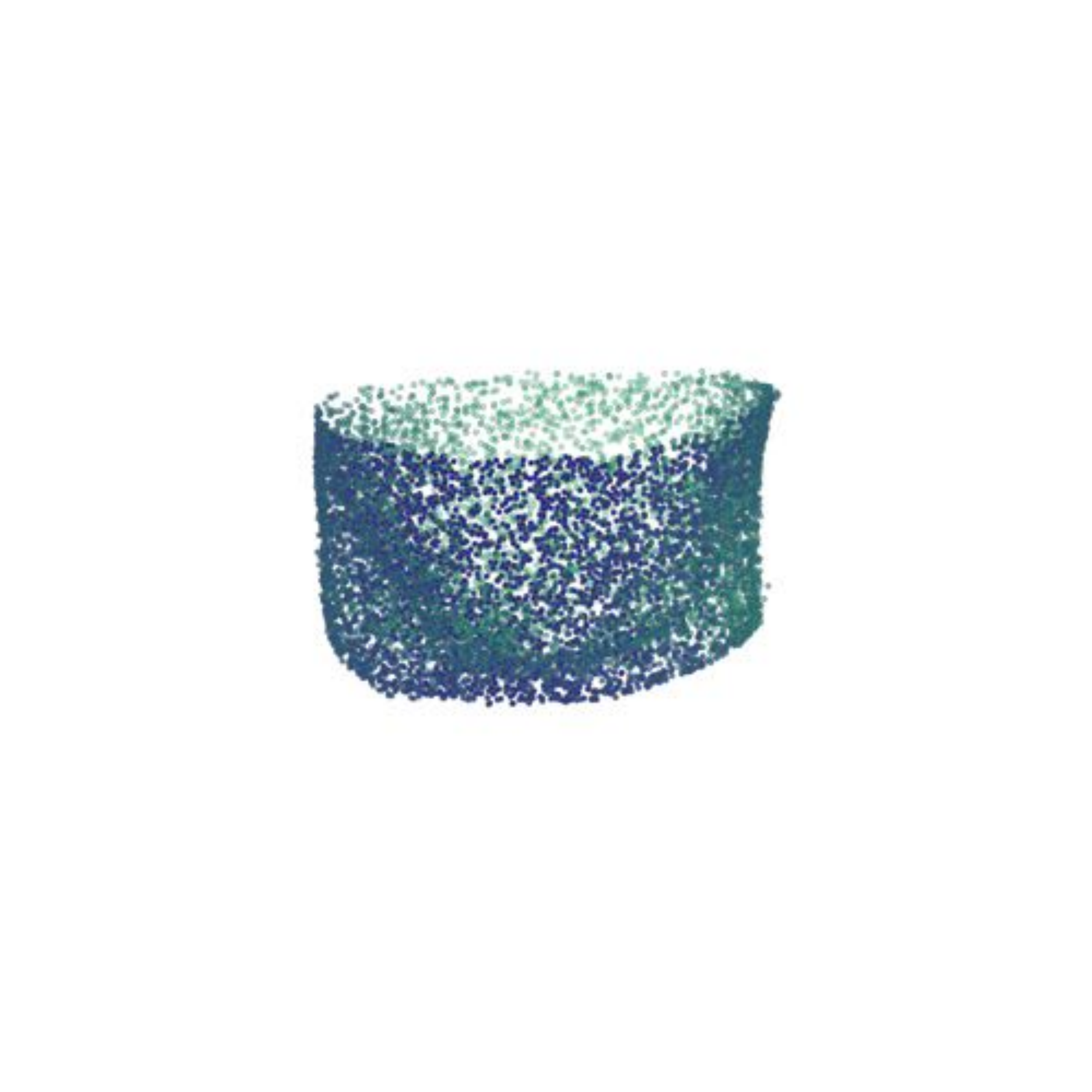}
\includegraphics[width=0.09\linewidth,trim={50mm 50mm 50mm 50mm},clip]{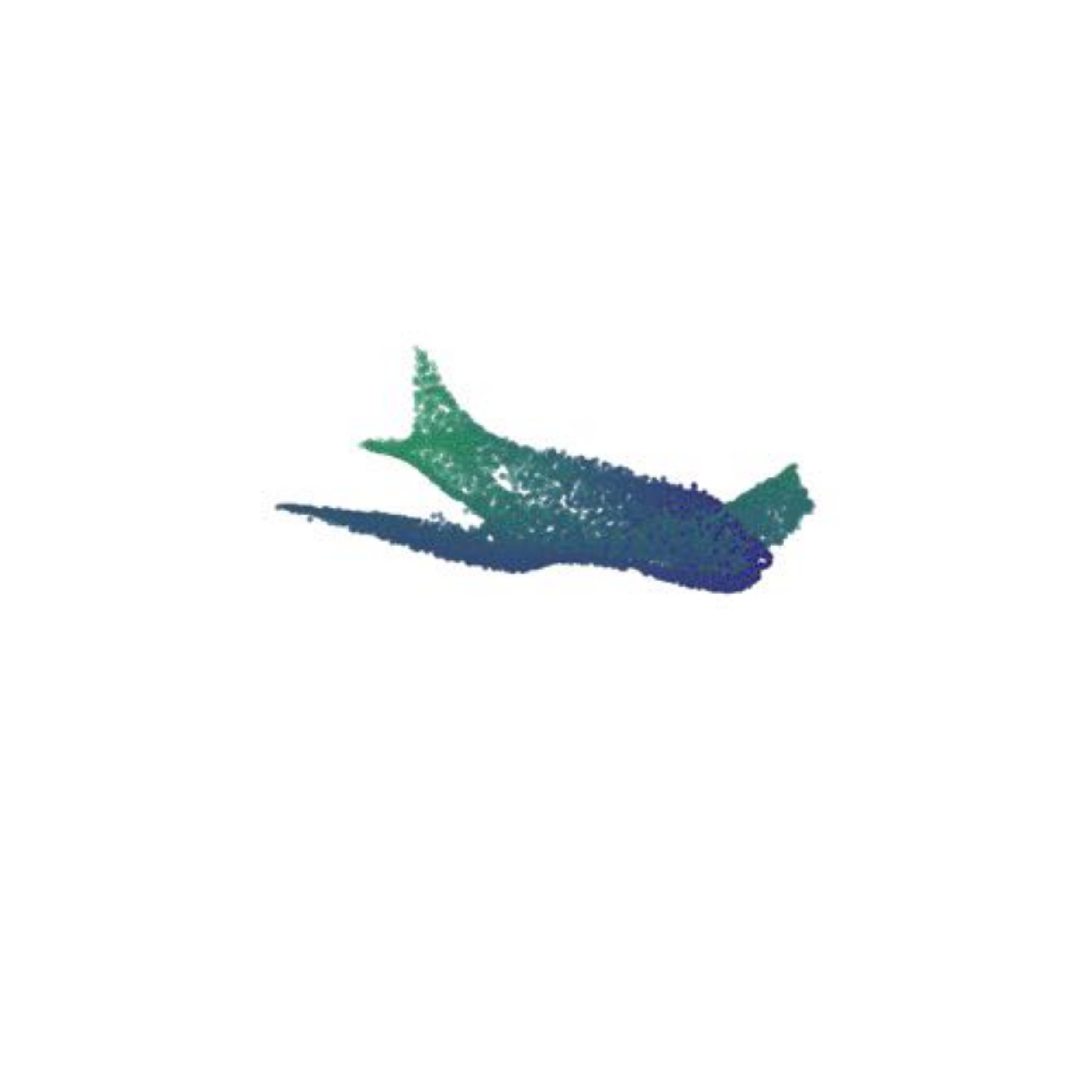}
\includegraphics[width=0.09\linewidth,trim={50mm 50mm 50mm 50mm},clip]{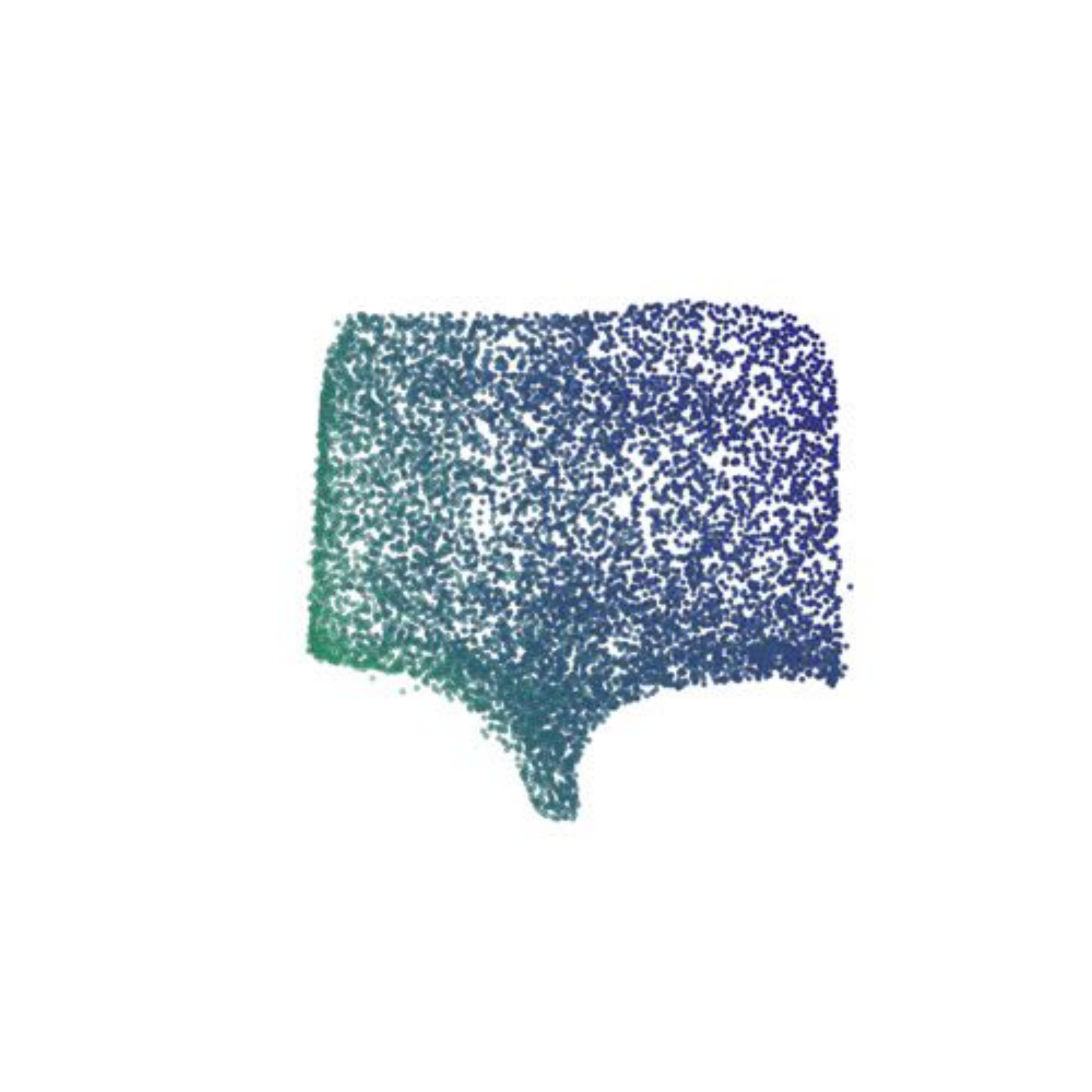}
\includegraphics[width=0.09\linewidth,trim={30mm 60mm 80mm 50mm},clip]{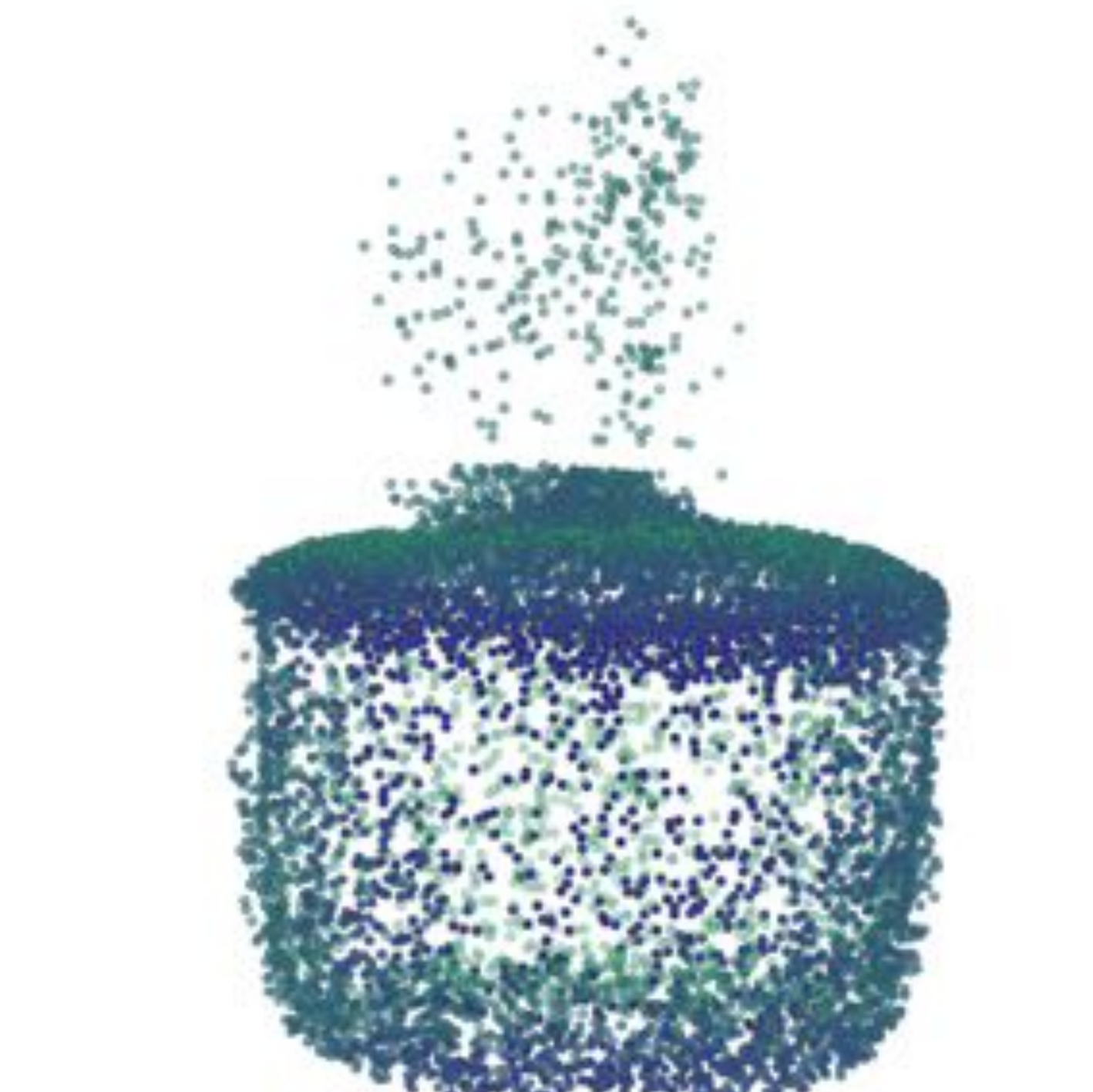}
\includegraphics[width=0.09\linewidth,trim={50mm 50mm 50mm 50mm},clip]{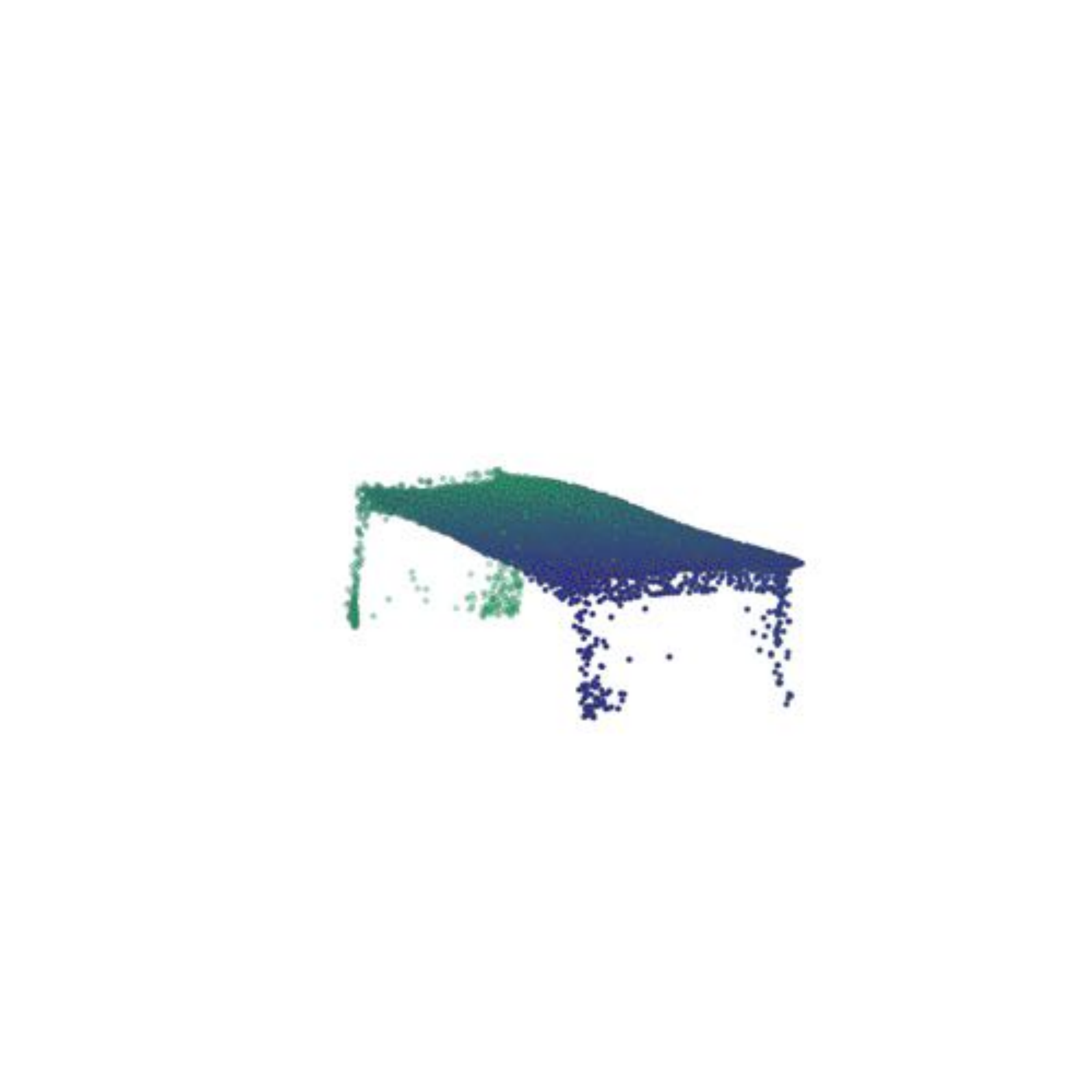}
\includegraphics[width=0.09\linewidth,trim={50mm 50mm 50mm 50mm},clip]{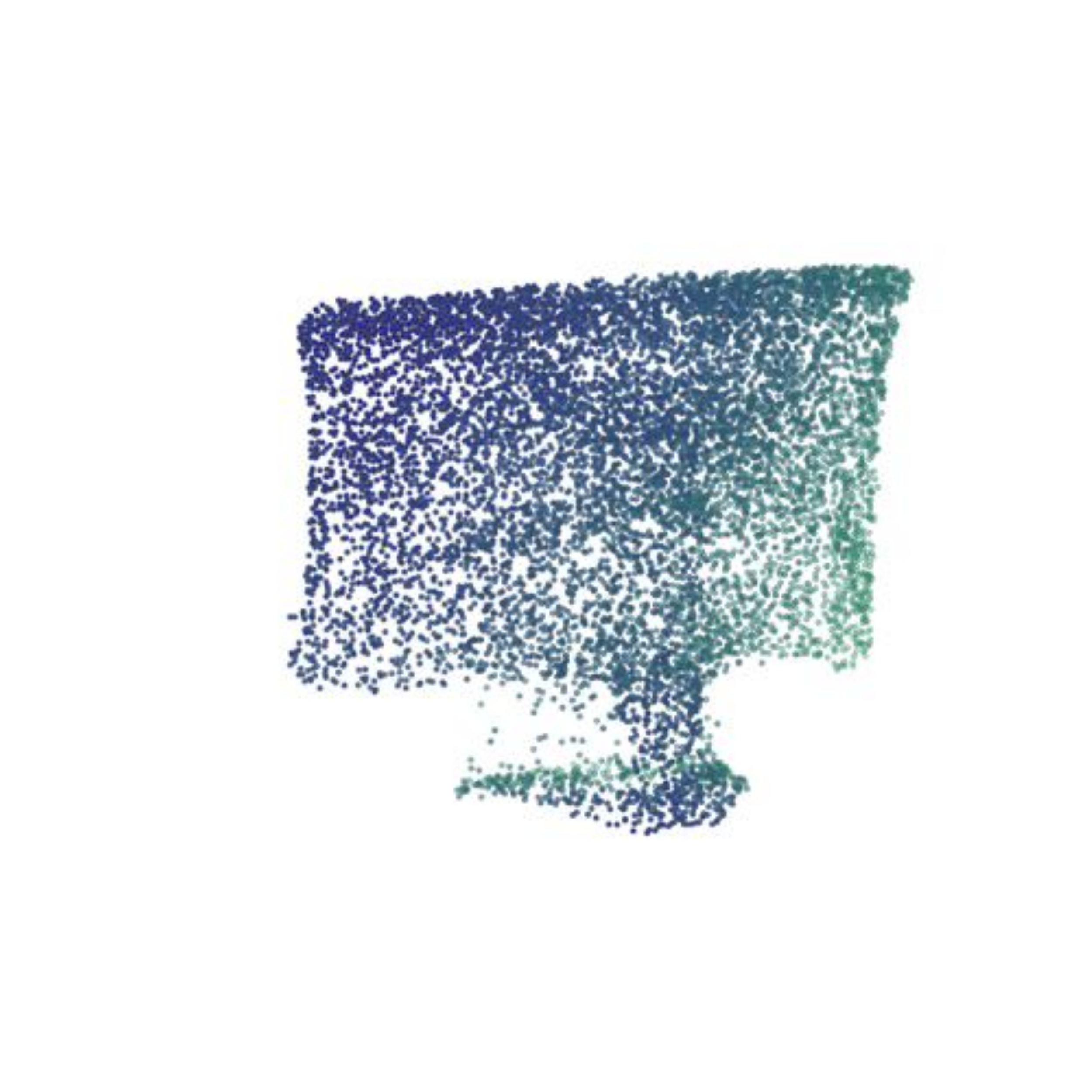}
\includegraphics[width=0.09\linewidth,trim={50mm 50mm 50mm 50mm},clip]{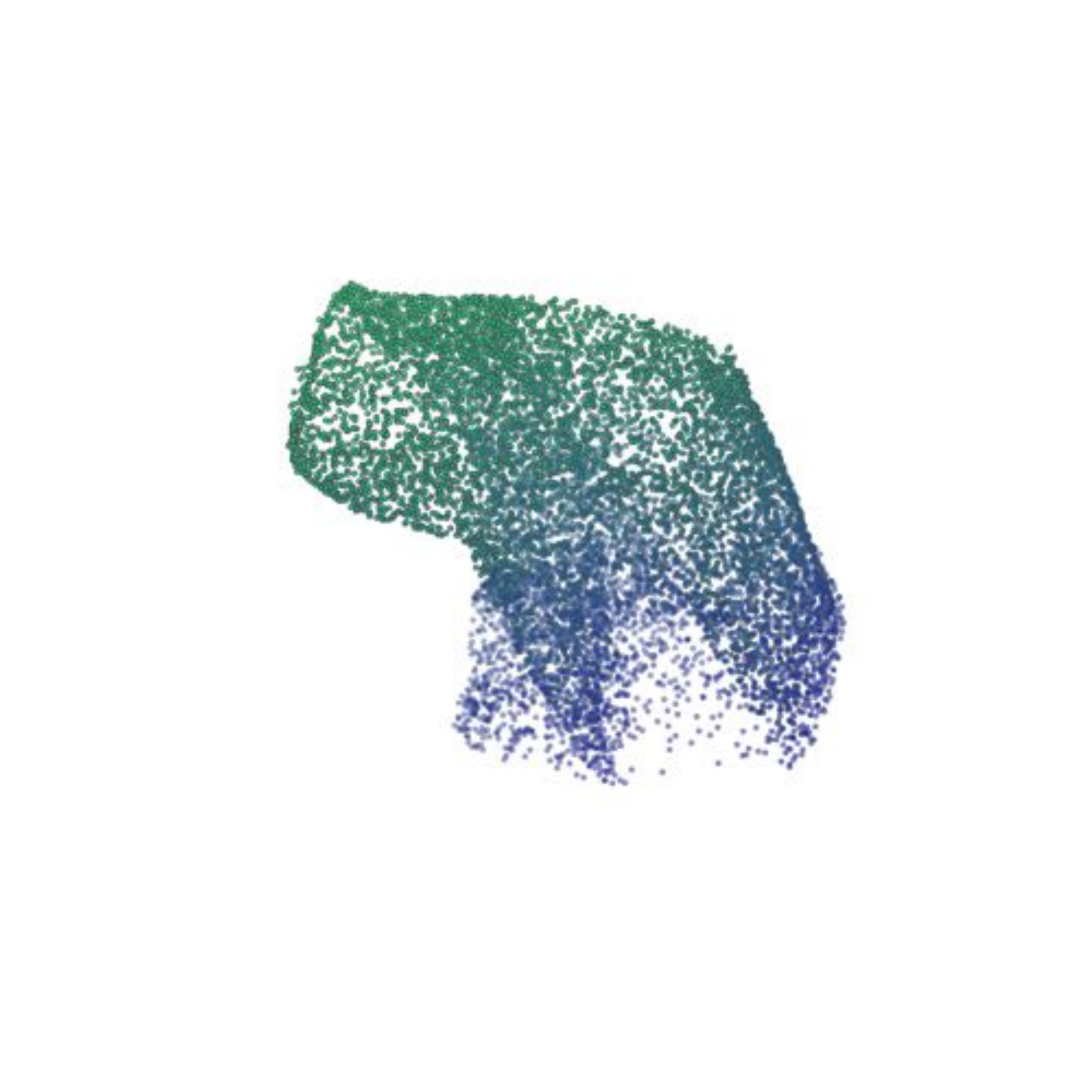}
\includegraphics[width=0.09\linewidth,trim={50mm 50mm 50mm 50mm},clip]{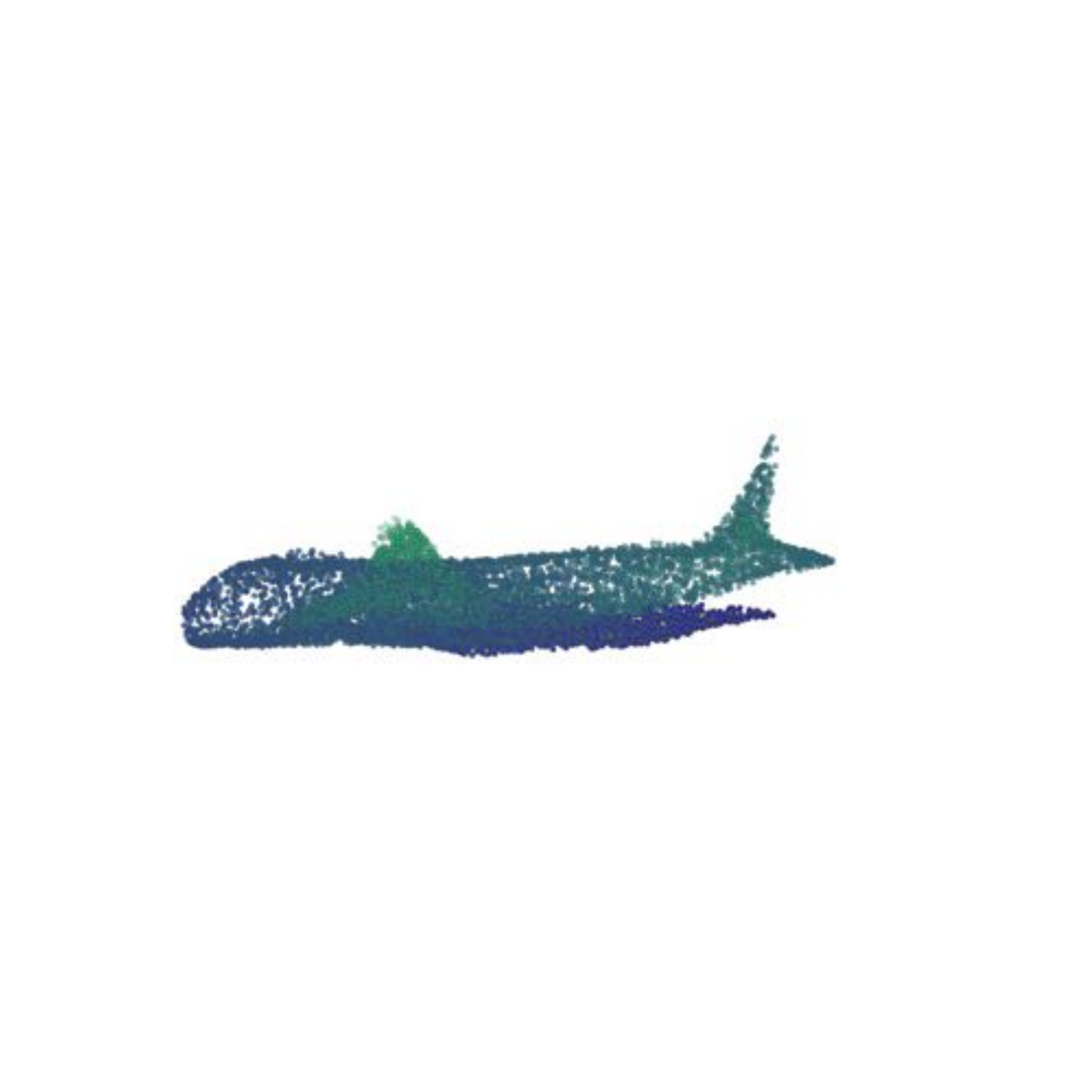}
\includegraphics[width=0.09\linewidth,trim={50mm 50mm 50mm 50mm},clip]{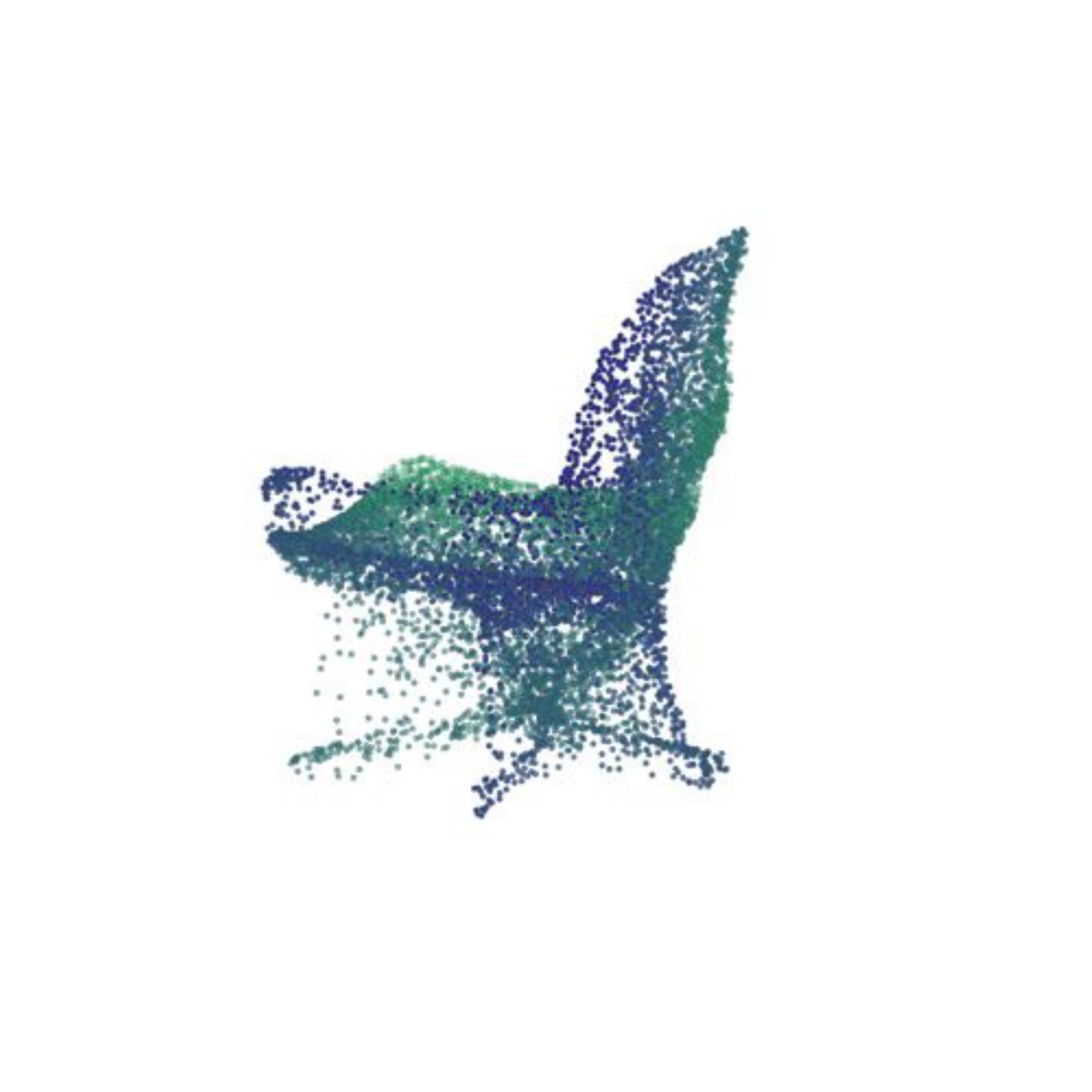}
\includegraphics[width=0.09\linewidth,trim={50mm 50mm 50mm 50mm},clip]{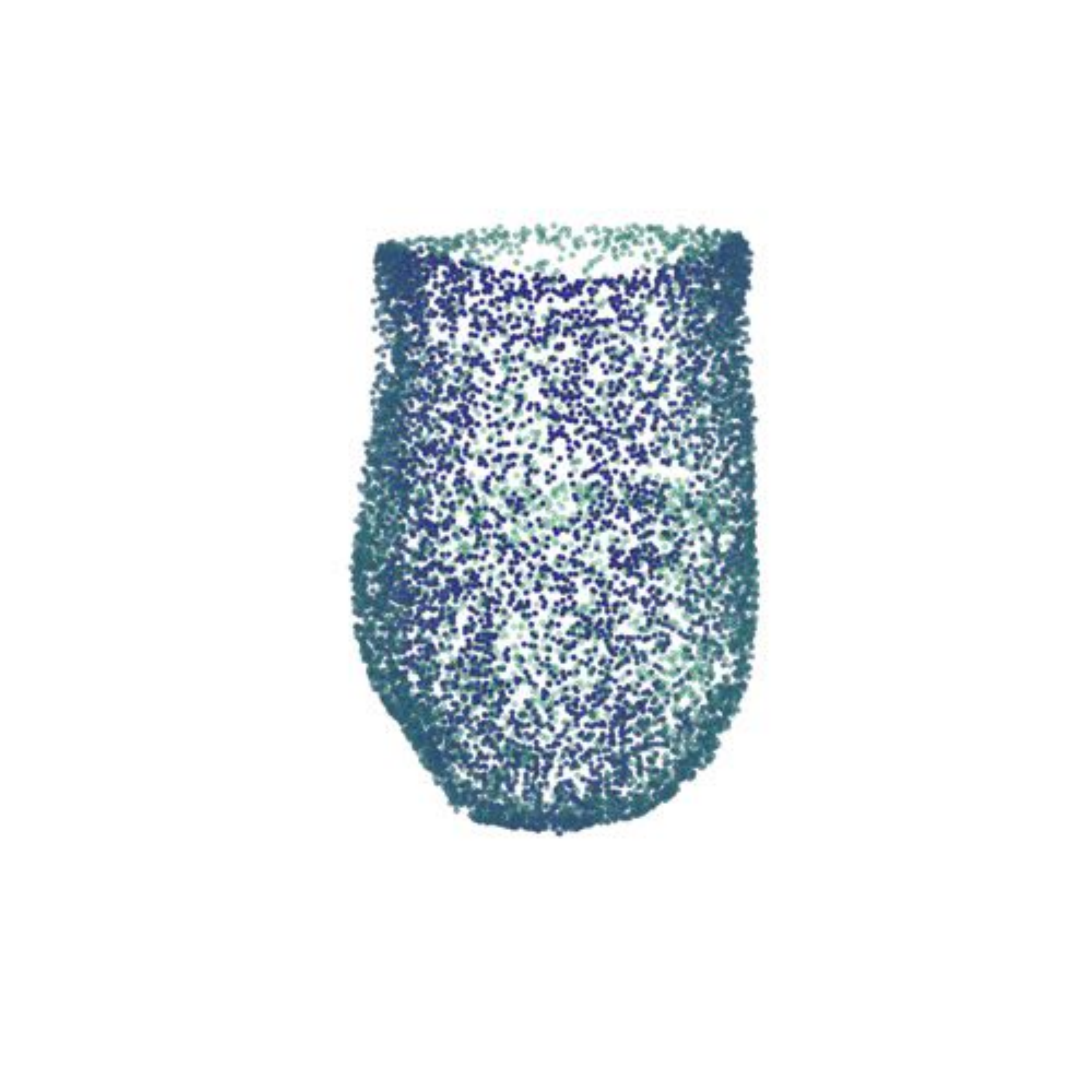}
\includegraphics[width=0.09\linewidth,trim={50mm 50mm 50mm 50mm},clip]{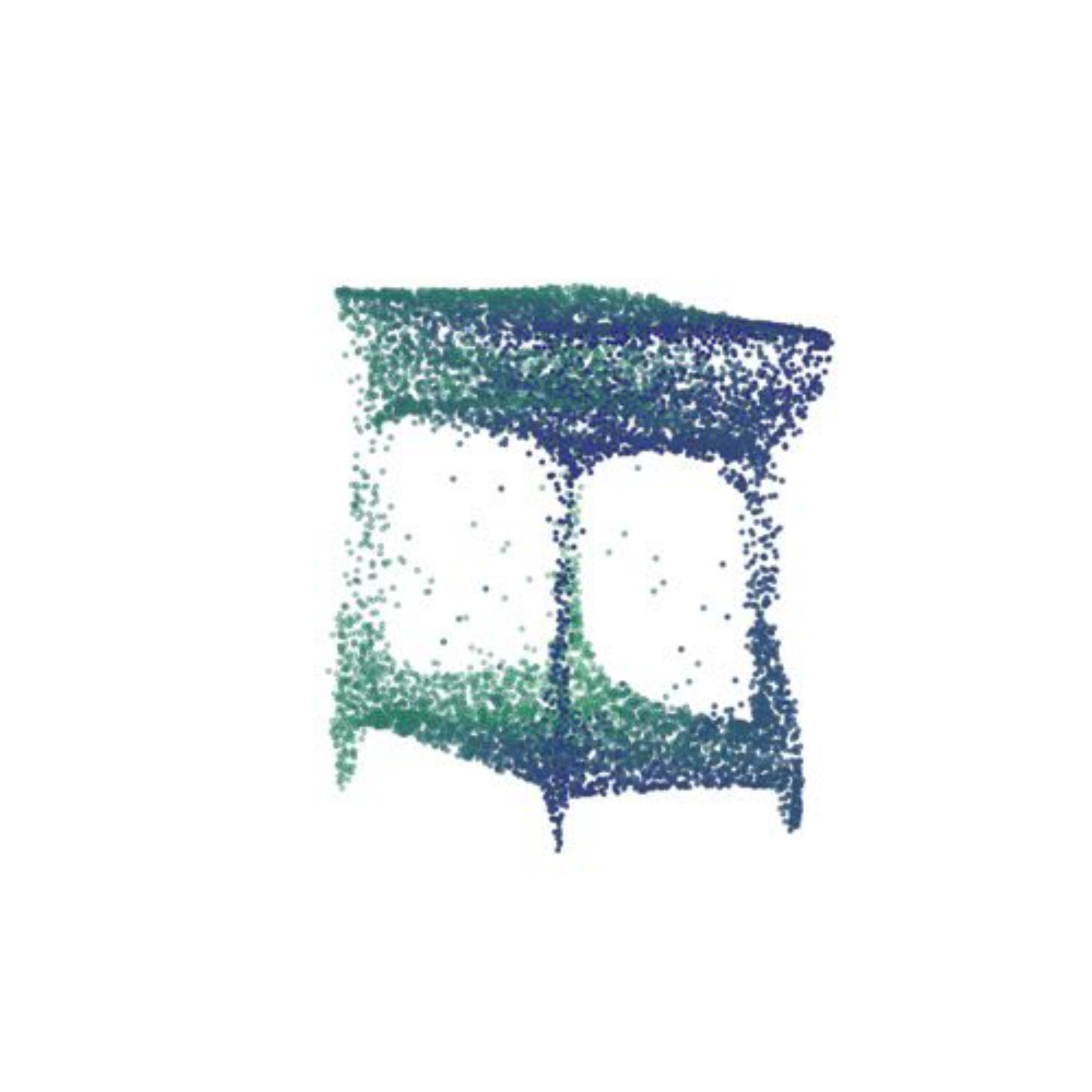}
\includegraphics[width=0.09\linewidth,trim={50mm 50mm 50mm 50mm},clip]{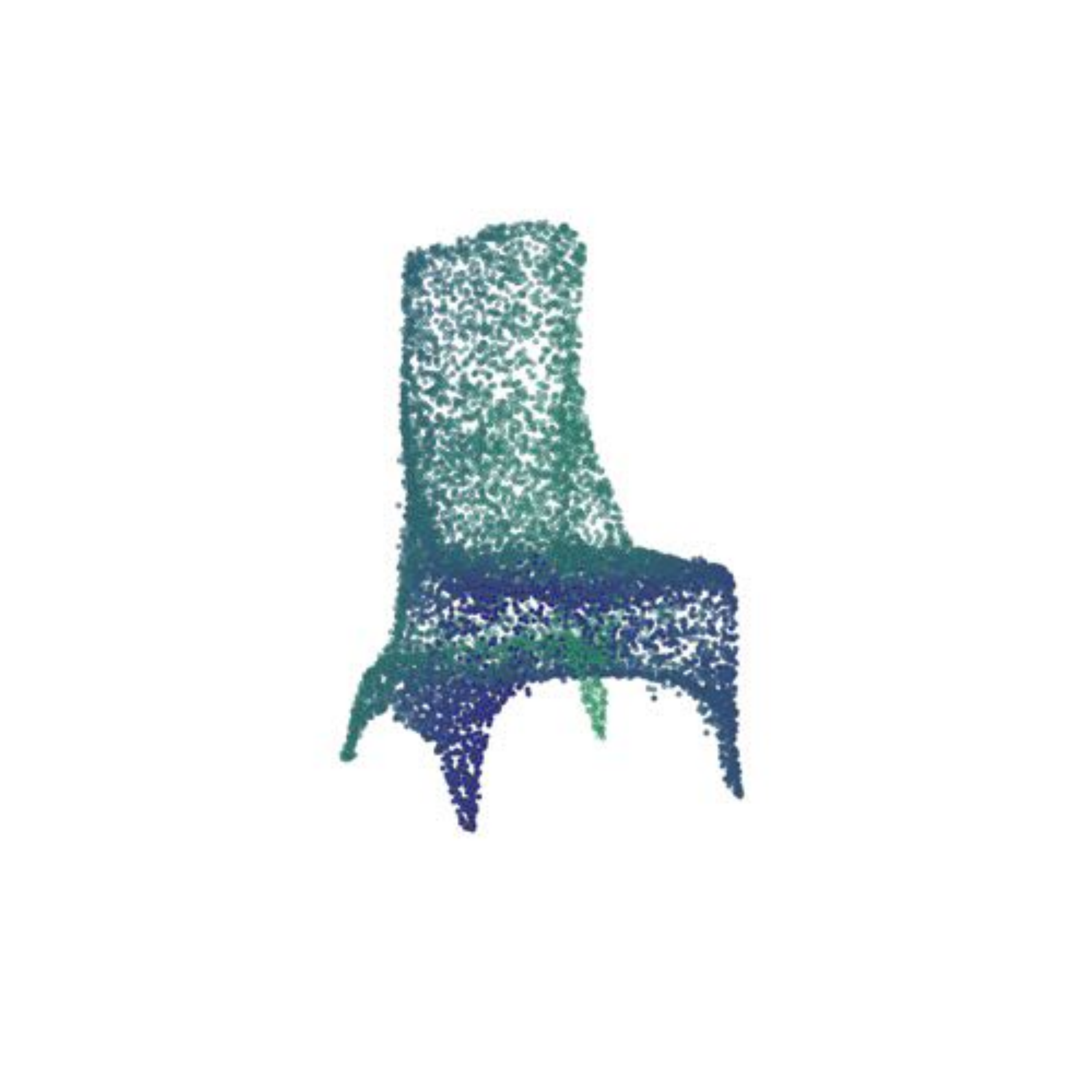}
\caption{Randomly sampled objects and corresponding point cloud from the hierarchical sampling. 
Even if there are some defects, the objects are smooth, symmetric and structured. 
It suggests PC-GAN captures inherent patterns 
and learns basic building blocks of objects.}
\label{fig:random}
\end{figure}

In Section~\ref{sec:algo}, we propose a hierarchical sampling process for sampling point clouds. In the first hierarchy,
the generator $G_\theta$ samples a object, while the second generator $G_x$ samples points to form the point cloud. 
The randomly sampled results without given any data as input are shown in Figure~\ref{fig:random}.
The point clouds are all smooth, structured and almost symmetric. It shows PC-GAN captures inherent symmetries and patterns in all the randomly sampled objects, even if overall object is not perfectly formed. This highlights that learning point-wise generation scheme encourages learning basic building blocks of objects.

\subsection{Understand the Learned Manifold}
\begin{figure}[h]
\hspace{-3mm}
\includegraphics[width=0.14\linewidth,trim={50mm 50mm 50mm 50mm},clip]{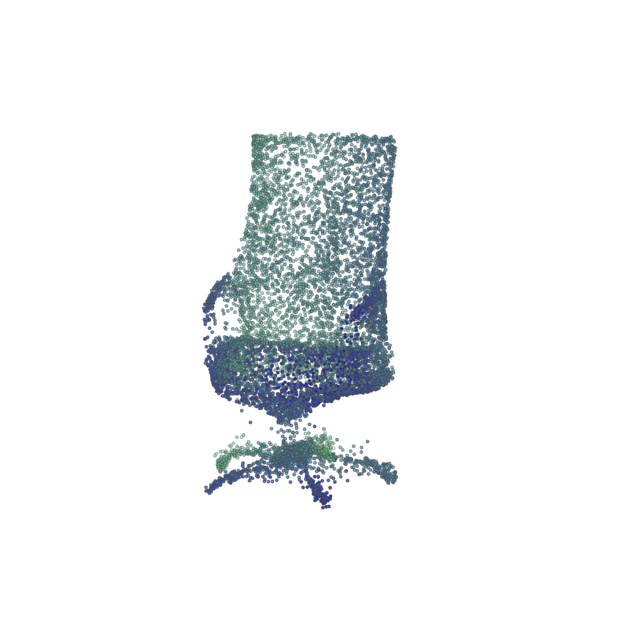}
\includegraphics[width=0.14\linewidth,trim={50mm 50mm 50mm 50mm},clip]{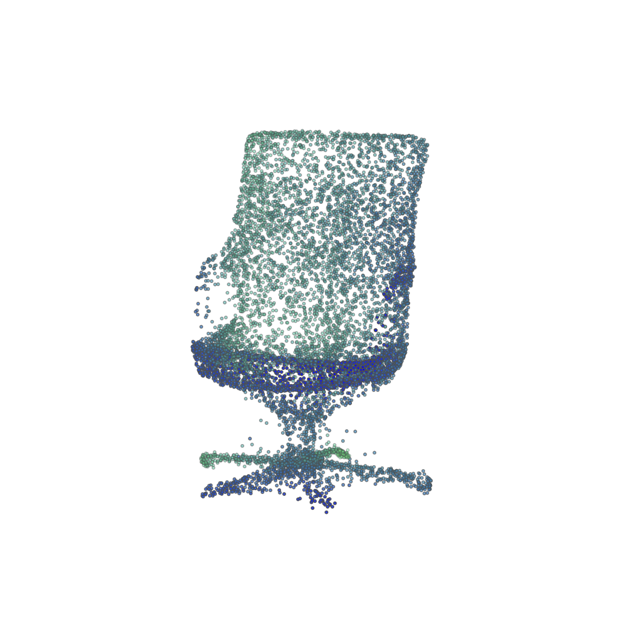}
\includegraphics[width=0.14\linewidth,trim={50mm 50mm 50mm 50mm},clip]{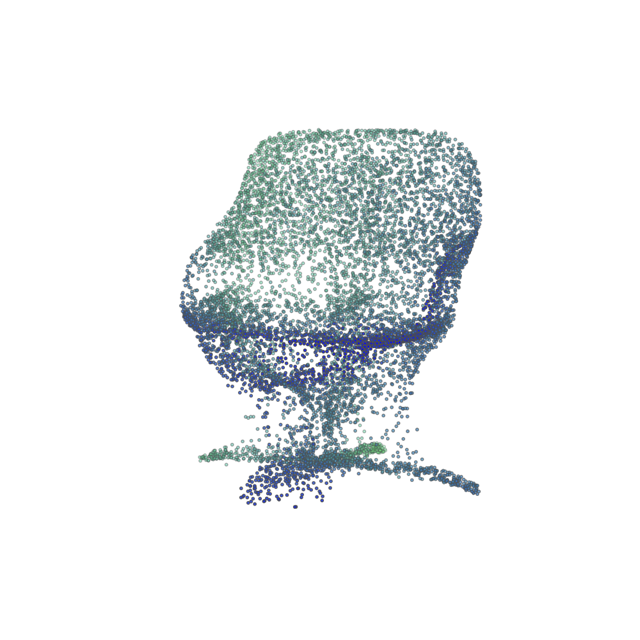}
\includegraphics[width=0.14\linewidth,trim={50mm 50mm 50mm 50mm},clip]{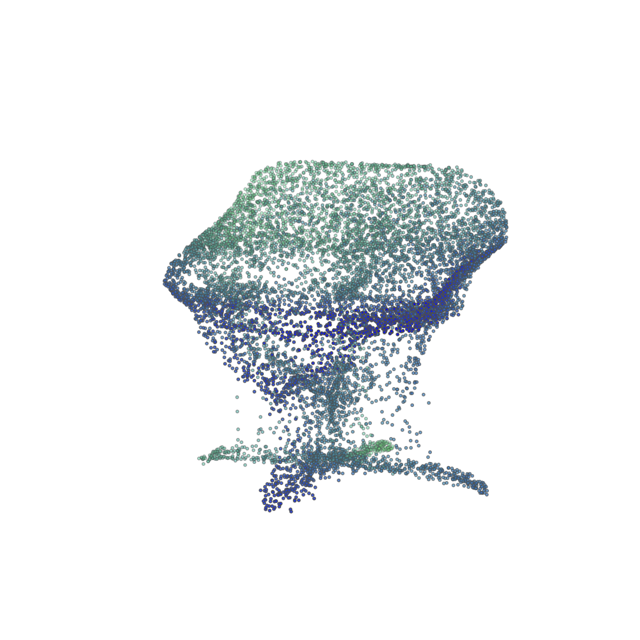}
\includegraphics[width=0.14\linewidth,trim={50mm 50mm 50mm 50mm},clip]{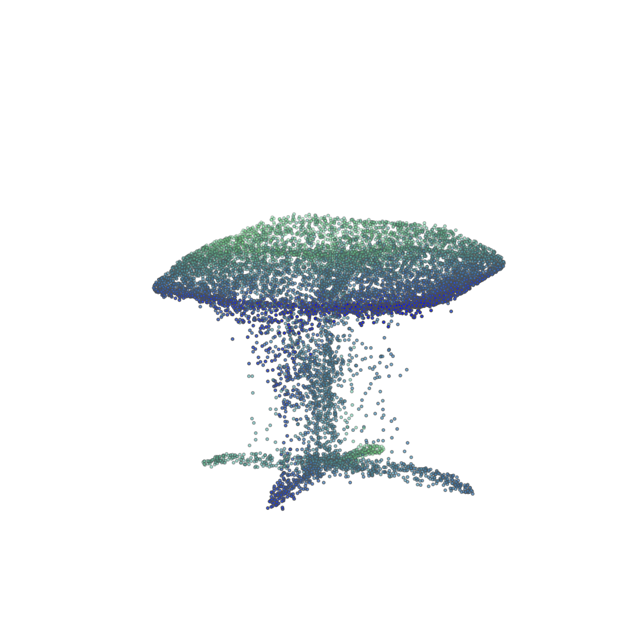}
\includegraphics[width=0.14\linewidth,trim={50mm 50mm 50mm 50mm},clip]{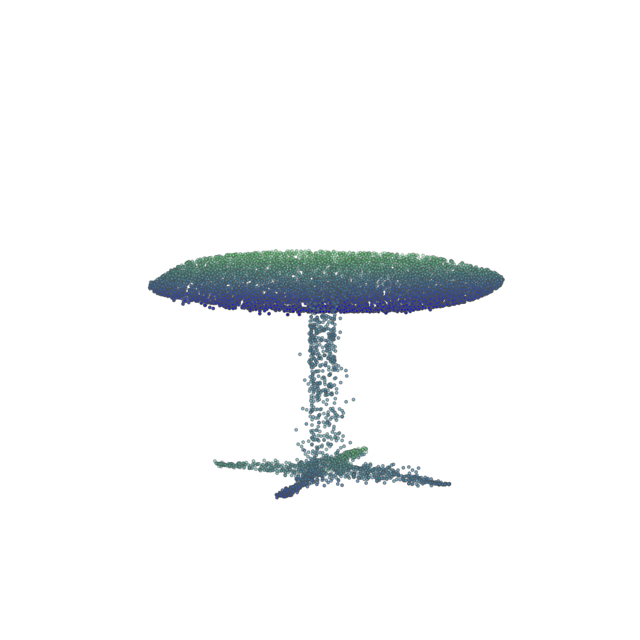}
\includegraphics[width=0.14\linewidth,trim={50mm 50mm 50mm 50mm},clip]{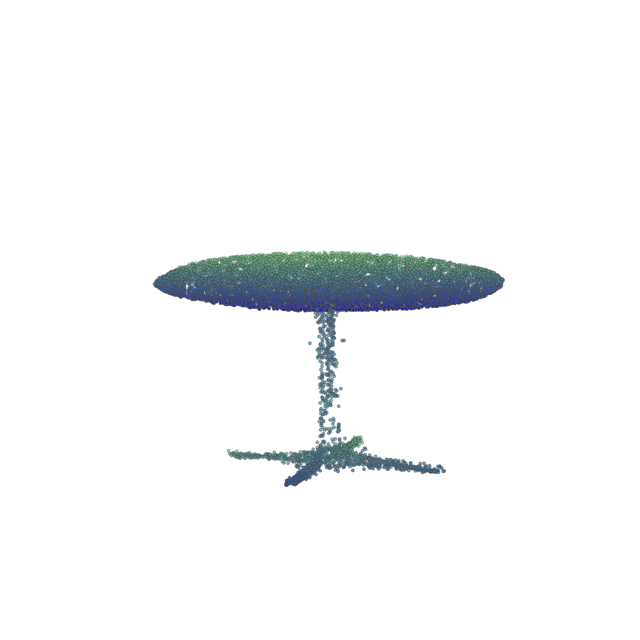}
\caption{Interpolating between a table and a chair point clouds, using our latent space representation.
\label{fig:inter}
}
\end{figure}

\paragraph{Interpolation}
A commonly used method to demonstrate quality of the learned latent space is showing whether the interpolation
between two objects on the latent space results in smooth change. 
We interpolate the inferred representations from two objects by the inference network, and use the generator to sample
points. The inter-class result is shown in Figure~\ref{fig:inter}.

\begin{figure}
\hspace{-3mm}
\includegraphics[width=0.14\linewidth,trim={50mm 100mm 50mm 60mm},clip]{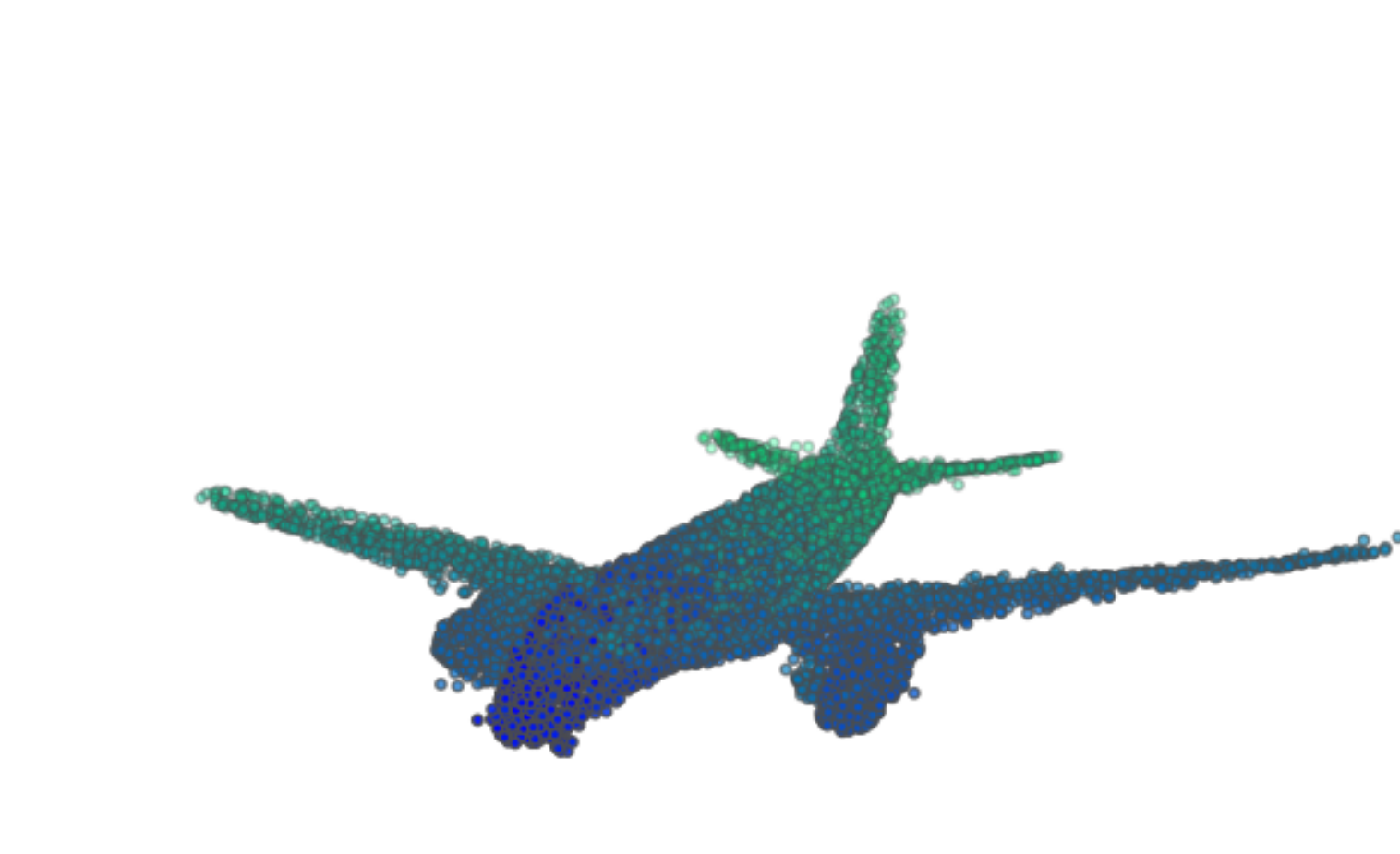}
\includegraphics[width=0.14\linewidth,trim={50mm 100mm 50mm 60mm},clip]{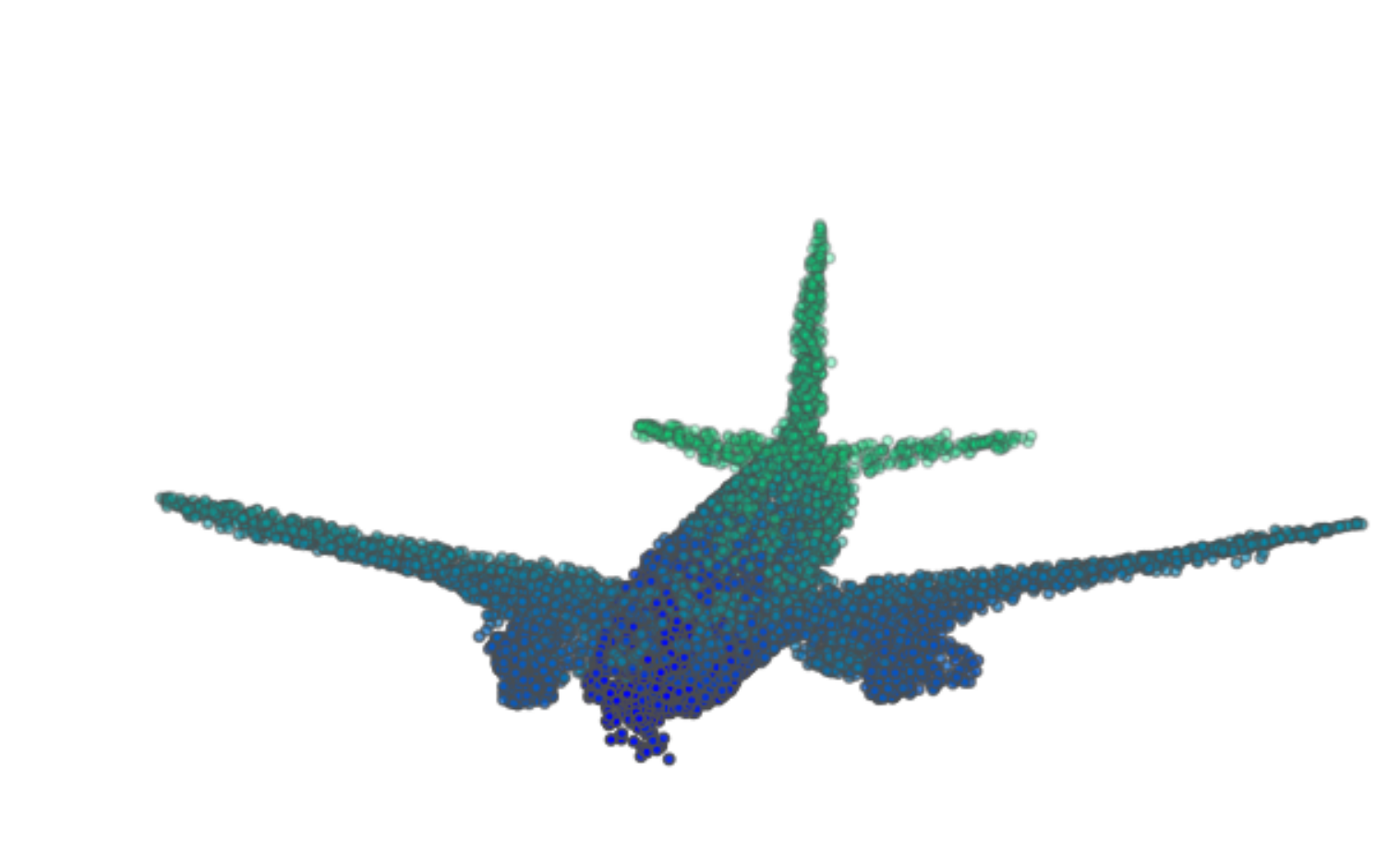}
\includegraphics[width=0.14\linewidth,trim={50mm 100mm 50mm 60mm},clip]{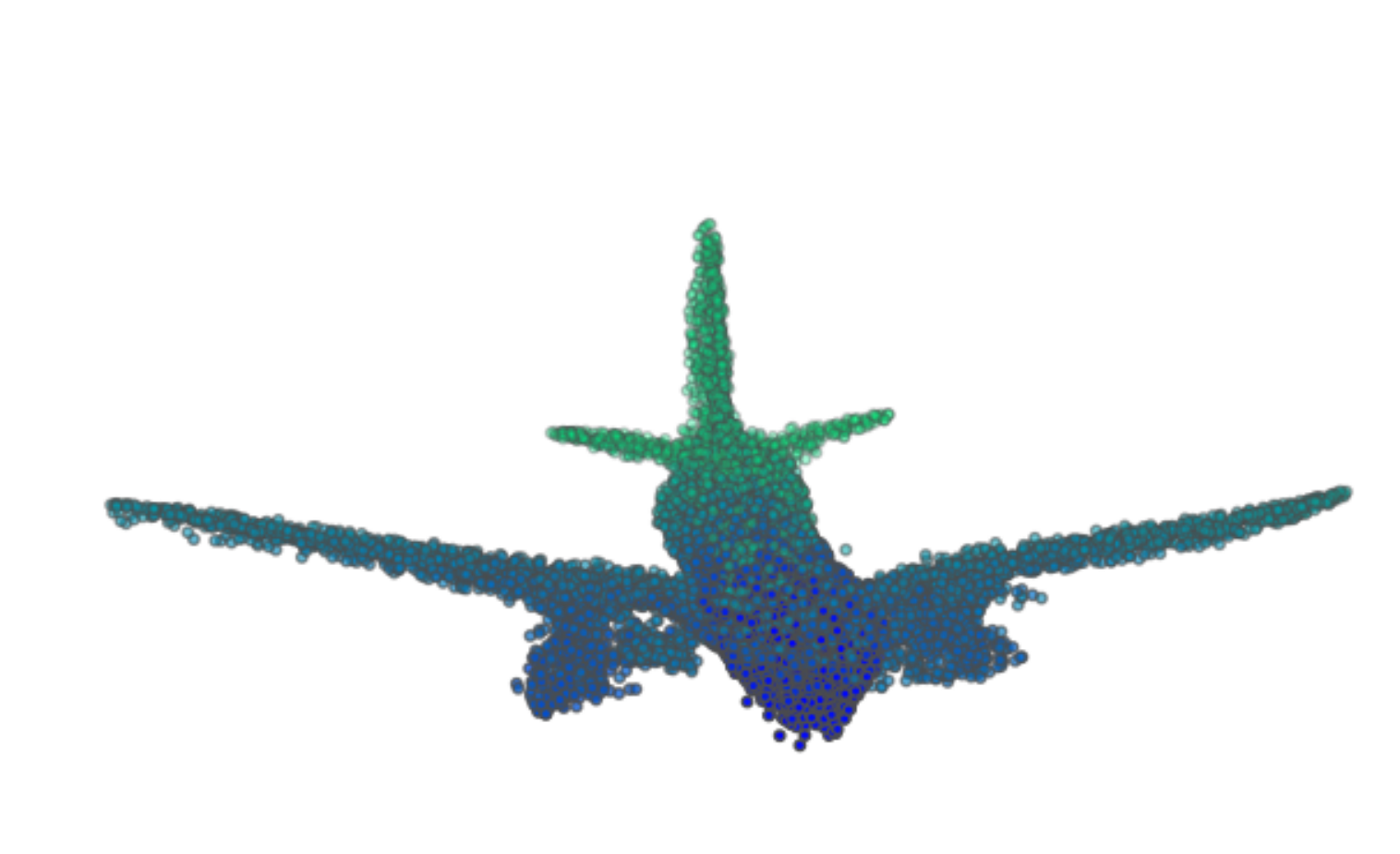}
\includegraphics[width=0.14\linewidth,trim={50mm 100mm 50mm 60mm},clip]{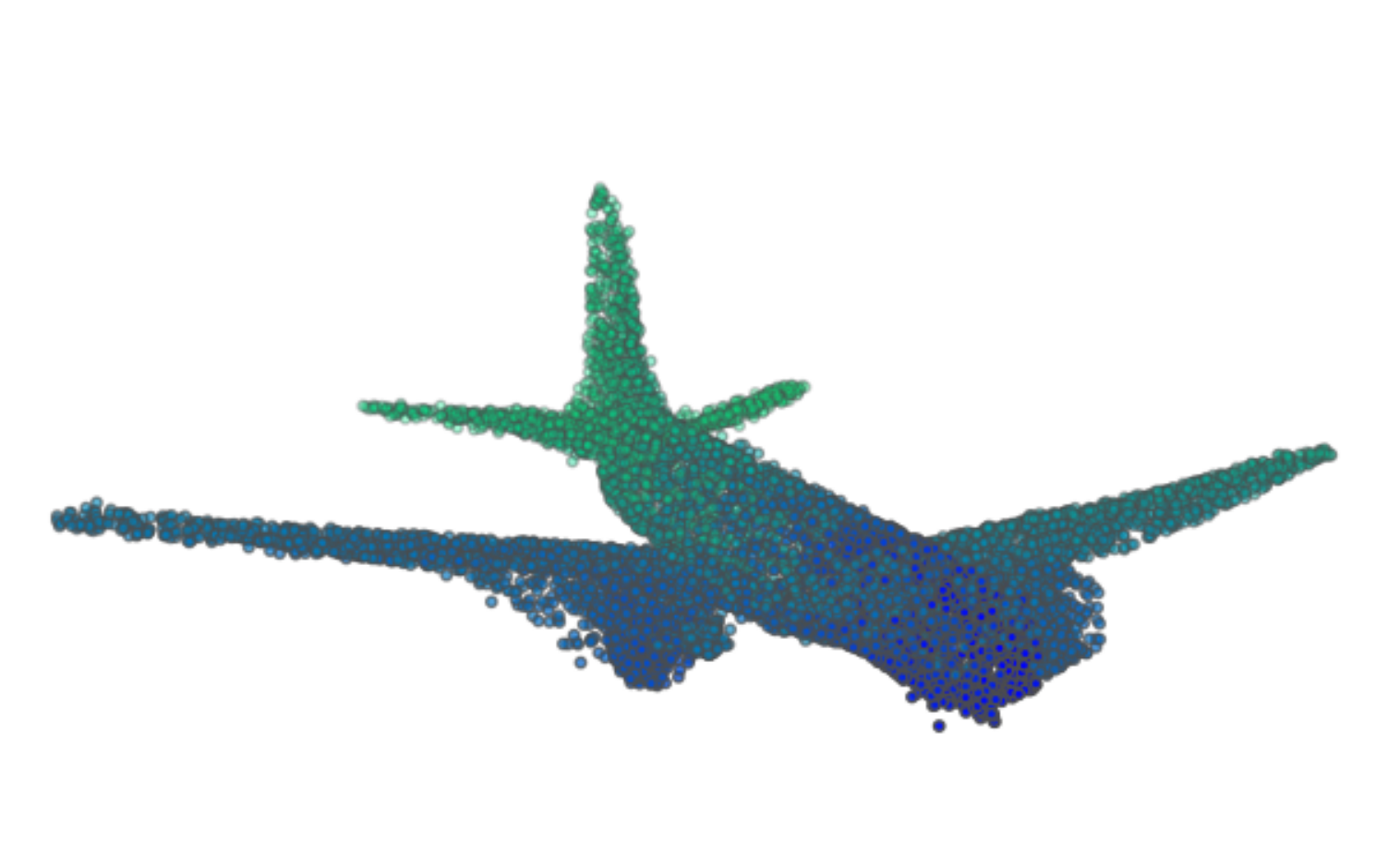}
\includegraphics[width=0.14\linewidth,trim={50mm 100mm 50mm 60mm},clip]{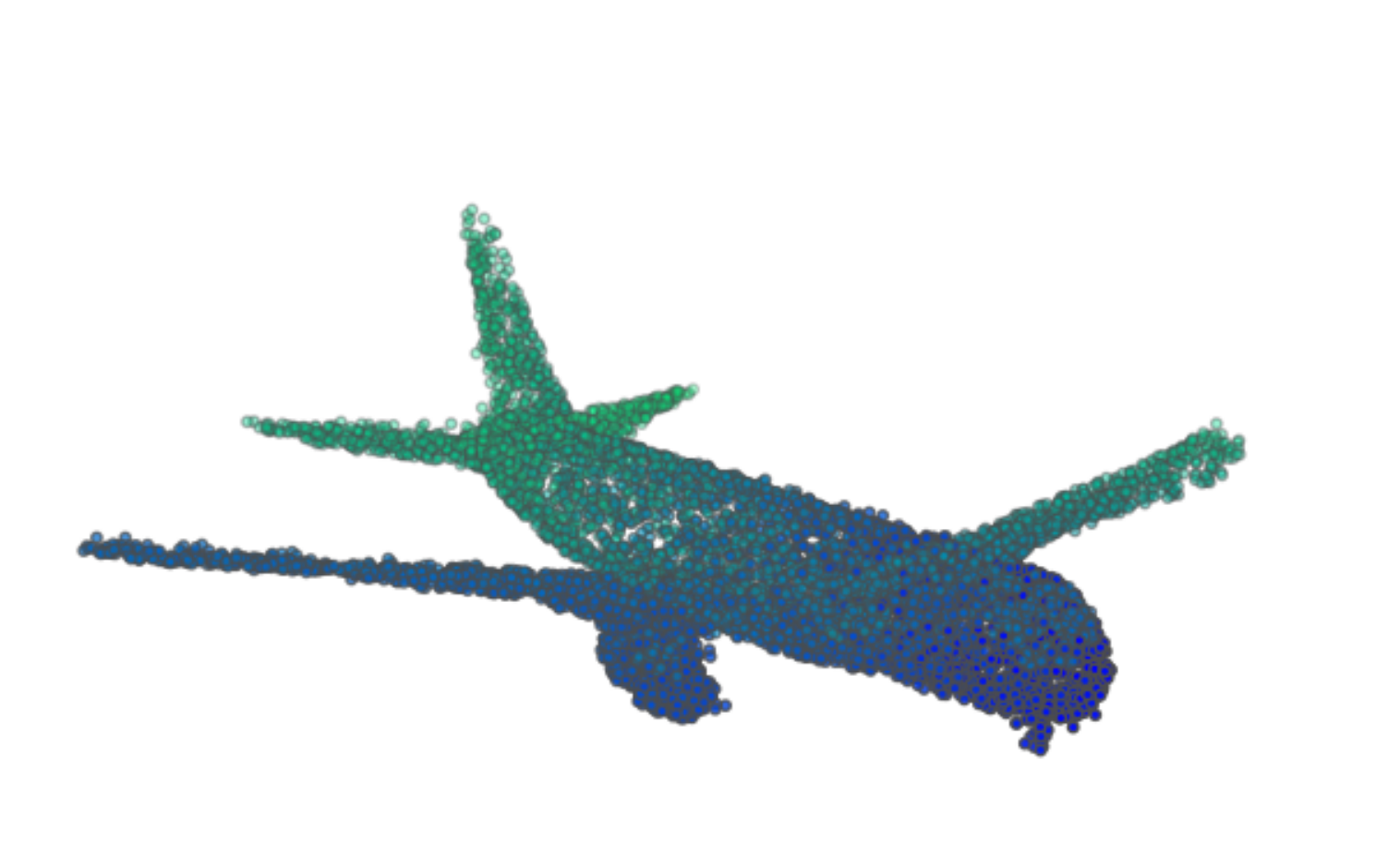}
\includegraphics[width=0.14\linewidth,trim={50mm 100mm 50mm 60mm},clip]{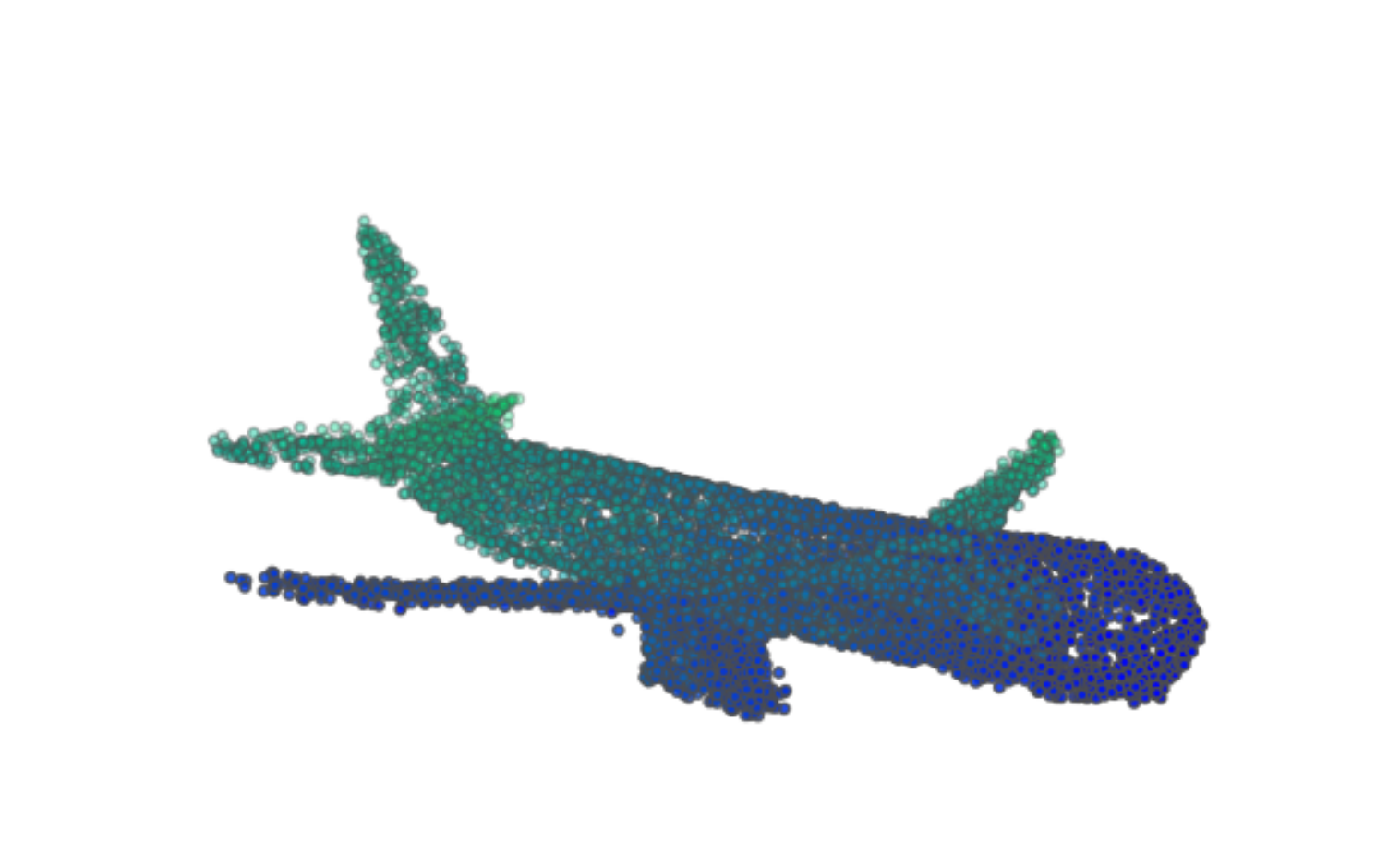}
\includegraphics[width=0.14\linewidth,trim={50mm 100mm 50mm 60mm},clip]{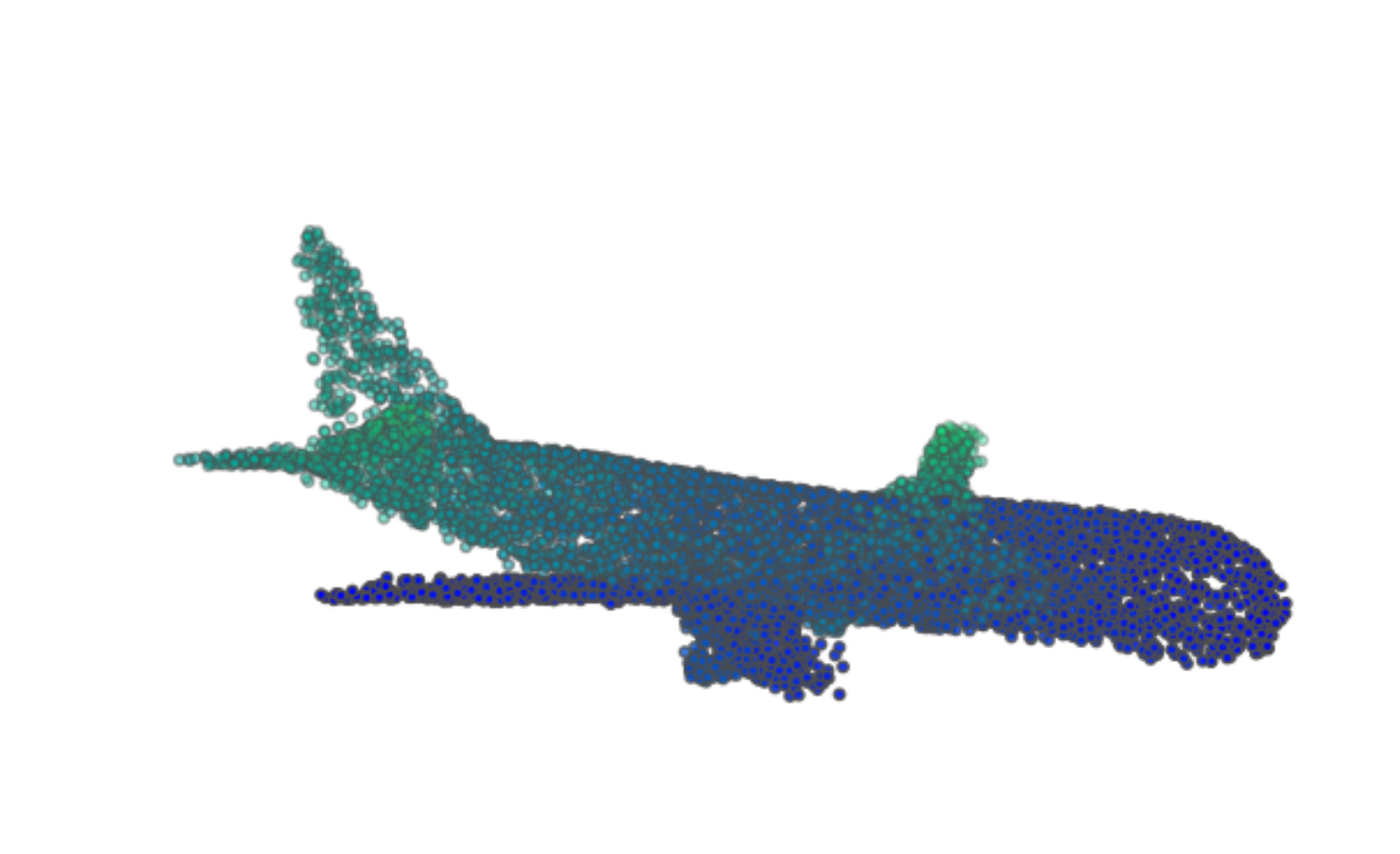}
\includegraphics[width=0.135\linewidth,trim={50mm 50mm 50mm 50mm},clip]{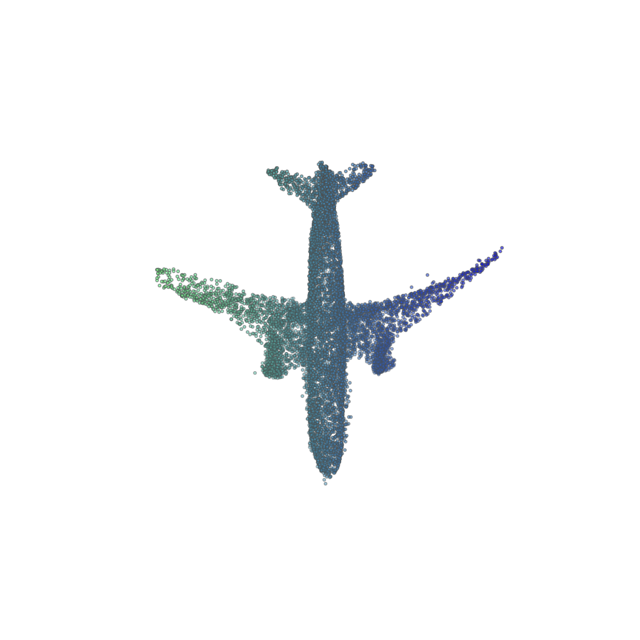}
\includegraphics[width=0.135\linewidth,trim={50mm 50mm 50mm 50mm},clip]{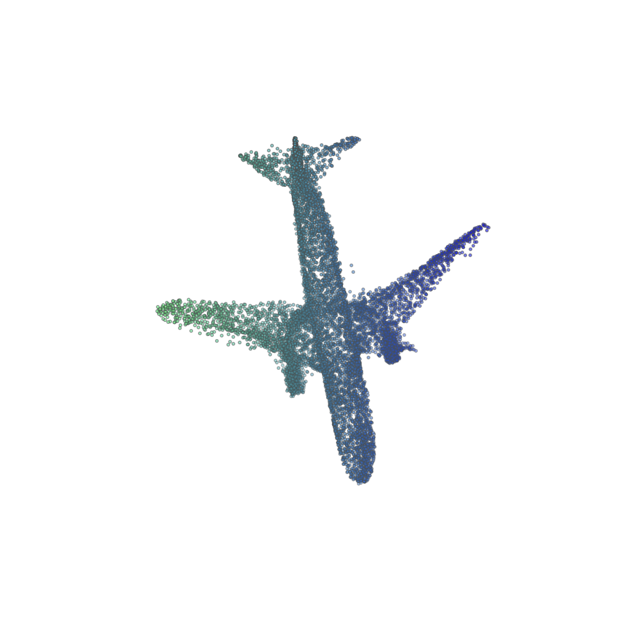}
\includegraphics[width=0.135\linewidth,trim={50mm 50mm 50mm 50mm},clip]{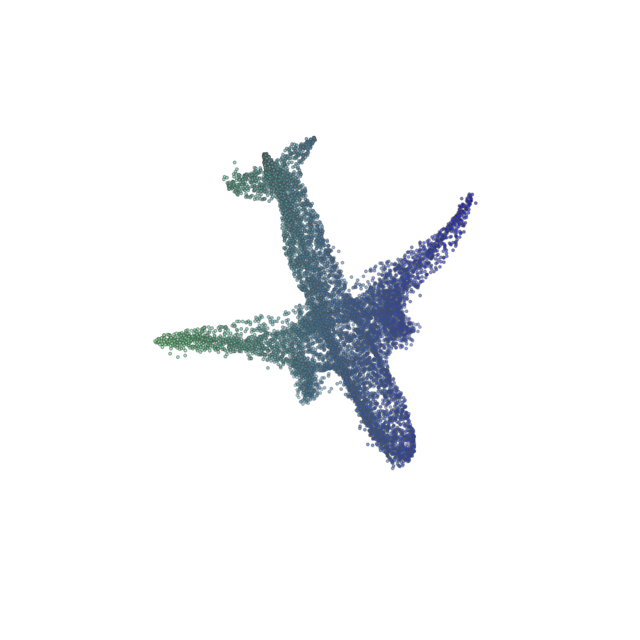}
\includegraphics[width=0.135\linewidth,trim={50mm 50mm 50mm 50mm},clip]{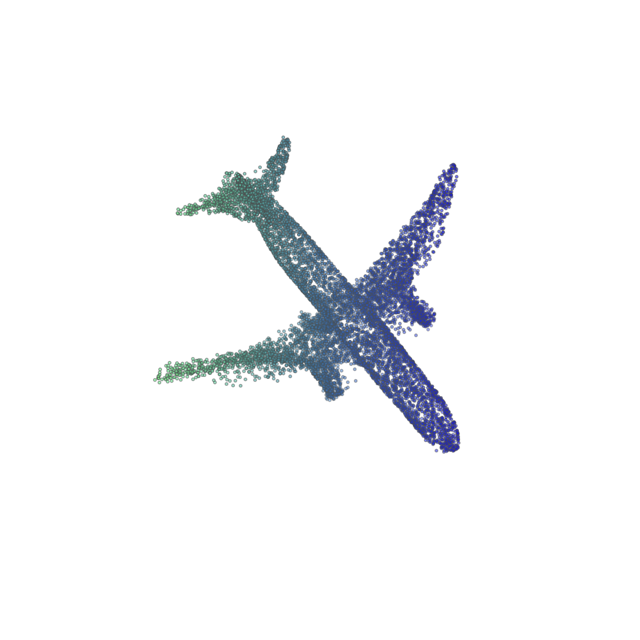}
\includegraphics[width=0.135\linewidth,trim={50mm 50mm 50mm 50mm},clip]{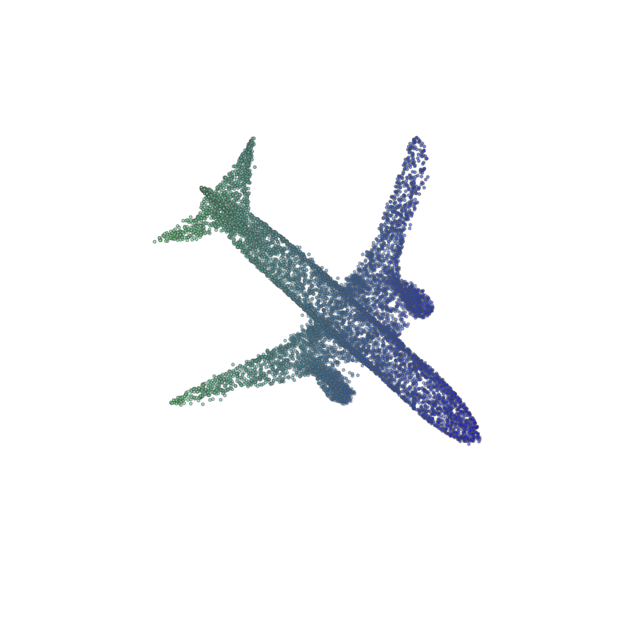}
\includegraphics[width=0.135\linewidth,trim={50mm 50mm 50mm 50mm},clip]{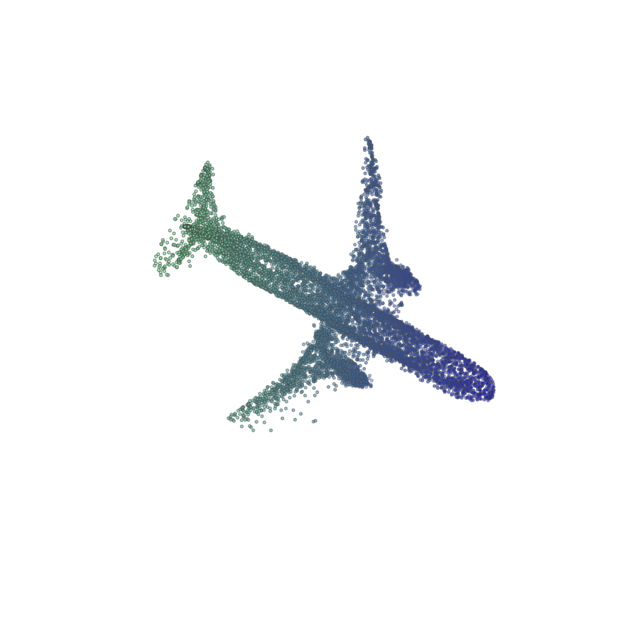}
\includegraphics[width=0.135\linewidth,trim={50mm 50mm 50mm 50mm},clip]{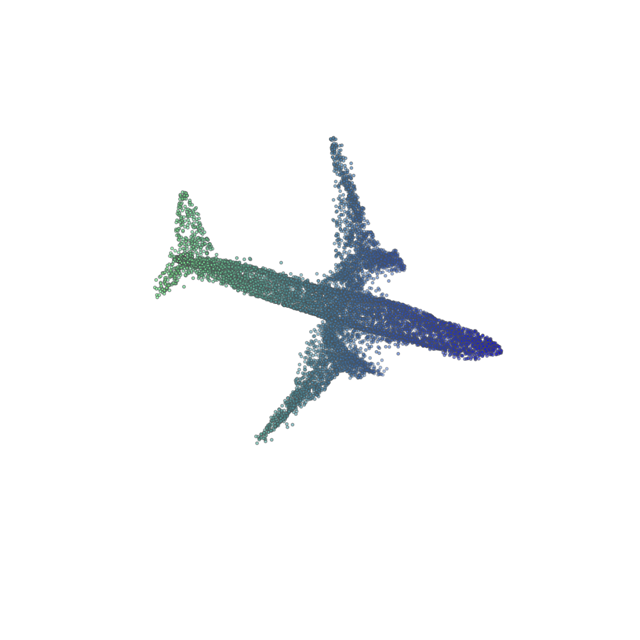}
\caption{Interpolating between rotation of an aeroplane, using our latent space representation.
\label{fig:rotation}
}
\end{figure}

It is also popular to show intra-class interpolation. 
In addition showing simple intra-class interpolations, where the objects are almost aligned, we present an interesting study on interpolations between rotations.  
During  the training, we only rotate data with $8$ possible angles for augmentation, here we show it generalizes to other
unseen rotations as shown in Figure~\ref{fig:rotation}.

However, if we linearly interpolate the code, the resulted change is scattered and not smooth as shown in
Figure~\ref{fig:rotation}. 
Instead of using linear interpolation, 
We train a 2-layer MLP with limited hidden layer size to be 16, where the input is the angle, output is the corresponding latent representation of rotated object.
We then generate the code for rotated planes with this trained MLP.
It suggests although the transformation path of rotation on the latent space is not linear, it follows a smooth
trajectory\footnote{By the capability of 1-layer MLP.}. 
It may also suggest the \emph{geodesic} path of the learned manifold may not be nearly linear between rotations. 
Finding the geodesic path with a principal method~\cite{shao2017riemannian} and
Understanding the geometry of the manifold for point cloud worth more deeper study as future work.

\paragraph{Classification} We evaluate the quality of the representation acquired from the learned inference network $Q$.
We train the inference network $Q$ and the generator $G_x$ on the training split of ModelNet40 with data augmentation 
as mentioned above for learning generative models without label information. 
We then extract the latent representation $Q(X)$ for each point clouds and train linear SVM on the that with its label. 
We apply the same setting to a linear classifier on the latent code of~\citet{achlioptas2017learning}.

We only sample $1000$ as input for our inference network $Q$. 
As the Deep Sets architecture for the inference network is invariant to number of points,
we can sample different number of points as input to the trained inference network for evaluation. 
Because of the randomness of sampling points for extracting latent representation, we repeat the experiments $20$ times
and report the average accuracy and standard deviation on the testing split in Table~\ref{tb:acc}. 
By using $1000$ points, we are already better than unsupervised algorithms, \citet{achlioptas2017learning} with $2048$
points and 3D Voxel GAN~\citep{wu2016learning}, and competitive
with the supervised learning algorithm Deep Sets.

\begin{table}[h]
\vspace{-0.25em}
\caption{Classification accuracy results.}
\label{tb:acc}
\centering
\begin{tabular}{@{}lrr@{}}
\toprule
Method & $\#$ points& Accuracy \\
\midrule
PC-GAN & 1000 & $87.5 \pm .6\%$ \\
PC-GAN & 2048 & $87.8 \pm .2\%$\\
AAE~\citep{achlioptas2017learning} & 2048 & $85.5 \pm .3\%$  \\
3D GAN~\citep{wu2016learning} & voxel & $83.3 \%$  \\
\midrule
Deep Sets~\citep{zaheer2017deep} & 1000 & $87 \pm 1\%$\\
Deep Sets~\citep{zaheer2017deep} & 5000 & $90 \pm .3\%$\\
MVCNN~\citep{su2015multi} & images & $90.1\%$ \\
RotationNet~\citep{kanezaki2018rotationnet} & images & $97.3\%$ \\
\bottomrule
\end{tabular}
\end{table}

\paragraph{Generalization on Unseen Categories}
In above, we studied the reconstruction of unseen testing objects, while PC-GAN still saw the point clouds from the same
class during training. \emph{Here we study a more challenging task.} We train PC-GAN on first 30 (Alphabetic order) class, 
and test on the other \emph{fully unseen 10 classes}. 
Some reconstructed (conditionally generated) point clouds are shown in Figure~\ref{fig:unseen}. 
More (larger) results can be found in Appendix~\ref{sec:larger}.
For the object from the unseen classes, the conditionally generated 
point clouds still recovers main shape and reasonable geometry structure, which confirms the advantage of the proposed
PC-GAN: by enforcing the point-wise transformation, the model is forced to learn the underlying geometry structure and
the shared building blocks, instead of naively copying the input from the conditioning. 
The resulted D2F and coverage are $57.4$ and $0.36$, which are only slightly worse than 
$48.4$ and $0.38$ by training on whole 40 classes in Table~\ref{tb:result} (ModelNet40), which also supports the claims
of the good generalization ability of PC-GAN.

\begin{figure}[h]
\centering
\begin{subfigure}[b]{.32\linewidth}
\includegraphics[width=0.48\linewidth,trim={80mm 100mm 70mm 80mm},clip]{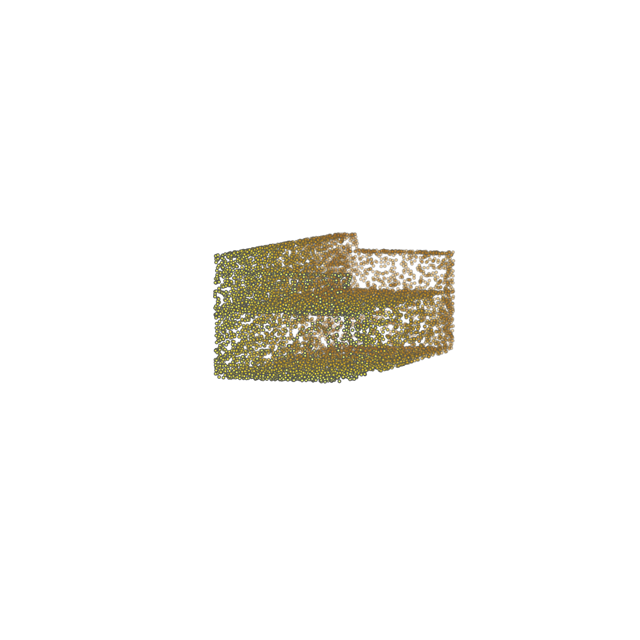}
\includegraphics[width=0.48\linewidth,trim={80mm 95mm 70mm 80mm},clip]{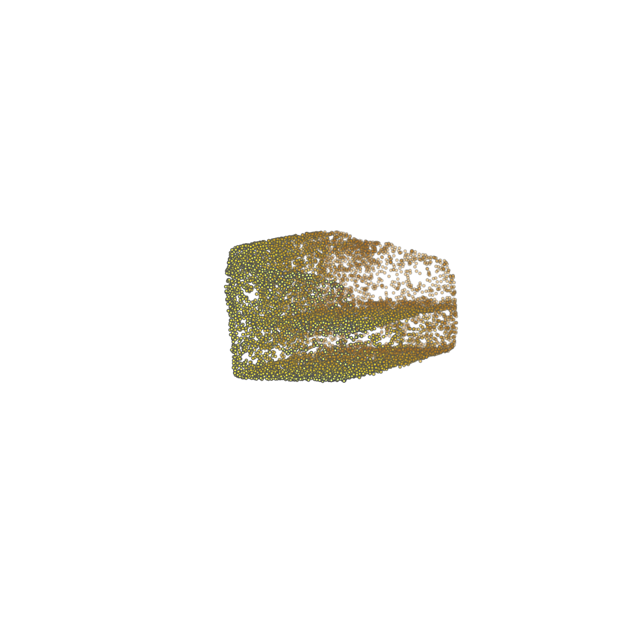}
\caption{Sofa}
\end{subfigure}
\begin{subfigure}[b]{.32\linewidth}
\includegraphics[width=0.48\linewidth,trim={10mm 30mm 10mm 30mm},clip]{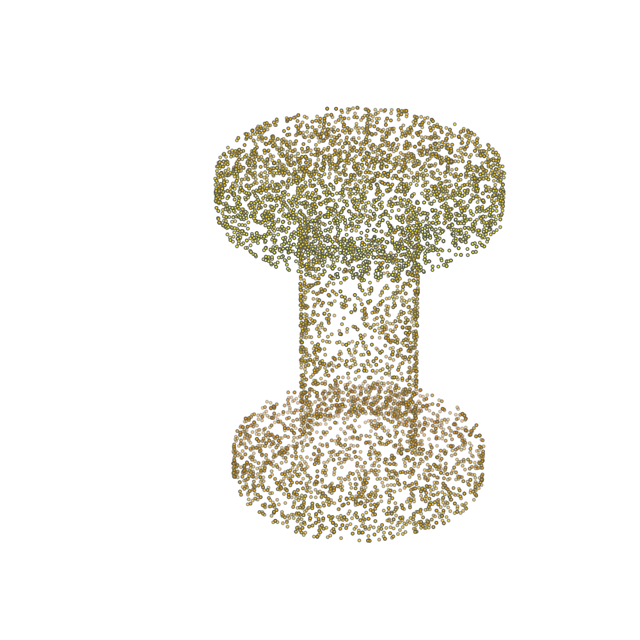}
\includegraphics[width=0.48\linewidth,trim={10mm 30mm 10mm 30mm},clip]{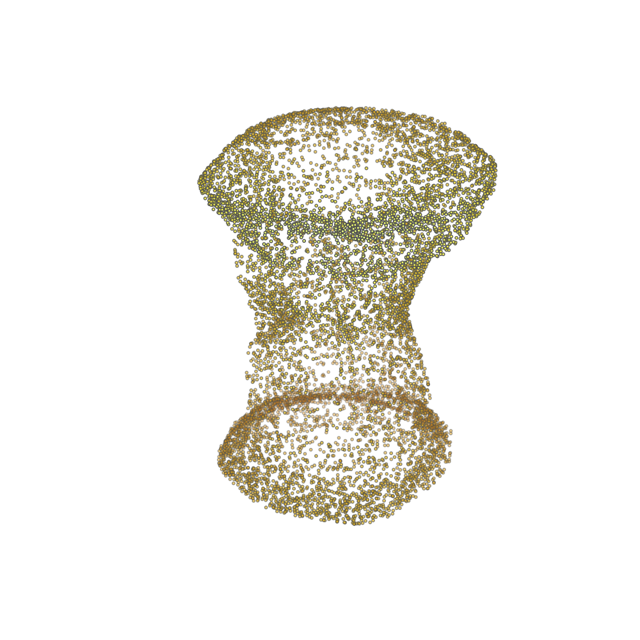}
\caption{Stool}
\end{subfigure}
\begin{subfigure}[b]{.32\linewidth}
\includegraphics[width=0.48\linewidth,trim={40mm 30mm 40mm 50mm},clip]{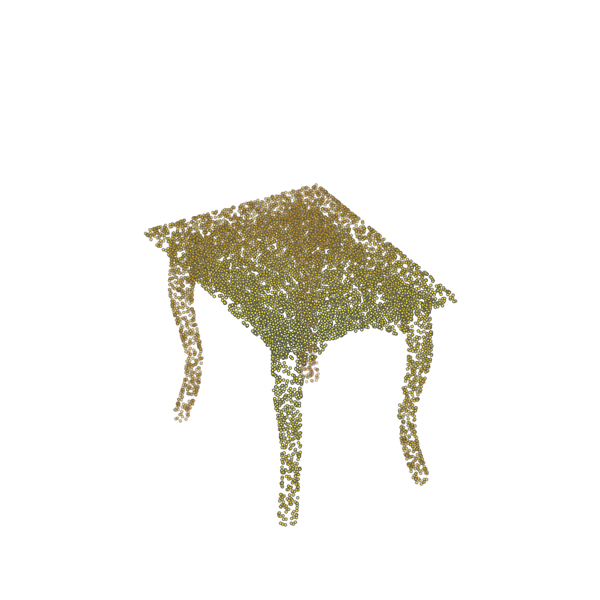}
\includegraphics[width=0.48\linewidth,trim={40mm 30mm 40mm 50mm},clip]{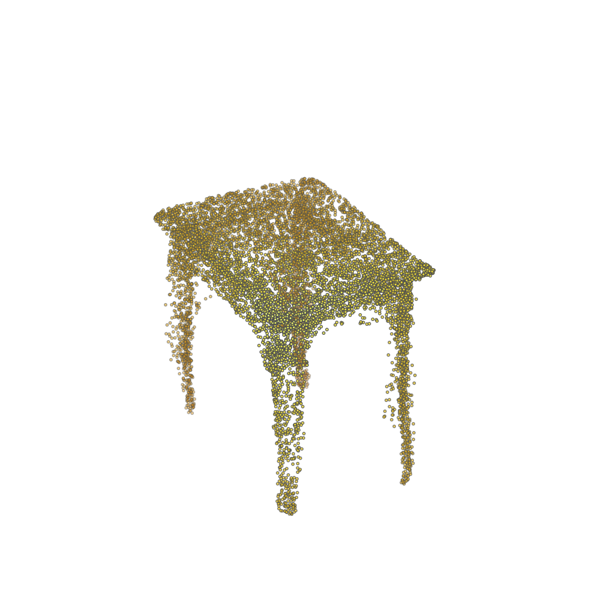}
\caption{Table}
\end{subfigure}
\begin{subfigure}[b]{.32\linewidth}
\includegraphics[width=0.48\linewidth,trim={40mm 50mm 40mm 50mm},clip]{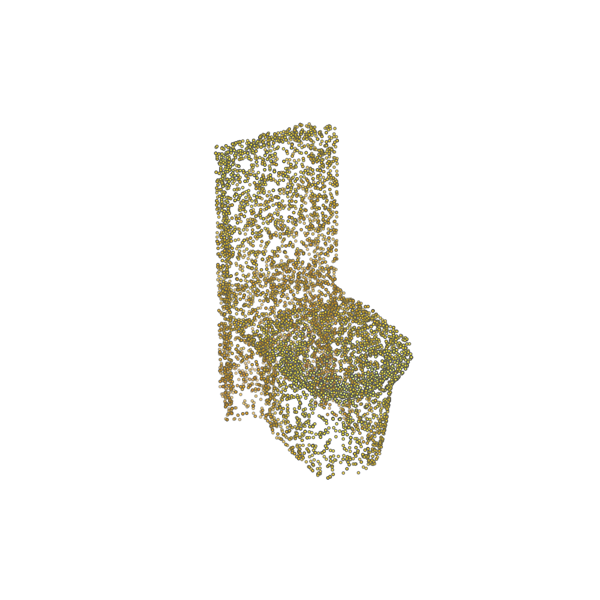}
\includegraphics[width=0.48\linewidth,trim={40mm 50mm 40mm 50mm},clip]{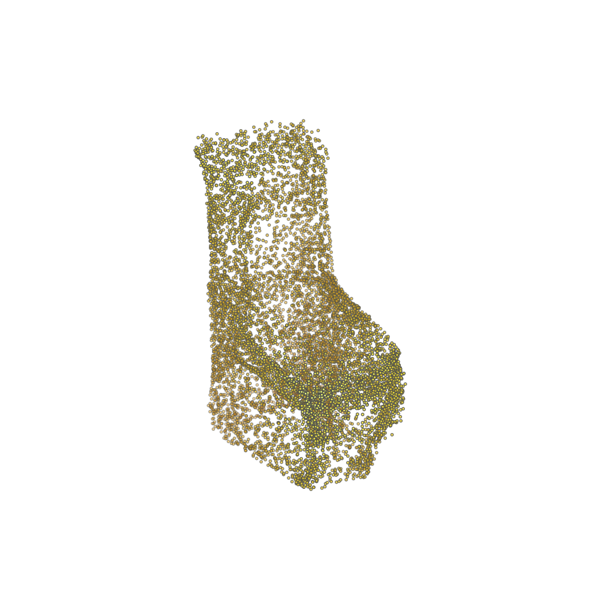}
\caption{Toilet}
\end{subfigure}
\begin{subfigure}[b]{.32\linewidth}
\includegraphics[width=0.48\linewidth,trim={40mm 50mm 40mm 50mm},clip]{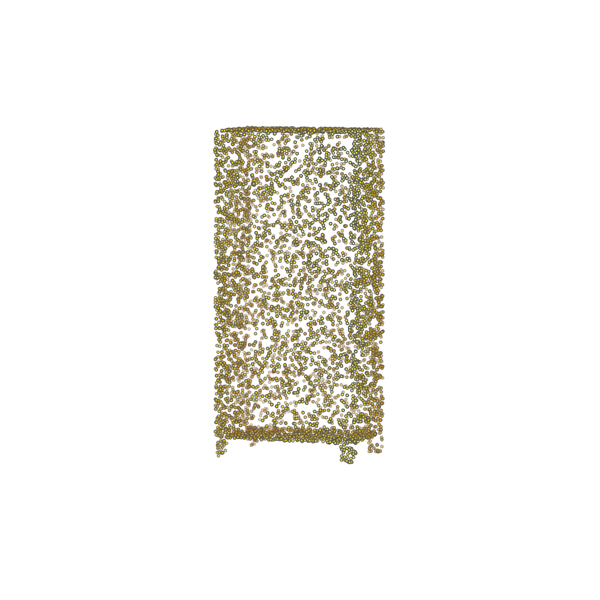}
\includegraphics[width=0.48\linewidth,trim={40mm 50mm 40mm 50mm},clip]{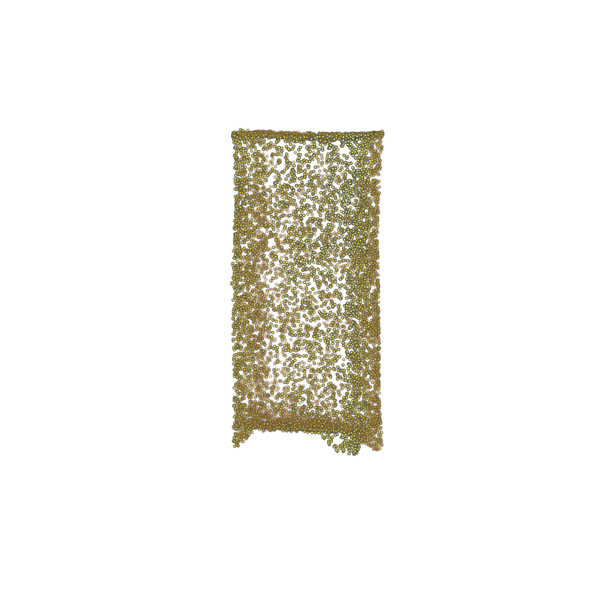}
\caption{TV Stand}
\end{subfigure}
\begin{subfigure}[b]{.32\linewidth}
\includegraphics[width=0.48\linewidth,trim={40mm 50mm 40mm 50mm},clip]{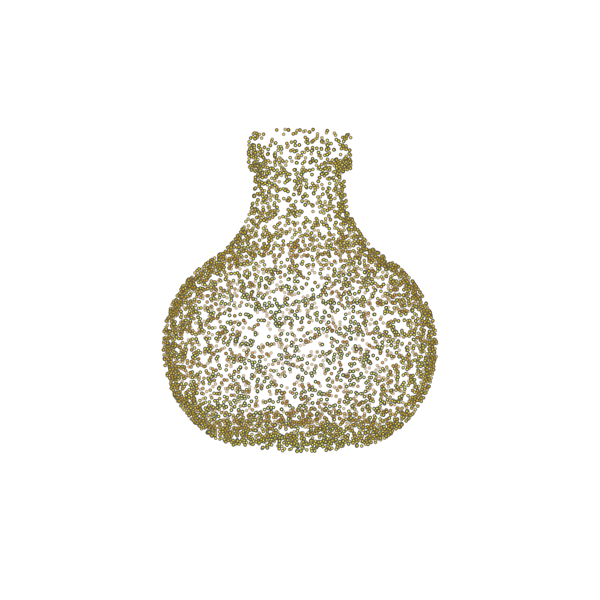}
\includegraphics[width=0.48\linewidth,trim={40mm 50mm 40mm 50mm},clip]{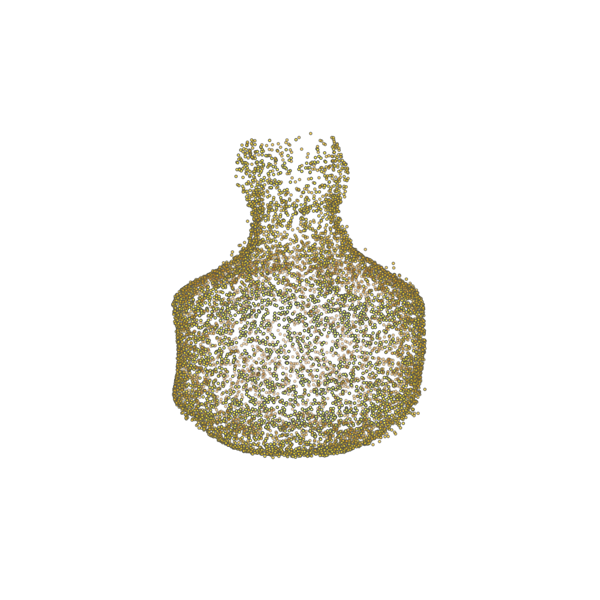}
\caption{Vase}
\end{subfigure}
\caption{The reconstructed objects from unseen categories. In each plot, LHS is true data while RHS is PC-GAN. PC-GAN generalizes well as it can match patterns and symmetries from categories seen in the past to new unseen categories.}
\label{fig:unseen}
\end{figure}

\subsection{Images to Point Cloud}
\begin{figure}[h]
\centering
\begin{subfigure}[b]{.4\linewidth}
\includegraphics[width=0.48\linewidth,trim={40mm 80mm 50mm 90mm},clip]{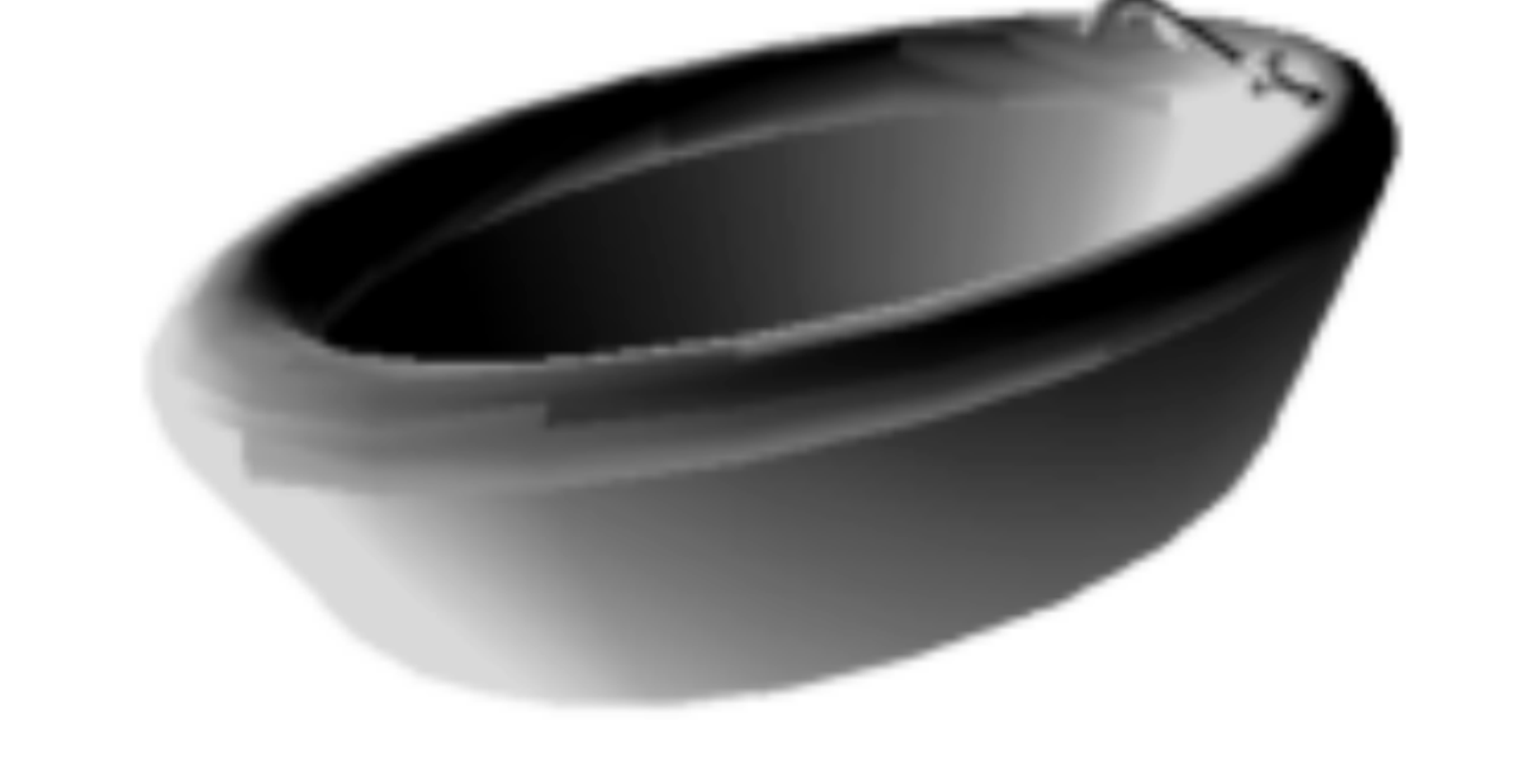}
\includegraphics[width=0.48\linewidth,trim={60mm 90mm 60mm 80mmm},clip]{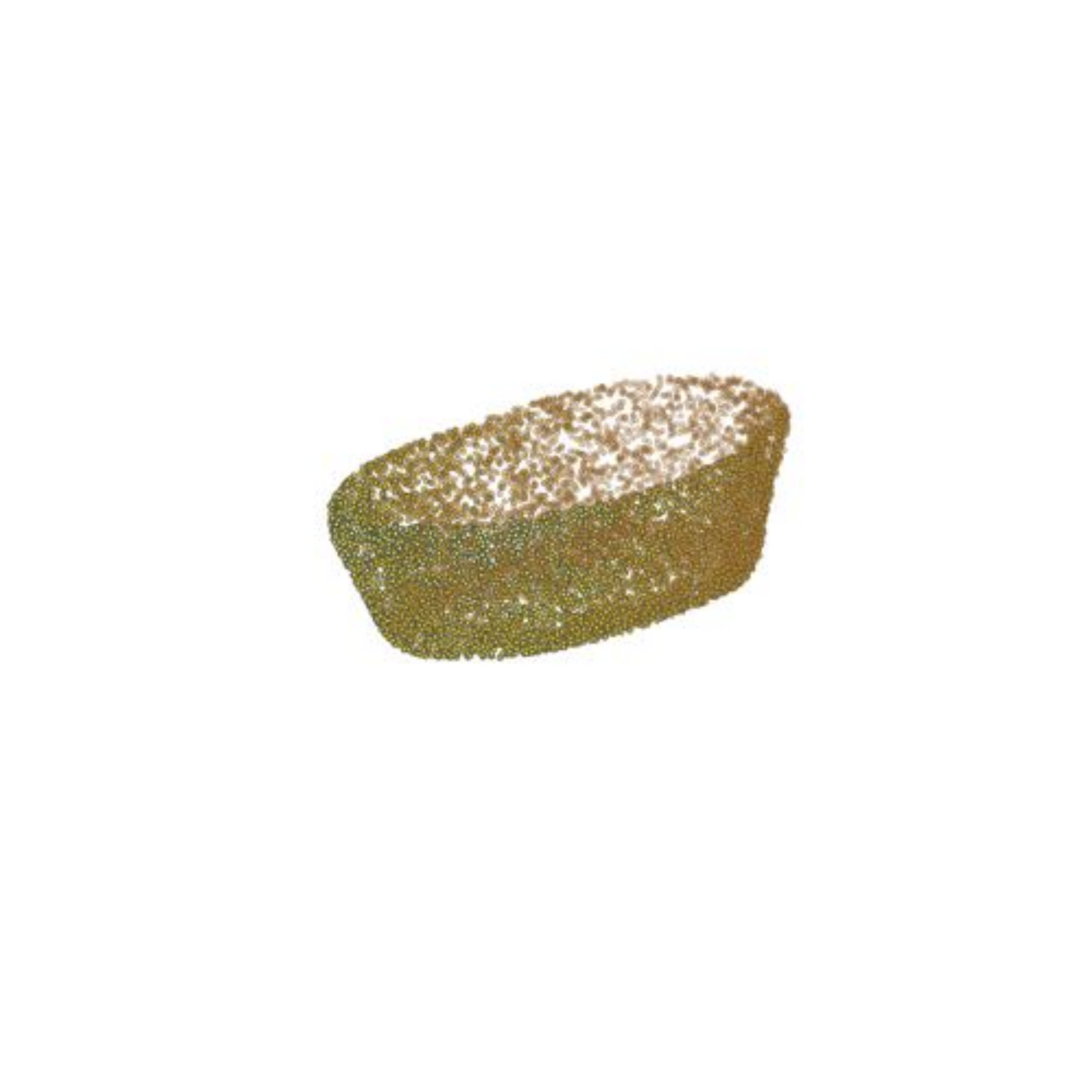}
\caption{Lamp}
\end{subfigure}
\begin{subfigure}[b]{.4\linewidth}
\includegraphics[width=0.48\linewidth,trim={10mm 50mm 10mm 80mm},clip]{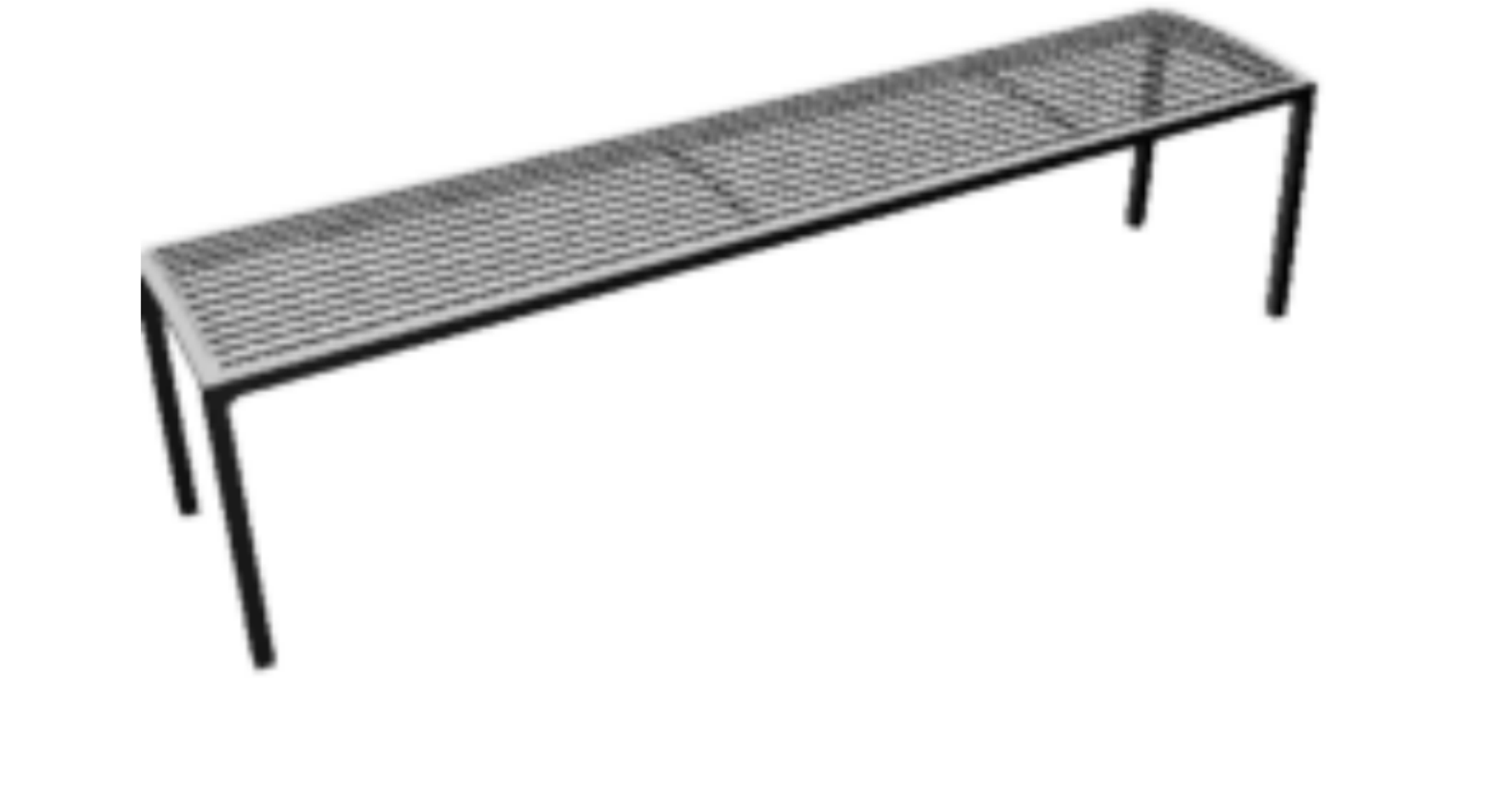}
\includegraphics[width=0.48\linewidth,trim={50mm 50mm 50mm 80mmm},clip]{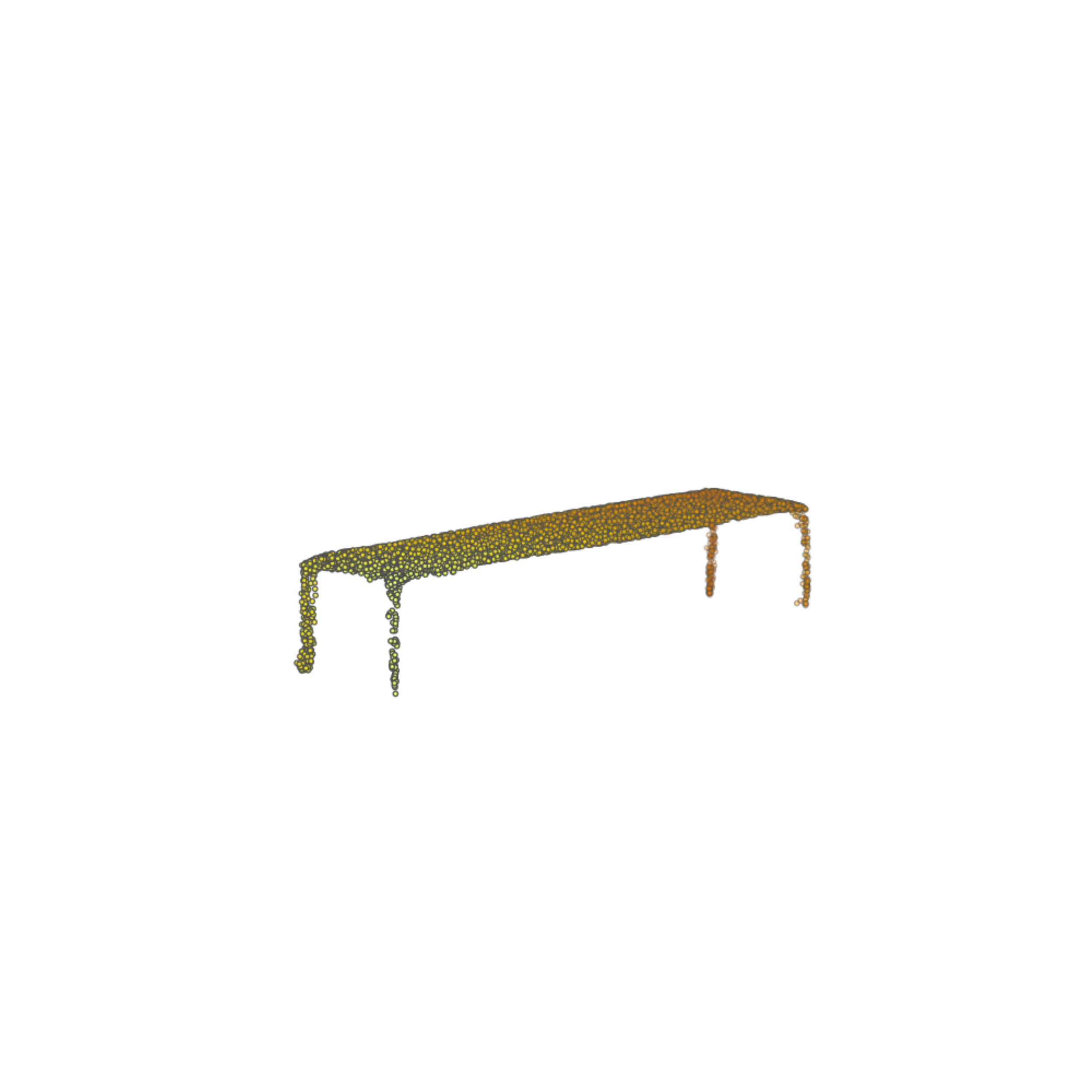}
\caption{Chair}
\end{subfigure}
\begin{subfigure}[b]{.4\linewidth}
\includegraphics[width=0.48\linewidth,trim={40mm 50mm 30mm 50mm},clip]{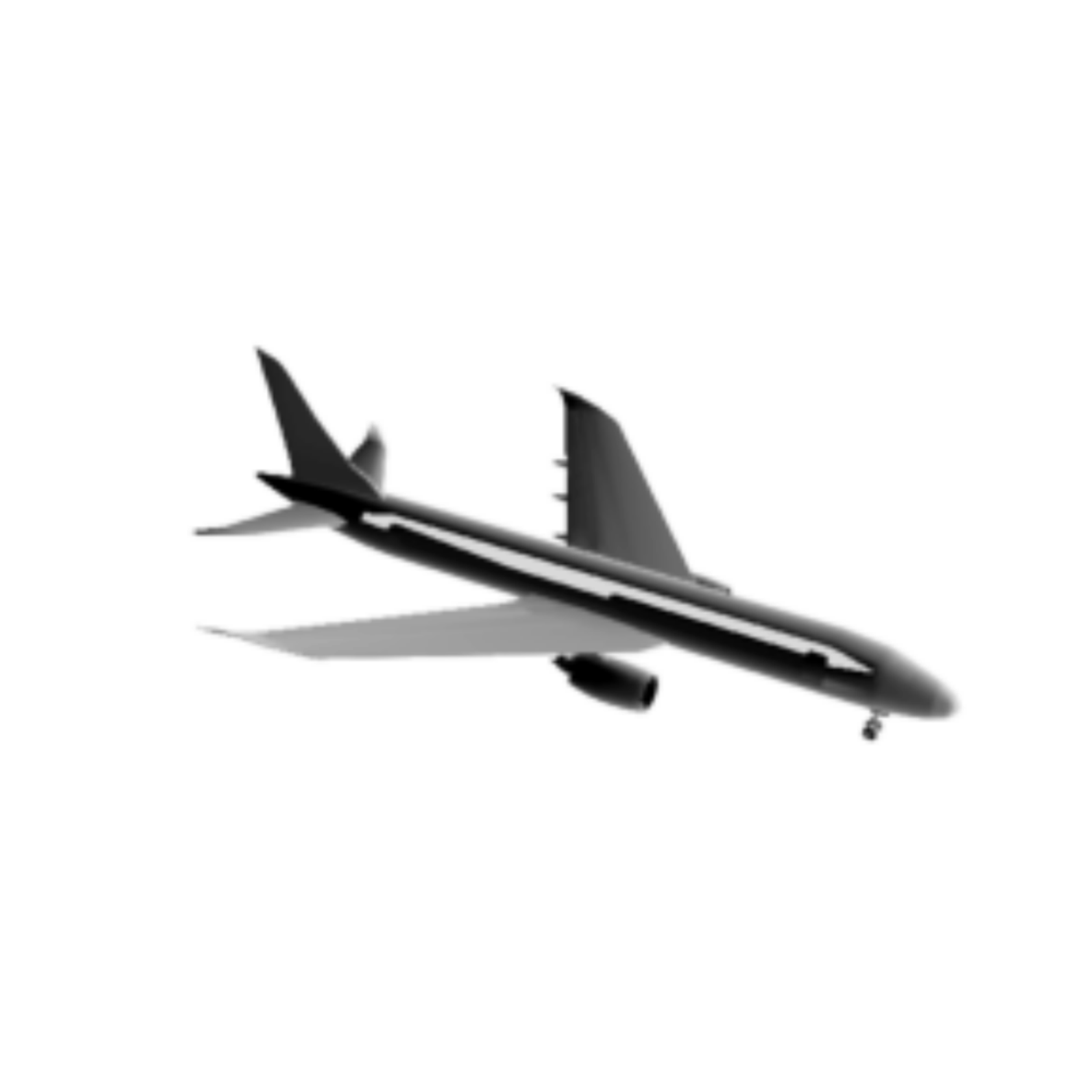}
\includegraphics[width=0.48\linewidth,trim={80mm 100mm 50mm 70mmm},clip]{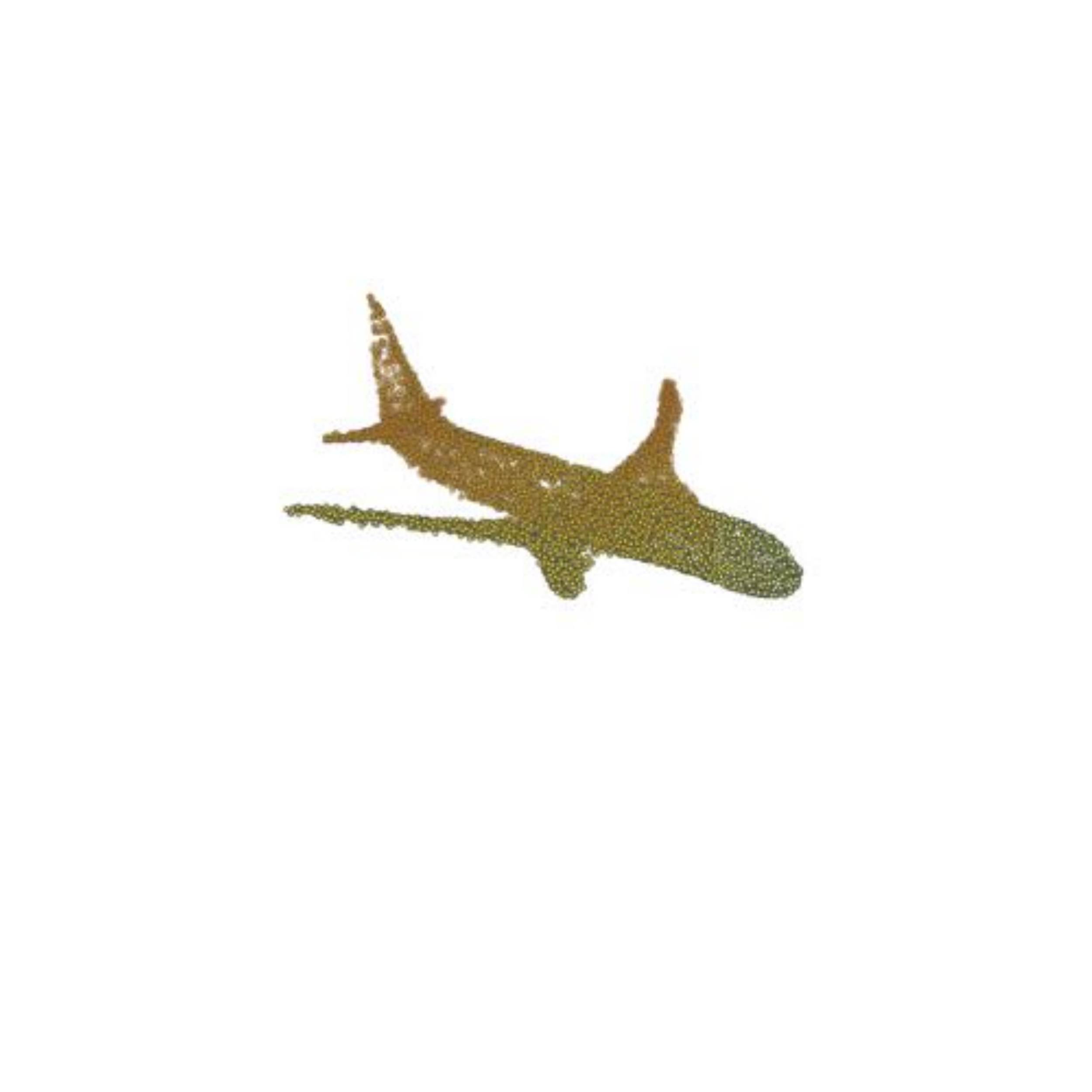}
\caption{Plane}
\end{subfigure}
\begin{subfigure}[b]{.4\linewidth}
\hfill
\includegraphics[width=0.4\linewidth,trim={10mm 10mm 10mm 10mm},clip]{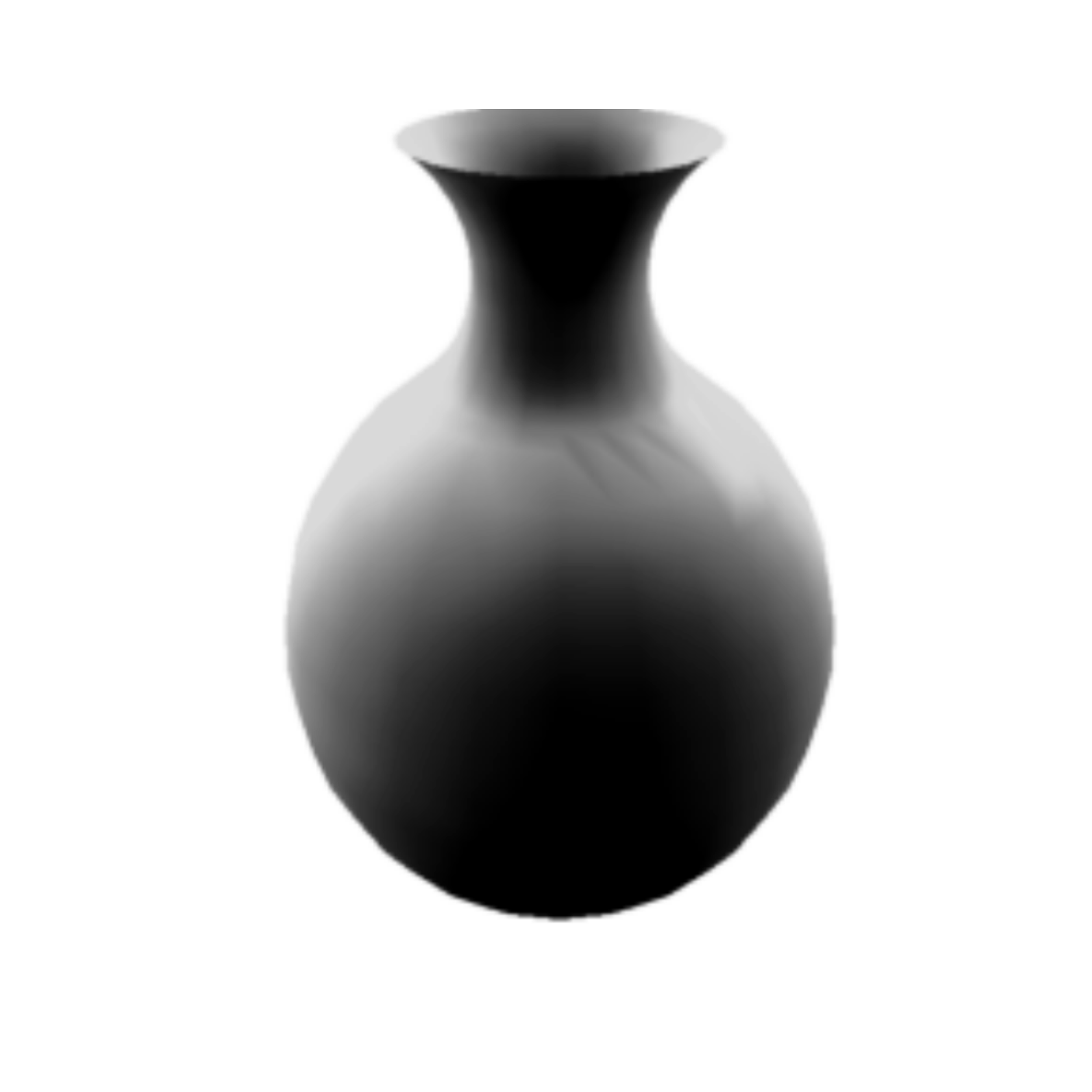}
\hfill
\includegraphics[width=0.4\linewidth,trim={45mm 50mm 45mm 45mmm},clip]{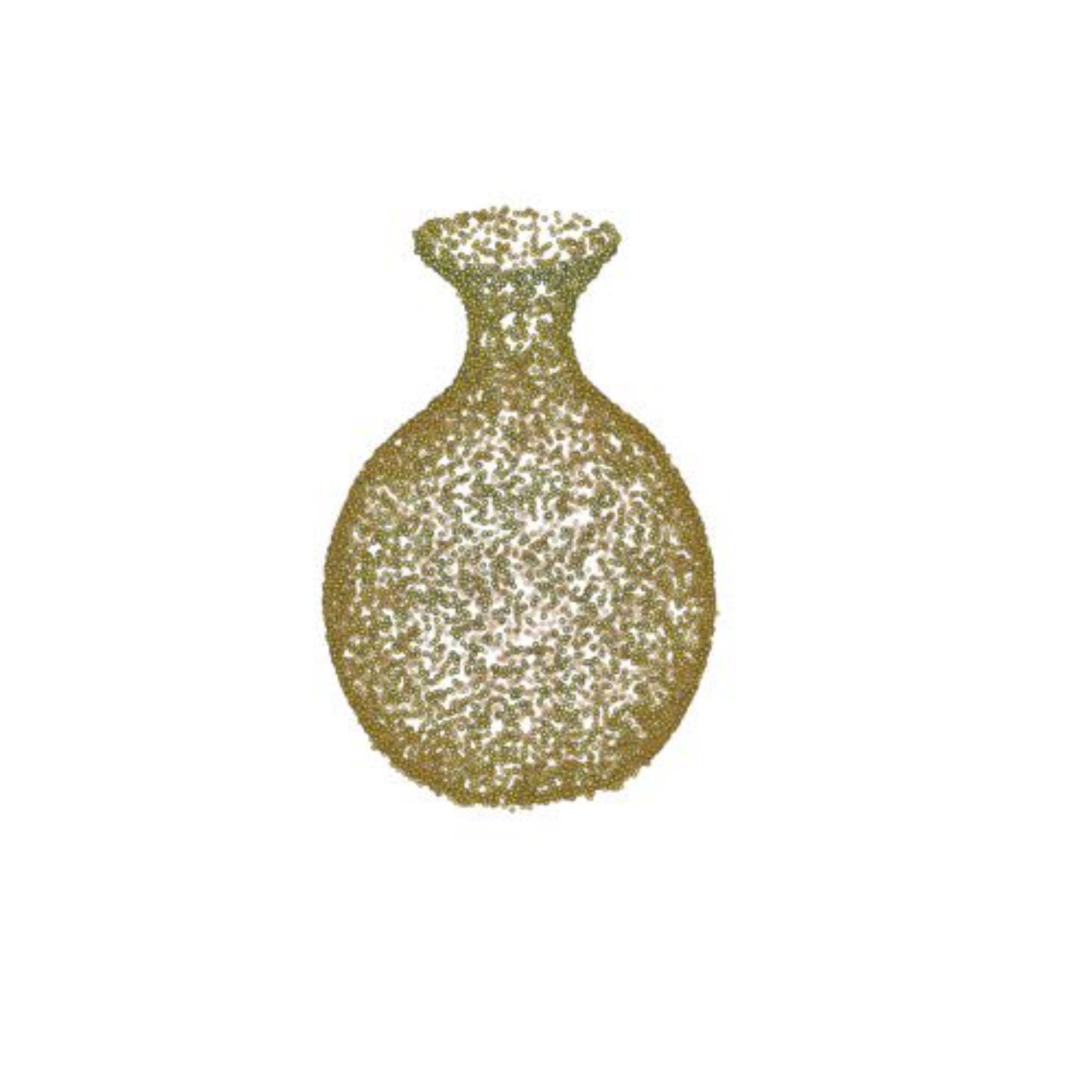}
\hfill
\caption{Vase}
\end{subfigure}
\caption{Images to Point Clouds.}
\label{fig:img2pc}
\end{figure}

Here we demonstrate a potential extension of the proposed PC-GAN for images to point cloud applications.
After training $Q$ as described in~\ref{sec:hierarchy}, instead of learning $G_\theta$ for hierarchical sampling, we
train a regressor $R$, where the input is the different views of the point cloud $X$, and the output is $Q(X)$. 
In this proof of concept experiment, we use the $12$ view data and the Res18 architecture in~\citet{su2015multi}, while
we change the output size to be $256$. 
Some example results on reconstructing testing data is shown in Figure~\ref{fig:img2pc}.
A straightforward extension is using end-to-end training instead of two-staged approached adopted here. Also, 
after aligning objects and take representative view along with traditional ICP techniques, we can also do single view
to point cloud transformation as~\citet{choy20163d, fan2017point, hane2017hierarchical, groueix2018atlasnet}, which is
not the main focus of this paper and we leave it for future work.

\section{Conclusion}

In this paper, 
we first showed a straightforward extension of existing GAN algorithm is not applicable to point clouds.
We then proposed a GAN modification (PC-GAN) that is capable of
learning to generate point clouds by using ideas both from hierarchical
Bayesian modeling and implicit generative models.
We further  propose a \emph{sandwiching} objective which results in a tighter Wasserstein distance estimate theoretically
and better performance empirically.

In contrast to some existing methods \citep{achlioptas2017learning}, PC-GAN can generate
arbitrary as many \emph{i.i.d.} points as we need to form a point clouds without
pre-specification. 
Quantitatively,
PC-GAN achieves competitive or better results using smaller network than existing methods. 
We also demonstrated  that PC-GAN can capture 
delicate details of point clouds and generalize well even on unseen data. 
Our method learns ``point-wise'' transformations which encourage the model to learn the
building components of the objects, instead of just naively copying the whole object. 
We also demonstrate other interesting results, including point cloud interpolation and image to point clouds.

Although we only focused on 3D applications in this paper, our framework can be naturally generalized to higher
dimensions. In the future we would like to explore higher dimensional applications, where each 3D point can
have other attributes, such as RGB colors and 3D velocity vectors.

\bibliographystyle{plainnat}
\bibliography{main}

\newpage
\appendix
\section{Technical Proof}
\setcounter{theorem}{0}
\begin{lemma}
\label{lem:sandwiching}
Suppose we have two approximators to Wasserstein distance: an upper bound $W_U$ and a lower $W_L$, such that $\forall
P,G: (1+\epsilon_1)w(\PP,\GG) \leq W_U(\PP,\GG) \leq (1+\epsilon_2)w(\PP,\GG)$ and $\forall P,G: (1-\epsilon_2)w(\PP,\GG) \leq W_L(\PP,\GG) \leq (1-\epsilon1)w(\PP,\GG)$ respectively, for some $\epsilon_2>\epsilon_1>0$ and $\epsilon_1>\epsilon_2/3$. Then, using the sandwiched estimator $W_\lambda$ from \eqref{eq:sandwich}, we can achieve tighter estimate of the Wasserstein distance than using either one estimator, \ie
\begin{equation}
\exists \lambda: |W_\lambda(\PP,\GG) - w(\PP,\GG)| < \min\{|W_U(\PP,\GG) - w(\PP,\GG)|, |W_L(\PP,\GG) - w(\PP,\GG)|\}
\end{equation}
\end{lemma}
\begin{proof}
We prove the claim by show that LHS is at most $\epsilon_1$, which is the lower bound for RHS.
\begin{equation}
\begin{aligned}
|W_\lambda&(\PP,\GG) - w(\PP,\GG)| \\
&= | (1-\lambda) W_U(\PP,\GG) + \lambda W_L(\PP,\GG) - w(\PP,\GG)|\\
&= | (1-\lambda) (W_U(\PP,\GG)-w(\PP,\GG)) - \lambda( w(\PP,\GG) - W_L(\PP,\GG))|\\
&\leq \max\{(1-\lambda)\underbrace{(W_U(\PP,\GG)-w(\PP,\GG))}_{\leq \epsilon_2}, \lambda\underbrace{( w(\PP,\GG) - W_L(\PP,\GG))}_{\leq \epsilon_2}\} \\
&\qquad\qquad - \min\{(1-\lambda)\underbrace{( W_U(\PP,\GG) - w(\PP,\GG))}_{\geq \epsilon_1}, \lambda\underbrace{( w(\PP,\GG) - W_L(\PP,\GG))}_{\geq \epsilon_1}\}\\
&\leq \max\{(1-\lambda), \lambda\}\epsilon_2 - \min\{(1-\lambda), \lambda\}\epsilon_1\\
\end{aligned}
\end{equation}
Without loss of generality we can assume $\lambda < 0.5$, which brings us to
\begin{equation}
|W_\lambda(\PP,\GG) - w(\PP,\GG)| \leq (1-\lambda)\epsilon_2 - \lambda \epsilon_1
\end{equation}
Now if we chose $\frac{\epsilon_2 - \epsilon_1}{\epsilon_2 + \epsilon_1} < \lambda < 0.5$, then $|W_\lambda(\PP,\GG) - w(\PP,\GG)| < \epsilon_1$ as desired.
\end{proof}

\begin{lemma}\label{prop:wc}
There exists $k>0$ such that 
\begin{equation}
	\max_{f\in \Omega_c \cap \Omega_{\phi}} \EE_{x \sim P}[f_\phi(x)] - \EE_{x \sim G}[f_\phi(x)] \leq \frac{1}{k}w(\PP,\GG)	
\end{equation}
\end{lemma}
\begin{proof}
Since there exists $k$ such that $\max_{f\in \Omega_c} \EE_{x \sim P}[f_\phi(x)] - \EE_{x \sim
G}[f_\phi(x)] \leq \frac{1}{k}w(\PP,\GG)$, it is clear that
\begin{equation}
\max_{f\in \Omega_c \cap \Omega_{\phi}} \EE_{x \sim P}[f_\phi(x)] - \EE_{x \sim G}[f_\phi(x)] \leq 
\max_{f\in \Omega_c} \EE_{x \sim P}[f_\phi(x)] - \EE_{x \sim G}[f_\phi(x)] \leq 
\frac{1}{k}w(\PP,\GG). 
\end{equation}
\end{proof}

\section{Permutation Equivariance Layers}
\label{sec:pelayer}
We briefly review the notion of Permutation Equivariance Layers proposed by~\citet{zaheer2017deep} as a background
required for this paper. 
For more details, please refer to \citet{zaheer2017deep}.

\citet{zaheer2017deep} propose a generic framework of deep learning for set data. The building block which can be stacked
to be deep neural networks is called Permutation Equivariance Layer. One Permutation Equivariance Layer example is defined 
as 
\[
	f(x_i) = \sigma( x_i +  \gamma \mbox{maxpool}(X) ),
\]
where  $\sigma$ can be any functions (\eg parametrized by neural networks) and $X = {x_1,\dots, x_n}$ is an input set. 
Also, the mox pooling operation can be replaced with mean pooling.
We note that PointNet\citet{qi2017pointnet} is a special case of using Permutation Equivariance Layer by properly defining
$\sigma(\cdot)$. 
In our experiments, we follow~\citet{zaheer2017deep} to set $\sigma$ to be a linear layer with output size $h$ followed
by any nonlinear activation function.

\section{Experiment Settings}
\label{sec:config}

\subsection{Synthetic Data}
The batch size is fixed to be $64$.
We sampled 10,000 samples for training and testing.

For the inference network, we stack $3$ mean Permutation Equivariance Layer~\citep{zaheer2017deep}, where the hidden
layer size (the output of the first two layers ) is $30$ and the final output size is $15$. 
The activation function are used SoftPlus.
For the generater is a $5$ layer MLP, where the hidden layer size is set to be $30$.
The discirminator is  $4$ layer MLP with hidden layer size to be $30$.
For ~\citet{achlioptas2017learning}, we change their implementation by replcing the number of filters for encoder to be
$[30,30,30,30,15]$, while the hidden layer width for decoder is $10$ or $20$ except for the output layer. 
The decoder is increased from 3 to 4 layers to have more capacity.

\subsection{ModelNet40}
We follow~\citet{zaheer2017deep} to do pre-processing. 
For each object, we sampled $10,000$ points from the mesh representation and normalize it to have zero mean (for each
axis) and unit (global) variance. 
During the training, we augment the data by uniformly rotating $0, \pi/8, \dots, 7\pi/8$ rad on the $x$-$y$ plane.
The random noise $z_2$ of PC-GAN is fixed to be $10$ dimensional for all experiments. 

For $Q$ of single class model, we stack $3$ max Permutation Equivariance Layer with output size to be $128$  for every
layer. On the top of the satck, we have a $2$ layer MLP with the same width and the output . 
The generator $G_x$ is a $4$ layer MLP where the hidden layer size is $128$ and output size is $3$. 
The discirminator is  $4$ layer MLP with hidden layer size to be $128$.

For training whole ModelNet40 training set, we increae the width to be $256$.  
The generator $G_x$ is a $5$ layer MLP where the hidden layer size is $256$ and output size is $3$. 
The discirminator is  $5$ layer MLP with hidden layer size to be $256$.
For hirarchical sampling, the top generator $G_\theta$ and discriminator are all $5$-layer MLP with hidden layer size to be $256$.

For AAE, we follow every setting used in ~\citet{achlioptas2017learning}, where the latent code size is $128$ and $256$
for single class model and whole ModelNet40 models.

\end{document}